\documentclass[11pt]{article}
\usepackage{enumerate}
\usepackage{pdfsync}
\usepackage[OT1]{fontenc}

\usepackage[usenames]{color}
\usepackage{smile}
\usepackage[colorlinks,
            linkcolor=red,
            anchorcolor=blue,
            citecolor=blue
            ]{hyperref}
\usepackage{fullpage}
\usepackage[protrusion=true,expansion=true]{microtype}
\usepackage{pbox}
\usepackage{setspace}
\usepackage{tabularx}
\usepackage{float}
\usepackage{wrapfig,lipsum}
\usepackage{enumitem}
\usepackage{microtype}
\usepackage{graphicx}
\usepackage{subfigure}
\usepackage{pgfplots}
\usepackage{mathtools}
\usepackage{caption}
\usepackage{framed}
\usetikzlibrary{arrows,shapes,snakes,automata,backgrounds,petri}
\usepackage{booktabs} 

\allowdisplaybreaks
\usepackage{colortbl}

\usepackage{xcolor}

\usepackage{makecell}
\usepackage{pifont}

\DeclareMathOperator{\msign}{msign}

\DeclareMathOperator{\EOS}{EOS}
\newcommand{\gammaGF}{\gamma_{\mathrm{GF}}}

\newcommand{\gammaSpecGF}{\gamma_{\mathrm{SpectralGF}}}

\makeatletter
\newcommand*{\rom}[1]{\expandafter\@slowromancap\romannumeral #1@}
\makeatother
\title{\huge Muon Learns Facts Better: Understanding the Role of Spectral Orthogonalization}
\author
{
    Xuheng Li\thanks{Department of Computer Science, University of California, Los Angeles, CA 90095, USA; e-mail: {\tt xuheng.li@cs.ucla.edu}}
    ~~
    Qiwei Di\thanks{Department of Computer Science, University of California, Los Angeles, CA 90095, USA; e-mail: {\tt qiwei2000@cs.ucla.edu}}
    ~~
    Yuan Cao\thanks{School of Computing and Data Science, the University of Hong Kong; e-mail: {\tt yuancao@hku.hk}}
    ~~
    Quanquan Gu\thanks{Department of Computer Science, University of California, Los Angeles, CA 90095, USA; e-mail: {\tt qgu@cs.ucla.edu}}
}

\ifdefined\final
\usepackage[disable]{todonotes}
\else
\usepackage[textsize=tiny]{todonotes}
\fi
\definecolor{shadecolor}{rgb}{0.78, 0.95, 0.89}

\definecolor{linkcolor}{HTML}{ED1C24}
\definecolor{LightCyan}{rgb}{0.58, 0.94, 0.85}
\hypersetup{colorlinks=true, linkcolor=linkcolor, urlcolor=black}
\usepackage[toc, page, header]{appendix}
\usepackage{titletoc}

\begin{document}
    \date{}
    \maketitle

\begin{abstract}
The Muon optimizer applies spectral orthogonalization to matrix-valued updates and has shown strong performance in large-scale neural network training, yet the mechanisms of this transformation in feature learning remain poorly understood.
In this work, we investigate this question through a tractable factual-recall model, where a fact maps each subject–relation pair to an answer, and a linear transformer learns the subject- and relation-dependent information required to recover this mapping.
The transformer is optimized with gradient flow (GF), spectral GF, or Sign GF, which are continuous-time limits of gradient descent, Muon, and Adam, respectively.
Prior studies~\citep{nichani2025understanding} have shown that when the number of subjects exceeds the number of relations, GF learns relation-dependent information before subject-dependent information, producing a feature-separation phase during training.
We characterize this separation with the learning times when the subject- and relation-dependent components of the prediction reach a target accuracy.
With $S$ subjects and $R$ relations, GF has a learning-time ratio of $\widetilde{\Theta}(\sqrt{S/R})$, whereas Spectral GF reduces this ratio to $\widetilde{\Theta}(1)$.
In addition, for fixed $S$ and $R$, the subject- and relation-dependent errors decay as $1/(T\log T)$ in training time $T$ under GF, but as $\exp(-\mathrm{poly}(T))$ under spectral GF.
Finally, we show that GF and spectral GF are equivariant under orthogonal transformations of the token embeddings, whereas Sign GF is not:
Different orthonormal embeddings can potentially produce no feature separation, a large feature-separation phase, or even a reversed learning order.
These results provide a mechanistic view of how spectral orthogonalization can fundamentally reshape feature-learning dynamics.
\end{abstract}

\section{Introduction}

\begin{figure}[ht]
\centering
\subfigure[Subject vs. relation accuracies]{\includegraphics[width=0.26\linewidth]{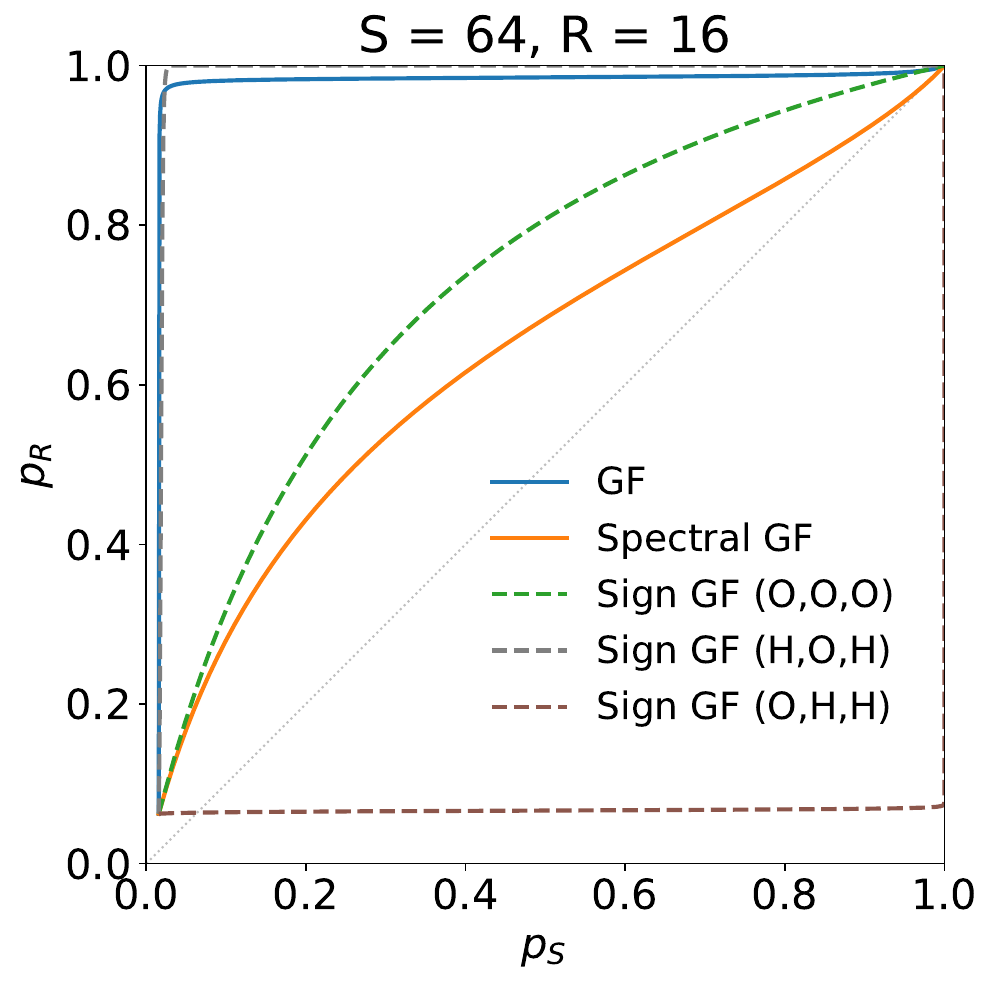}}
\subfigure[Dynamics of subject and relation components accuracies in GF]{\includegraphics[width=0.36\linewidth]{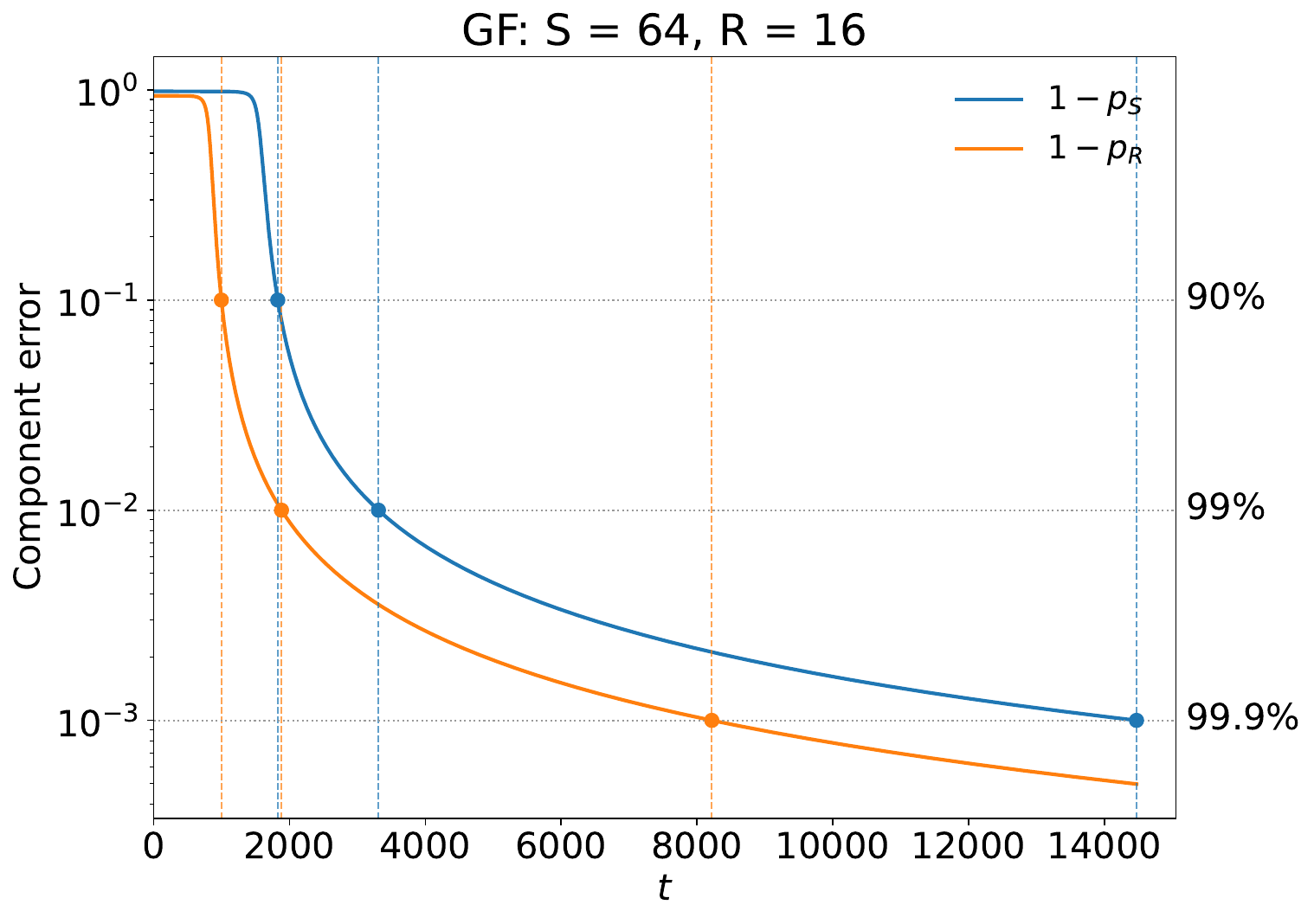}}
\subfigure[Dynamics of subject and relation component accuracies in Spectral GF]{\includegraphics[width=0.36\linewidth]{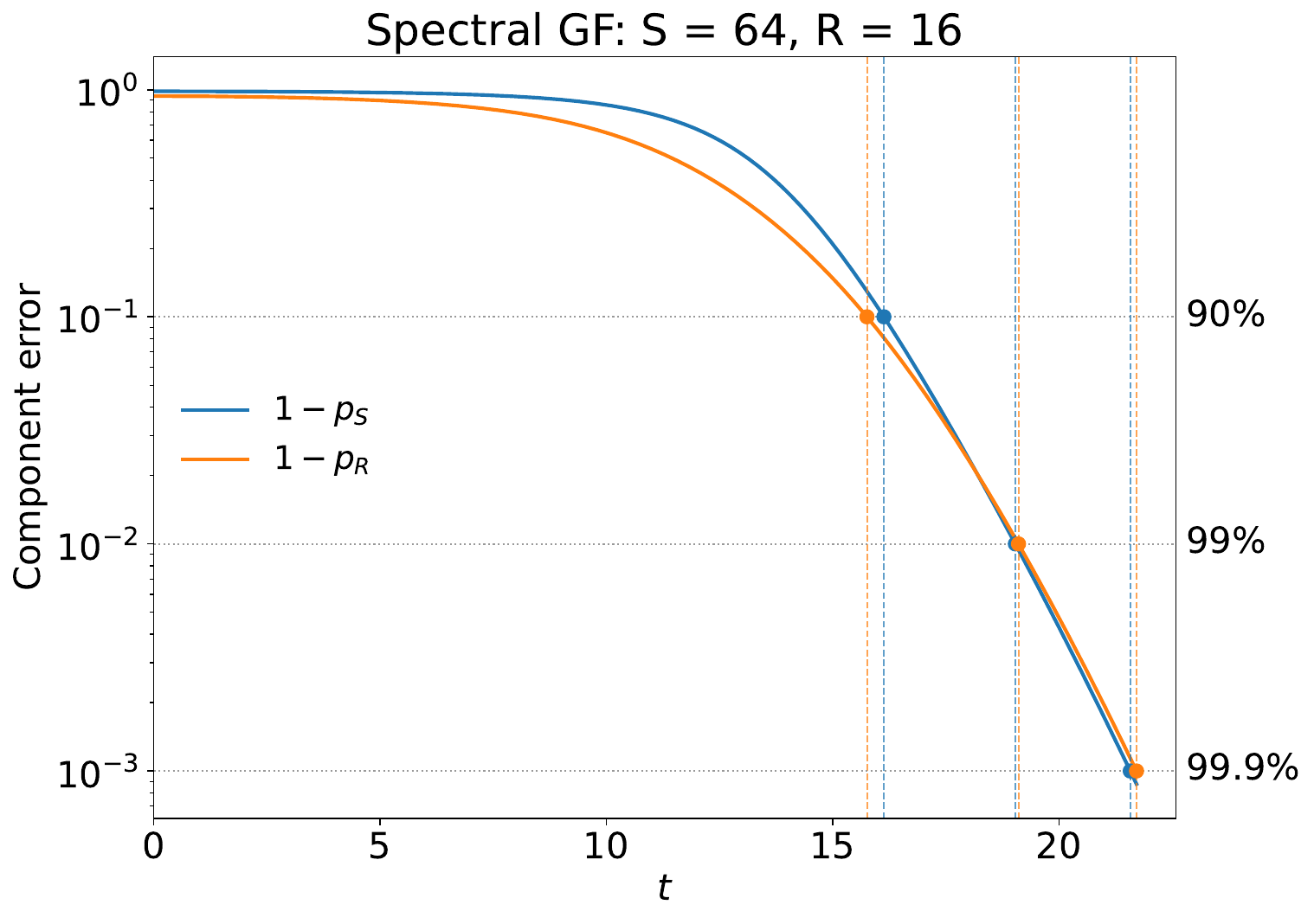}}
\caption{
Illustration of feature separation and softmax saturation. \textbf{(a):} Subject vs. relation components of model prediction accuracy. The top left portion of the plot for GF stands for feature separation, i.e., high accuracy of relation component but low accuracy of subject component; plot of Spectral GF is closer to the diagonal, indicating weaker feature separation; different embeddings in Sign GF can cause strong feature separation, weak separation, or reversed learning order (bottom right). \textbf{(b, c):} Dynamics of subject and relation component accuracies in GF and Spectral GF. For $(S, R)=(64, 16)$, the learning-time ratio of GF is approximately $2$ (consistent with theoretical prediction $\tilde\Theta(\sqrt{S/R})$); the ratio is close to $1$ in Spectral GF. Additionally, GF slows down when entering the high-accuracy regime due to softmax saturation, while Spectral GF does not.
}
\end{figure}
The Muon optimizer~\citep{jordan2024muon} has been widely applied to training large language models~\citep{liu2025muon, zeng2025glm, team2025kimi, xu2026deepseek}.
At the core of Muon is the spectral orthogonalization applied to momentum of matrix-valued parameters, e.g., QKV projections in self-attention and the linear weights in feed-forward networks.
For the corresponding momentum $\Mb$ corresponding to the weight $\Wb$, let $\Mb=\Ub\bSigma\Vb^\top$ be its singular value decomposition (SVD), where $\bSigma$ is a diagonal matrix with positive entries, and $\Ub$ and $\Vb$ have orthonormal columns.
Muon then calculates an approximation of the spectral orthogonalization
$\tilde{\msign}(\Mb)\approx\msign(\Mb):=\Ub\Vb^\top$, usually with Newton-Schulz iterations~\citep{guo2006schur, bernstein2024old}, and performs update
\begin{align*}
\Wb\gets\Wb-\gamma\cdot\tilde{\msign}(\Mb).
\end{align*}
Despite the empirical success of the Muon optimizer, the mechanism of the spectral orthogonalization is not fully understood.
A recent line of work suggests that spectral orthogonalization prevents frequent data from dominating the update, and achieves better performance on data with lower frequency, including
\citet{wang2025muon, vasudeva2026muon, li2026muon}.

\begin{table}[ht]
\caption{Comparison of learning times of GF, Spectral GF, and Sign GF dynamics in the factual recall task with $S$ subjects and $R$ relations, assuming $S\ge R\ge 4$.
The triplet after Sign GF stands for the embeddings corresponding to the subject, relation, and answer tokens: ``O'' and ``H'' stand for one-hot and Hadamard embeddings, respectively (See Section~\ref{subsection:embedding} for definition).
We use $T_S^\delta$ and $T_R^\delta$ to denote the learning times when the subject- and relation-dependent components of the prediction reach the target accuracy of $1-\delta$ under assumptions about the model initialization, and the ratio $T_S^\delta/T_R^\delta$ measures the duration of the feature-separation phase.
}\label{table:compare_all}
\centering\small
\begin{tabular}{c|c>{\color{blue}}cccc}
\toprule
\makecell{Optimization\\dynamics} & GF & Spectral GF & \makecell{Sign GF\\(O, O, O)} & \makecell{Sign GF\\(H, O, H)} & \makecell{Sign GF\\(O, H, H)}\\
\midrule
$T_S^\delta$ & $\tilde\Theta(S\sqrt R\cdot\textcolor{red}{\delta^{-1}})$ & $\tilde\Theta((SR)^{1/4})$ & $\tilde\Theta(1)$ & $\tilde\Theta((SR)^{1/4})$ & $\tilde\Theta((R/S)^{1/4})$\\
$T_R^\delta$ & $\tilde\Theta(R\sqrt S\cdot\textcolor{red}{\delta^{-1}})$ & $\tilde\Theta((SR)^{1/4})$ & $\tilde\Theta(1)$ & $\tilde\Theta((S/R)^{1/4})$ & $\tilde\Theta((SR)^{1/4})$\\
$T_S^\delta/T_R^\delta$ & $\tilde\Theta(\sqrt{S/R})$ & $\tilde\Theta(1)$ & $\tilde\Theta(1)$ & $\tilde\Theta(\sqrt R)$ & $\tilde\Theta(1/\sqrt S)$\\
\bottomrule
\end{tabular}
\end{table}

In practice, however, a prediction is not usually a simple mapping from the example to the class label. Instead, it can depend on several different \textbf{features} hidden in a single example.
The factual recall model~\citep{nichani2025understanding} models these tasks as a mapping $(s, r)\mapsto a$ from the subject-relation pair to the answer, e.g., $(\text{France}, \text{capital})\mapsto\text{Paris}$.
The factual recall model facilitates separate characterizations of the acquisition of subject- and relation-dependent information during training, which can occur at different timescales.
\citet{nichani2025understanding} studied how the attention block, whose weights have been interpreted as associative memory~\citep{geva2021transformer, bietti2023birth, cabannes2024learning}, learns the factual recall task when trained with gradient flow (GF).
They showed that the attention module learns the subject-dependent features later than relation-dependent features if the set of subjects is significantly larger than the set of relations.
We refer to this phenomenon as feature separation (See Figure~\ref{fig:main_text}).
We then raise the following questions about Muon:
\begin{quote}\centering\textit{
How does the Muon optimizer behave on the factual recall task? Does the similar feature separation phenomenon occur?}
\end{quote}

To answer this question, we study the GF, Spectral GF, and Sign GF, which are continuous-time limits of GD, Muon, and Adam, respectively.
We show that compared with GF, Spectral GF significantly reduces the learning-time ratio for the subject and relation components to reach the same target accuracy, indicating that spectral orthogonalization can suppress feature separation and save the training budget spent waiting for later-learned features to catch up.
More detailed contributions are summarized as follows:
\begin{itemize}[leftmargin=*]
\item[1.] We first compare GF and Spectral GF in a factual recall task with $S$ subject tokens and $R$ relation tokens ($S\ge R\ge4$). We show that the matrix orthogonalization of Muon/Spectral GF has two effects in the factual recall task: (i) it balances the learning of subject- and relation-dependent features, and reduces the learning-time ratio to $\tilde\Theta(1)$, while the learning-time ratio of GF is $\tilde\Theta(\sqrt{S/R})$; (ii) it additionally weakens the slowdown caused by softmax saturation, and reduces the $\delta^{-1}$-dependency in the learning times of GF to a logarithmic scale.
\item[2.] We show that both GF and Spectral GF are equivariant under orthogonal transformations of embeddings, while Sign GF is not. Sign GF combined with different orthonormal embeddings can thus display different dynamics, including a long feature-separation phase, no feature separation as in Spectral GF, or even reversed learning times of subject and relation.
\item[3.] Technically, we identify a low-dimension manifold $\cM$ from the data symmetry. On this manifold, the model prediction probability can be factored into the product of subject- and relation-dependent components. We further show that $\cM$ is an invariant manifold for all the learning dynamics of interest, including the Sign GF dynamics under different embeddings. This enables us to study the high-dimensional dynamics of the attention weights by focusing on the small number scalar parameters defining the manifold $\cM$.
\end{itemize}\label{fig:main_text}

\noindent\textbf{Notations.}
We use normal-face letters for scalars, boldface lower-case letters for vectors, and boldface upper-case letters for matrices.
$\one$ denotes the all-one vector or matrix; $\eb_i$ denotes the one-hot vector with the $i$-th entry being $1$.
$\delta_{i, j}$ denotes the Kronecker delta, i.e., $\delta_{i, j}=1$ iff. $i=j$; otherwise $\delta_{i, j}=0$.
For $N\in\NN_+$, $[N]$ denotes the set $\{1, 2, \dots, N\}$.
$\odot$ and $\oslash$ denotes element-wise product and division, and $\otimes$ denotes the Kronecker product.
We use the standard asymptotic notation $\Theta(\cdot)$, and $\tilde\Theta(\cdot)$ hides logarithmic factors.
We use $\simeq$, $\lesssim$, and $\gtrsim$ to hide constant factors in $=$, $\le$, and $\ge$.

\section{Related Work}

\label{section:related_work}

\subsection{Understanding the Muon Optimizer}

\noindent\textbf{Implicit bias and operator-norm margins.}
One line of work characterizes the solutions selected by spectral optimization as the training loss approaches zero.
For multiclass linear classification on separable data with cross-entropy loss, \citet{fan2025implicit} show that spectral descent and Muon achieve maximum margin with respect to the spectral, or operator, norm of the classifier matrix.
For homogeneous neural networks, \citet{tsilivis2024flavors} relate steepest-descent flows under general norms to the corresponding margin-maximization problems.
\citet{gronich2026implicit} extend this perspective to momentum-based methods, including Muon, under suitable decaying learning-rate schedules.
In these nonlinear settings, the guarantees concern KKT points of the margin problem rather than global maximizers.
These results characterize asymptotic implicit bias, whereas our analysis resolves the finite-time acquisition of subject- and relation-dependent information and the learning times required to reach a prescribed accuracy.

\noindent\textbf{Optimization geometry, convergence, and empirical behavior.}
Another line of work studies the geometry and convergence properties of matrix-valued updates.
\citet{bernstein2024old} interpret several optimizers through steepest descent under different norms, and \citet{kovalev2025understanding} places gradient orthogonalization in a spectral-norm trust-region framework.
\citet{shen2025convergence} derive convergence guarantees and identify Hessian structures under which Muon can improve over GD, while \citet{shulgin2026beyond} quantify the effects of inexact orthogonalization.
Empirically, \citet{khona2026soap} compare SOAP, Muon, and AdamW at large pretraining scales, including their stability and orthogonalization behavior.
\citet{ruan2026muon} investigate robustness and transferability and connect these properties to margins and feature diversity.
Our work complements these perspectives by explicitly characterizing how spectral orthogonalization changes the coupled feature-learning dynamics of an attention model.

\noindent\textbf{Balancing components and learning infrequent data.}
A closely related line of work explains Muon's benefits through more balanced learning across classes or spectral components.
\citet{wang2025muon} study class-imbalanced linear associative memories and show that Muon learns tail classes more effectively, while Adam's behavior can depend strongly on the embeddings.
\citet{vasudeva2026muon} analyze spectral gradient descent on Gaussian-mixture data with linear and deep linear models, showing that it learns principal components at equal rates and improves class-balanced performance early in training.
\citet{li2026muon} study linear associative memories with hierarchical query-answer frequencies, obtaining faster learning of low-frequency components and improved scaling laws.
In matrix factorization, \citet{rhee2026muon} likewise show that Muon avoids slow saddle-to-saddle dynamics and learns spectral modes concurrently.

Our results extend this picture to complementary features within a single prediction.
In our factual-recall task, subject-relation pairs and answer classes are sampled uniformly, and every example contains both a subject and a relation.
Nevertheless, when $S>R$, each particular subject occurs less often than each particular relation, and GF acquires the two components on different timescales.
We characterize how this asymmetry propagates through the coupled attention weights: Spectral GF reduces the learning-time ratio from $\tilde\Theta(\sqrt{S/R})$ to $\tilde\Theta(1)$ and weakens the slowdown caused by softmax saturation.
Thus, balancing the acquisition of features needed to identify one answer complements previous analyses of balancing errors across frequent and infrequent answer classes.

More balanced learning need not improve every task.
\citet{dragutinovic2026use} analyze simplified Muon in deep linear networks and linear attention and show that it can remove the sequential simplicity bias of GD.
Their experiments demonstrate that this can impair the discovery of shared representations across tasks or encourage reliance on spurious features; the relative strengths of the features also influence which optimizer performs better.
Our findings are consistent with theirs because the tasks reward different behavior.
Their routing task evaluates generalization to unseen input-output domain pairings, while their spurious-feature experiments examine reliance on features that may fail under distribution shift.
Our population factual-recall task requires learning both subject and relation information to identify the correct answer, and our guarantees concern recall under the same data distribution.
Accelerating the later-learned component is beneficial in this setting without implying that removing simplicity bias universally improves generalization.

\subsection{Factual Recall and Associative Memory}

\noindent\textbf{Linear associative memory.}
A linear associative memory stores mappings between fixed input and output embeddings in a trainable matrix, with retrieval determined by scores between transformed inputs and candidate outputs.
This setting isolates storage and retrieval from the surrounding sequence-processing architecture.
\citet{cabannes2024scaling} study memories formed from outer products of embeddings and derive scaling laws in sample size and parameter count.
\citet{cabannes2024learning} analyze gradient-descent training of a single memory module, showing how token frequencies and embedding correlations affect learning dynamics and interference between stored associations.
The Muon analyses of \citet{wang2025muon} and \citet{li2026muon} use this setting to study optimization under frequency imbalance.
Going beyond orthogonal embeddings, \citet{kim2026sharp} consider Gaussian input and output embeddings, allowing the number of associations to exceed the embedding dimension.
They characterize one-step recovery and storage capacity under power-law frequencies and analyze multi-step dynamics under a thresholded-gradient approximation.
These studies clarify how an optimizer learns associations stored in one matrix; our setting additionally requires learning interacting attention weights that combine subject and relation information from a sequence.

\noindent\textbf{Associative memory in transformer architectures.}
Associative memory also provides a framework for understanding how transformer components store and retrieve information.
\citet{geva2021transformer} interpret feed-forward layers as key-value memories, with keys detecting textual patterns and values contributing to the output-token distribution.
\citet{bietti2023birth} use a memory viewpoint to study a simplified transformer, explaining the learning of global bigram statistics and the subsequent emergence of an induction-head mechanism for in-context prediction.
These works connect individual memory modules to the computations performed by a sequence model.

Our closest architectural and task-level antecedent is \citet{nichani2025understanding}.
They show that a shallow transformer can combine attention and MLP associative memories to store facts with capacity nearly proportional to its parameter count.
They also analyze a simplified linear-attention model that learns relation information before subject information under GF.
We adopt their factual-recall setting and study how optimizer geometry changes this sequential learning behavior.
Specifically, we compare GF, Spectral GF, and Sign GF on an invariant manifold, quantify the subject and relation learning times, and identify the embedding dependence of Sign GF.
Our focus is therefore the dynamics of acquiring complementary factual information through attention.

\section{Problem Setting}

In this section, we sketch the data model and the linear attention model, following~\citet{nichani2025understanding}. We then introduce the continuous-time limits of different optimizers and their interplay with the embedding selection.

\subsection{Data Model}

We use the factual recall task~\citep{nichani2025understanding} to model predictions depending on multiple features. Let $\cS$, $\cR$, and $\cA$ be the sets of subject, relation, and answer tokens, respectively. A fact is a mapping $a_*:\cS\times\cR\to\cA$ from a subject-relation pair to an answer, e.g., $(\text{France}, \text{Capital})\mapsto\text{Paris}$. We assume that $a_*$ is a bijection.

The facts represented by triplets can be easily learned by a simple classification model, but in reality, facts are stored in sentences with complex structures and irrelevant information, e.g., ``the \textcolor{blue}{capital} of \textcolor{blue}{France} is \textcolor{blue}{Paris}''. To model such sentences, we study the following model where subject-relation pair is hidden in a sequence: Let $T\ge2$ be the context length. Let $\cN$ be the set of noise tokens, and $\EOS$ be a special token denoting the end of sequence. The sequence $(z_t)_{t\in[T+1]}$ is generated as follows:
\begin{itemize}[leftmargin=*]
\item[1.] Sample subject token $s\sim\mathrm{Uniform}(\cS)$ and relation token $r\sim\mathrm{Uniform}(\cR)$. The answer, which is the label of prediction represented by the next token of the sequence, is given by $z_{T+1}=a_*(s, r)$.
\item[2.] Uniformly sample two distinct indices $i, j\in[T-1]$ Set $z_i=s$ and $z_j=r$, and set $z_T=\EOS$.
\item[3.] For any index $k\in[T-1]$ such that $k\neq i, j$, draw $z_k$ uniformly from the noise tokens.
\end{itemize}
In this data model, all subjects, relations, and answers have equal probabilities. Therefore, the behavior of the optimizers on this data model is not affected by the data imbalance, and we can isolate the effect of \textbf{feature imbalance} with different optimizers because the only asymmetry lies in the different sizes of the subject and relation sets.

\subsection{Linear Attention}

We use a single-head linear attention to learn the factual recall task.
Let $\cV=\cS\cup\cR\cup\cA\cup\cN\cup\{\EOS\}$ denote the vocabulary, and $\bphi:\cV\to\RR^d$ is an embedding mapping that is fixed during training. We assume that the embeddings are orthonormal, i.e., for any $v, v'\in\cV$, we have $\langle\bphi(v), \bphi(v')\rangle=\delta_{v, v'}$. For input sequence $z_{1:T}$, let $\Xb=[\xb_1, \xb_2, \dots, \xb_T]^\top\in\RR^{T\times d}$ be the embedded sequence with $\xb_k=\bphi(z_k)$.
Position embedding is not considered because the answer $z_{T+1}$ is irrelevant to the ordering of the sequence $z_{1:T}$.
The linear attention model is given by
\begin{align*}
\fb(\Xb; \bTheta)=\Wb_{OV}\Xb^\top\Xb\Wb_{KQ}\xb_T,
\end{align*}
where $\bTheta=(\Wb_{OV}, \Wb_{KQ})$ are trainable parameters. $\Wb_{OV}$ merges the O and V parameters, and $\Wb_{KQ}$ merges the K and Q parameters, following~\citet{elhage2021mathematical, von2023transformers}. Let $\hat p_{\bTheta}(\cdot|z_{1:T})$ be the predicted next token distribution on the input sequence $z_{1:T}$:
\begin{align*}
\hat p_{\bTheta}(a|z_{1:T})=\frac{\exp(\langle\bphi(a), \fb(\Xb; \bTheta)\rangle)}{\sum_{a'\in\cA}\exp(\langle\bphi(a'), \fb(\Xb; \bTheta)\rangle)}.
\end{align*}
The model is trained by minimizing the cross entropy loss
\begin{gather*}
L(\bTheta)=\EE_{z_{1:T+1}}[\ell(z_{1:T+1}; \bTheta)],\quad \text{where}\quad\ell(z_{1:T+1}; \bTheta)=-\log\hat p_{\bTheta}(z_{T+1}|z_{1:T}).
\end{gather*}

\subsection{Optimizers and Continuous-Time Dynamics}

We consider the continuous-time limit of the GD, Adam, and Muon optimizers using the gradient of the population loss $L(\bTheta)$. Let $\gamma>0$ denote the learning rate.

\noindent\textbf{Gradient Flow (GF).} GF is the continuous-time limit of GD, given by
$\dot\bTheta(t)=-\nabla L(\bTheta(t))$.

\noindent\textbf{Sign Gradient Flow (Sign GF).} 
We follow \citet{zou2021understanding, ma2022qualitative, cattaneo2023implicit, xie2024implicit} and study Sign GF as an idealized continuous-time model of Adam. Let $\Gb_t=\nabla L(\bTheta_{t-1})$. The Adam update is given by
\begin{gather*}
\Mb_t=\beta_1\cdot\Mb_{t-1}+(1-\beta_1)\cdot\Gb_t,\quad\Vb_t=\beta_2\cdot\Vb_{t-1}+(1-\beta_2)\cdot\Gb_t\odot\Gb_t,\\
\bTheta_t=\bTheta_{t-1}-\gamma\cdot\Mb_t\oslash(\sqrt{\Vb_t}+\epsilon).
\end{gather*}
where $\beta_1, \beta_2\in(0, 1)$ are hyperparameters for the first-order and second-order momentum, respectively. If $(\beta_1, \beta_2)$ are fixed and $\gamma\to0$, these gradients are evaluated at nearly the same point and are thus all close to $\Gb_t$. Thus, $\Mb_t\approx\Gb_t$ and $\Vb_t\approx\Gb_t^{\odot2}$. If we also let $\epsilon\to0$, then $\Mb\oslash(\sqrt{\Vb_t}+\epsilon)\approx\sign(\Gb_t)$, where $\sign(\cdot)$ denotes the element-wise sign of a matrix. The Sign GF dynamics are thus given by $\dot\bTheta(t)=-\sign(\nabla L(\bTheta(t)))$.

\noindent\textbf{Spectral Gradient Flow (Spectral GF).}
The update of the Muon optimizer is
\begin{gather*}
\Mb_t^{OV}=\beta\cdot\Mb_{t-1}^{OV}+\nabla_{\Wb_{OV}}L(\bTheta_{t-1}),\quad\Wb_t^{OV}=\Wb_{t-1}^{OV}-\gamma\cdot\msign(\Mb_t^{OV});\\
\Mb_t^{KQ}=\beta\cdot\Mb_{t-1}^{KQ}+\nabla_{\Wb_{KQ}}L(\bTheta_{t-1}),\quad\Wb_t^{KQ}=\Wb_{t-1}^{KQ}-\gamma\cdot\msign(\Mb_t^{KQ});
\end{gather*}
where $\beta\in(0, 1)$ is momentum hyperparameter.
We study the ideal setting where the calculation of $\msign(\cdot)$ is exact. Similar to modeling Adam as Sign GF, we model Muon as Spectral GF:
\begin{align*}
\dot\Wb_{OV}(t)=-\msign\big(\nabla_{\Wb_{OV}}L(\bTheta(t))\big),\quad\dot\Wb_{KQ}(t)=-\msign\big(\nabla_{\Wb_{KQ}}L(\bTheta(t))\big).
\end{align*}

\subsection{Orthogonal Equivariance and Embedding Dependence}

In this section, we discuss how the choice of embeddings $\bphi(\cdot)$ can affect the learning dynamics. The following proposition shows that GF and Spectral GF are equivariant under different embeddings.

\begin{proposition}\label{prop:rotation_invariance}
Suppose that $\bTheta(t)$ follows either GF or Spectral GF dynamics under the embedding $\bphi(\cdot)$. Let $\Qb\in\RR^{d\times d}$ be an orthogonal matrix. Then $\tilde\bTheta(t)=(\Qb\Wb_{OV}(t)\Qb^\top, \Qb\Wb_{KQ}(t)\Qb^\top)$ also follows the GF or Spectral GF dynamics under the embedding $\tilde\bphi(\cdot)=\Qb\bphi(\cdot)$.
\end{proposition}
Proposition~\ref{prop:rotation_invariance} establishes that both GF and Spectral GF are equivariant under orthogonal transformations of embeddings (even when the embeddings are not orthonormal). Thus, for any two sets of orthonormal embeddings $\bphi$ and $\tilde\bphi$, let $\Qb$ be the orthogonal matrix satisfying $\tilde\bphi(\cdot)=\Qb\bphi(\cdot)$. Then by Proposition~\ref{prop:rotation_invariance}, there is a bijection between the dynamics under embeddings $\bphi(\cdot)$ and $\tilde\bphi(\cdot)$.

\begin{remark}\label{remark:signGF_embedding_dependent}
In general, Proposition \ref{prop:rotation_invariance} does not hold for Sign GF. A crucial property for Proposition~\ref{prop:rotation_invariance} to hold for Spectral GF is that for any matrix $\Mb$ and orthogonal matrices $\Ub$, $\Vb$, the spectral orthogonalization satisfies
$\msign(\Ub\Mb\Vb^\top)=\Ub\cdot\msign(\Mb)\cdot\Vb^\top$ (Lemma~\ref{lemma:msign_rotation}).
However, this property does not hold for the element-wise sign operator in general. In Section~\ref{section:SignGF}, we will study how the Sign GF dynamics change across different orthonormal embeddings of the same dictionary.
\end{remark}

\section{GF versus Spectral GF}\label{section:GF_SpectralGF}

We compare the GF and Spectral GF dynamics in this section. 
In general, the dynamics of $(\Wb_{OV},\Wb_{KQ})$ are high-dimensional and coupled, obscuring how the model learns subject- and relation-dependent information. Our data model, however, treats all subjects and relations symmetrically, suggesting that this learning process can be described by a small number of shared quantities rather than separate parameters for each subject and relation. 

Motivated by this symmetry, we introduce a low-dimensional manifold $\cM$ that isolates the feature directions relevant to factual recall.
By comparing the dynamics of GF and Spectral GF reduced to the manifold and their learning times, we identify two mechanisms of Spectral GF: reweighing different features in the update and alleviating softmax saturation.

\subsection{Reduced Parameterization and Learning Times}

We first define a low-dimensional manifold $\cM$, which plays a central role in our later analysis.
\begin{align}
\Wb_{OV}&=\sum_{s\in\cS}(\alpha_S\bar\vb_s+\gamma_S\bar\vb)\bphi(s)^\top+\sum_{r\in\cR}(\alpha_R\bar\vb_r+\gamma_R\bar\vb)\bphi(r)^\top+\sum_{n\in\cN}\gamma_N\bar\vb\bphi(n)^\top,\label{eq:def_WOV}\\
\Wb_{KQ}&=\bigg(\beta_S\sum_{s\in\cS}\bphi(s)+\beta_R\sum_{r\in\cR}\bphi(r)+\beta_N\sum_{n\in\cN}\bphi(n)\bigg)\bphi(\EOS)^\top,\label{eq:def_WKQ}
\end{align}
where $(\alpha_S, \alpha_R, \gamma_S, \gamma_R, \gamma_N, \beta_S, \beta_R, \beta_N)$ are parameters, and
\begin{align}\label{eq:def_barv}
\bar\vb_s=\frac{1}{\sqrt R}\sum_{r'\in\cR}\bphi(a_*(s, r')),\quad\bar\vb_r=\frac1{\sqrt S}\sum_{s'\in\cS}\bphi(a_*(s', r)),\quad\bar\vb=\frac1{\sqrt{SR}}\sum_{s'\in\cS}\sum_{r'\in\cR}\bphi(a_*(s', r')).
\end{align}
On this manifold, the model predictions can be expressed using only the scalar coefficients governing subject and relation learning, which allows us to define and analyze their respective learning times.
The subsequent sections show that the optimization dynamics of interest preserve this manifold, i.e., \textcolor{red}{if $\bTheta(0)\in\cM$, then $\bTheta(t)\in\cM$ for all $t\ge0$} (See Propositions~\ref{prop:GF}, \ref{prop:spectralGF}, and \ref{prop:signGF}).
Therefore, the reduced dynamics exactly describe training from the corresponding initialization.

The following proposition provides a clean expression of the model predictions and accuracy, provided that the model parameters $\bTheta\in\cM$:
\begin{proposition}\label{prop:prediction}
Suppose $\bTheta\in\cM$. Then the prediction of the linear attention model satisfies that for any $(s', r')$,
\begin{align*}
\hat p_{\bTheta}(a_*(s', r')|z_{1:T})=\frac{e^{\delta_{s', s(z_{1:T})}\cdot\alpha_S\beta_S/\sqrt R}}{S-1+e^{\alpha_S\beta_S/\sqrt R}}\cdot\frac{e^{\delta_{r', r(z_{1:T})}\cdot\alpha_R\beta_R/\sqrt S}}{R-1+e^{\alpha_R\beta_R/\sqrt S}}.
\end{align*}
Specifically, the model accuracy satisfies
\begin{align*}
\hat p_{\bTheta}(z_{T+1}|z_{1:T})=\frac{1}{1+(S-1)e^{-\alpha_S\beta_S/\sqrt R}}\cdot\frac{1}{1+(R-1)e^{-\alpha_R\beta_R/\sqrt S}}.
\end{align*}
\end{proposition}

Proposition~\ref{prop:prediction} shows that the model prediction $p(a|z_{1:T})$ is related to four parameters of the manifold $\cM$, namely $(\alpha_S, \beta_S, \alpha_R, \beta_R)$. The other parameters of the manifold $\cM$, including $\gamma_S, \gamma_R, \gamma_N$, and $\beta_N$, are irrelevant to the model prediction or accuracy. Therefore, to study the dynamics of the model accuracy, it suffices to study the dynamics of the parameters $(\alpha_S, \beta_S, \alpha_R, \beta_R)$.

Proposition~\ref{prop:prediction} also shows that the model prediction $p(a|z_{1:T})$ can be factored into the product of subject- and relation-dependent components. Thus, the model accuracy is also the product of two components. We define the margins as
\begin{align}\label{eq:def_q}
q_S=\alpha_S\beta_S/\sqrt R-\log(S-1),\quad q_R=\alpha_R\beta_R/\sqrt S-\log(R-1).
\end{align}
Then the model accuracy is $(1+e^{-q_S})^{-1}\cdot(1+e^{-q_R})^{-1}$.

\noindent\textbf{Learning times and feature-separation phase.}
To achieve the target accuracy of $1-\delta$, we require both $(1+e^{-q_S})^{-1}\ge1-\delta$ and $(1+e^{-q_R})^{-1}\ge1-\delta$, i.e., $\min\{q_S, q_R\}\ge\log(\delta^{-1}-1)$. Additionally, if the condition $\min\{q_S, q_R\}\ge\log(\delta^{-1}-1)$ holds, then the model accuracy is at least $1-2\delta$. Therefore, we consider the time when $q_S$ and $q_R$ reach the threshold $\log(\delta^{-1}-1)$:
\begin{align}\label{eq:def_Tdelta}
T_S^\delta:=\inf\{t: q_S(t)\ge\log(\delta^{-1}-1)\},\quad T_R^\delta:=\inf\{t:q_R(t)\ge\log(\delta^{-1}-1)\}.
\end{align}
The interval between $T_R^\delta$ and $T_S^\delta$ characterizes a \textbf{feature-separation phase} where the relation-dependent component of the prediction has already reached accuracy $1-\delta$, while the subject-dependent component has not. In this regime, the model has acquired substantially more information about the faster one of the two components, even though overall prediction accuracy is still limited by the slower component. This phase is referred to the hallucination phase in the literature~\citep{nichani2025understanding} because the model output $a\sim \hat p_{\bTheta}(\cdot|z_{1:T})$ corresponds to the correct relation token, while the subject token corresponding to $a$ is highly randomized. Intuitively, the model produces an answer of the right type but fails to associate it with the queried subject, yielding a plausible yet potentially factually incorrect response.

In subsequent sections, we use the ratio $T_S^\delta/T_R^\delta$ to quantify the duration of the feature-separation phase. This ratio largely removes the effect of the learning-rate because the constant learning rate $\gamma$ in the continuous-time dynamics is equivalent to time reparameterization, which introduces a shared factor of $\gamma^{-1}$ to both $T_S^\delta$ and $T_R^\delta$ (See Appendix~\ref{section:LR} for details).

\noindent\textbf{Initialization.} Throughout our analysis, we assume small, positive, and symmetric initializations of subject parameters $(\alpha_S, \beta_S)$ and relation parameters $(\alpha_R, \beta_R)$. The following assumption characterizes the positivity and symmetry of initializations, and we leave the discussion of the initialization scale to detailed analysis of optimization dynamics in subsequent sections.
\begin{assumption}\label{assumption:init}
The model parameters satisfy $\bTheta(0)\in\cM$, and there exist $\alpha_0, \beta_0>0$ such that
\begin{align*}
\alpha_S(0)=\alpha_R(0)=\alpha_0,\quad\beta_S(0)=\beta_R(0)=\beta_0.
\end{align*}
\end{assumption}
This initialization places the two sets of manifold parameters at the same magnitude at initialization, so differences between the learning times arise from the data and optimization dynamics exclusively.
The positivity assumption avoids sign changes in the reduced dynamics on the manifold and simplifies the expressions of learning times.

\subsection{GF Dynamics}

In this section, we study the GF dynamics in the manifold $\cM$.
We will calculate the expression of the gradient $\nabla L(\bTheta)$ and show that $\bTheta(t)\in\cM$ throughout training as long as $\bTheta(0)\in\cM$. We then immediately obtain the dynamics of the parameters $(\alpha_S, \beta_S, \alpha_R, \beta_R)$. Finally, we use the dynamics to characterize $T_S^\delta$ and $T_R^\delta$, and show that the feature separation ratio satisfies $T_S^\delta/T_R^\delta=\tilde\Theta(\sqrt{S/R})$.

The following proposition provides the expression of the gradient $\nabla L(\bTheta)$ when $\bTheta\in\cM$:
\begin{proposition}\label{prop:GF}
Suppose the model parameters $\bTheta\in\cM$. Define $\tilde\vb_s$ and $\tilde\vb_r$ as
\begin{align}\label{eq:def_tildev}
\tilde\vb_s=\bar\vb_s-\bar\vb/\sqrt{S},\quad\tilde\vb_r=\bar\vb_r-\bar\vb/\sqrt R.
\end{align}
Then the gradient at $\bTheta$ satisfies
\begin{align*}
-\nabla_{\Wb_{OV}}L(\bTheta)&=\frac{\beta_S/\sqrt R}{S-1+e^{\alpha_S\beta_S/\sqrt R}}\sum_{s\in\cS}\tilde\vb_s\bphi(s)^\top+\frac{\beta_R/\sqrt S}{R-1+e^{\alpha_R\beta_R/\sqrt S}}\sum_{r\in\cR}\tilde\vb_r\bphi(r)^\top;\\
-\nabla_{\Wb_{KQ}}L(\bTheta)&=\bigg[\frac{(S-1)\alpha_S/(S\sqrt R)}{S-1+e^{\alpha_S\beta_S/\sqrt R}}\sum_{s\in\cS}\bphi(s)+\frac{(R-1)\alpha_R/(R\sqrt S)}{R-1+e^{\alpha_R\beta_R/\sqrt S}}\sum_{r\in\cR}\bphi(r)\bigg]\bphi(\EOS)^\top.
\end{align*}
\end{proposition}
Proposition~\ref{prop:GF} shows that due to the definition of $\tilde\vb_s$ and $\tilde\vb_r$ in \eqref{eq:def_tildev}, the gradient $\nabla L(\bTheta)\in\cM$ as long as $\bTheta\in\cM$. Combined with the assumption that $\bTheta(0)\in\cM$, we conclude that \textbf{$\cM$ is an invariant manifold of the GF dynamics}. From the expression of the gradient, we also obtain the dynamics of $(\alpha_S, \beta_S, \alpha_R, \beta_R)$ expressed by the following ODE:
\begin{align*}
\dot\alpha_S=\frac{\beta_S/\sqrt R}{(S-1)(1+e^{q_S})},\quad\dot\beta_S=\frac{\alpha_S/\sqrt R}{S(1+e^{q_S})},\quad\dot\alpha_R=\frac{\beta_R/\sqrt S}{(R-1)(1+e^{q_R})},\quad\dot\beta_R=\frac{\alpha_R/\sqrt S}{R(1+e^{q_R})}.
\end{align*}
Although the parameters $\gamma_S$ and $\gamma_R$ change during training due to the presence of $\bar\vb$ in $\tilde\vb_s$ and $\tilde\vb_r$, their change does not affect the model accuracy, as is shown by Proposition~\ref{prop:prediction}. We also notice that the dynamics of $(\alpha_S, \beta_S, \alpha_R, \beta_R)$ are irrelevant to other manifold parameters. Therefore, to characterize the learning times $T_S^\delta$ and $T_R^\delta$, it suffices to study the above ODE. The following theorem provides an accurate estimate of $T_S^\delta$ and $T_R^\delta$:
\begin{theorem}\label{theorem:GF}
Assume that $S, R\ge2$, $\delta\le0.5$, and Assumption~\ref{assumption:init} holds with $\alpha_0, \beta_0<1$.
Then the learning times of the GF dynamics satisfy
\begin{align*}
T_S^\delta&\simeq\sqrt{SR(S-1)}\bigg\{\log\frac{\sqrt R\log(S-1)}{\alpha_0+\beta_0}+\frac{\delta^{-1}-1}{\log[(\delta^{-1}-1)(S-1)]}\bigg\},\\
T_R^\delta&\simeq\sqrt{SR(R-1)}\bigg\{\log\frac{\sqrt S\log(R-1)}{\alpha_0+\beta_0}+\frac{\delta^{-1}-1}{\log[(\delta^{-1}-1)(R-1)]}\bigg\}.
\end{align*}
\end{theorem}
The assumption of the initialization can be relaxed to $|\alpha_S(0)|, |\beta_S(0)|, |\alpha_R(0)|, |\beta_R(0)|\le1$, and the results are similar with minor changes to the term $\alpha_0+\beta_0$. We do not further analyze more general initializations because $\alpha_0+\beta_0$ contributes to only a logarithmic factor in $T_S^\delta$ and $T_R^\delta$.

\begin{remark}
We notice from the ODE that the subject parameters $(\alpha_S, \beta_S)$ and the relation parameters $(\alpha_R, \beta_R)$ form two isolated systems (the dynamics are independent from each other). We hereby interpret the behavior of the subject and relation parameters in Theorem~\ref{theorem:GF} separately, each consisting of two phases. For the subject parameters $(\alpha_S, \beta_S)$, the term
\begin{align*}
\sqrt{SR(S-1)}\log\frac{\sqrt R\log(S-1)}{\alpha_0+\beta_0}\simeq T_S^{0.5}
\end{align*}
corresponds to \textbf{Phase I} where $q_S\le1$. In this phase, the term $S-1+e^{\alpha_S\beta_S/\sqrt R}$, which stems from the normalizing factor of softmax, satisfies $S-1\ge e^{\alpha_S\beta_S/\sqrt R}$, so the evolution of $(\alpha_S, \beta_S)$ is close to a linear ODE system. The term
\begin{align*}
\sqrt{SR(S-1)}\cdot\frac{\delta^{-1}-1}{\log[(\delta^{-1}-1)(S-1)]}
\end{align*}
corresponds to \textbf{Phase II} where $q_S\ge1$. In this phase, we have $e^{\alpha_S\beta_S/\sqrt R}\ge S-1$, so the increase of $\alpha_S$ and $\beta_S$ slows down due to the saturated softmax. It takes time $\tilde\Theta(\delta^{-1})$, ignoring factors other than $\delta$, for the subject-dependent component in the model accuracy to reach $1-\delta$, which is consistent with prior works~\citep{soudry2018implicit, ji2018risk, ji2019implicit, lyu2019gradient}. Interpretations for the relation-dependent parameters $(\alpha_R, \beta_R)$ are similar.
\end{remark}
\begin{remark}
We now consider the learning time ratio $T_S^\delta/T_R^\delta$. If the target error rate $\delta$ is viewed as a constant, then the terms in $T_S^\delta$ and $T_R^\delta$ corresponding to Phase II can be ignored. If we further assume that the initialization is in the order of constants, i.e., $\alpha_0+\beta_0=\Theta(1)$, then
\begin{align*}
\frac{T_S^\delta}{T_R^\delta}\simeq\sqrt{\frac{S-1}{R-1}}\cdot\frac{\log(\sqrt R\log(S-1))}{\log(\sqrt S\log(R-1))}.
\end{align*}
If we aim for high accuracy, i.e., $\delta^{-1}\ge\max\{S, R, \log(\alpha_0+\beta_0)^{-1}\}$, then the term corresponding to Phase II is dominating. In this case, the learning-time ratio satisfies
\begin{align*}
T_S^\delta/T_R^\delta\simeq\sqrt{(S-1)/(R-1)}.
\end{align*}
In both cases, the learning-time ratio is $\tilde\Theta(\sqrt{S/R})$.
\end{remark}

\subsection{Spectral GF Dynamics}

Similar to the analysis of GF, we first provide closed-form expressions of $\msign(-\nabla_{\Wb_{OV}}L(\bTheta))$ and $\msign(-\nabla_{\Wb_{KQ}}L(\bTheta))$ in the following proposition:
\begin{proposition}\label{prop:spectralGF}
Suppose the model parameters $\bTheta\in\cM$. Let $\tilde\vb_s$ and $\tilde\vb_r$ be defined in \eqref{eq:def_tildev}. Then the spectral orthogonalization of the gradient at $\bTheta$ satisfies
\begin{align*}
&\msign(-\nabla_{\Wb_{OV}}L(\bTheta))=\sign(\beta_S)\cdot\sum_{s\in\cS}\tilde\vb_s\bphi(s)^\top+\sign(\beta_R)\cdot\sum_{r\in\cR}\tilde\vb_r\bphi(r)^\top.\\
&\msign(-\nabla_{\Wb_{KQ}}L(\bTheta))=\bigg[\frac{(S-1)^2\alpha_S^2}{(S-1+e^{\alpha_S\beta_S/\sqrt R})^2}+\frac{(R-1)^2\alpha_R^2}{(R-1+e^{\alpha_R\beta_R/\sqrt S})^2}\bigg]^{-1/2}\\
&\quad\cdot\bigg[\frac{(S-1)/\sqrt S\cdot\alpha_S}{S-1+e^{\alpha_S\beta_S/\sqrt R}}\sum_{s\in\cS}\bphi(s)+\frac{(R-1)/\sqrt R\cdot\alpha_R}{R-1+e^{\alpha_R\beta_R/\sqrt S}}\sum_{r\in\cR}\bphi(r)\bigg]\cdot\bphi(\EOS)^\top.
\end{align*}
\end{proposition}
To calculate $\msign(-\nabla_{\Wb_{OV}}L(\bTheta))$, we notice that the spectral orthogonalization of both $\sum_{s\in\cS}\tilde\vb_s\bphi(s)^\top$ and $\sum_{r\in\cR}\tilde\vb_r\bphi(r)^\top$ are themselves, and the two matrices have orthogonal rows (See Lemma~\ref{lemma:msign_grad}). To calculate $\msign(-\nabla_{\Wb_{KQ}}L(\bTheta))$, we notice that the gradient is a rank-one matrix $\ub\cdot\bphi(\EOS)^\top$, so its spectral orthogonalization is $(\ub/\|\ub\|_2)\cdot\bphi(\EOS)^\top$, i.e., the original matrix normalized by the magnitude of $\ub$.

Similar to Proposition~\ref{prop:GF} in the GF dynamics, Proposition~\ref{prop:spectralGF} shows that if the model initialization satisfies $\bTheta(0)\in\cM$, then $\bTheta(t)\in\cM$ throughout the Spectral GF dynamics. Therefore, $\cM$\textbf{ is an invariant manifold of the Spectral GF dynamics}. We then obtain dynamics of $(\alpha_S, \beta_S, \alpha_R, \beta_R)$:
\begin{align*}
&\dot\alpha_S=\sign(\beta_S),\quad\dot\beta_S=\frac{\alpha_S(1+e^{q_R})/\sqrt S}{\sqrt{\alpha_R^2(1+e^{q_S})^2+\alpha_S^2(1+e^{q_R})^2}};\\
&\dot\alpha_R=\sign(\beta_R),\quad\dot\beta_R=\frac{\alpha_R(1+e^{q_S})/\sqrt R}{\sqrt{\alpha_R^2(1+e^{q_S})^2+\alpha_S^2(1+e^{q_R})^2}}.
\end{align*}
If we assume the symmetric and positive initialization in Assumption~\ref{assumption:init},
then for all $t>0$, we have $\alpha_S(t)=\alpha_R(t)=t+\alpha_0$, and the $\alpha_S$ and $\alpha_R$ terms in the dynamics of $\beta_S$ and $\beta_R$ are canceled.
The analysis of the spectral GF is based on the simplified dynamics of $\beta_S$ and $\beta_R$, so it depends more directly on the symmetric initialization in Assumption~\ref{assumption:init}. Without Assumption~\ref{assumption:init}, the asymmetric initialization of $\alpha_S$ and $\alpha_R$ can complicate the analysis of the dynamics $\beta_S$ and $\beta_R$ in the early training phase. We leave the study of the general initialization for future work.

 From the simplified dynamics of $\beta_S$ and $\beta_R$, we observe that spectral orthogonalization affects the training dynamics with two major mechanisms:
\begin{itemize}[leftmargin=*]
\item[1.] \textbf{Spectral orthogonalization reweighs subject- and relation-dependent factors in the gradient of $\Wb_{OV}$.} It removes coefficients before $\sum_{s\in\cS}\tilde\vb_s\bphi(s)$ and $\sum_{r\in\cR}\tilde\vb_r\bphi(r)$, which cause the relation parameters to receive larger updates than subject parameters in the GF dynamics. This mechanism is consistent with the observations of prior works on GD/GF in classification tasks.
\item[2.] \textbf{Spectral orthogonalization weakens the slowdown caused by softmax saturation.} The dynamics of $\alpha_S$ and $\alpha_R$ are explicitly free of the terms $1+e^{q_S}$ and $1+e^{q_R}$ that cause softmax saturation in GF; although the ODEs of $\beta_S$ and $\beta_R$ contain such terms, their dynamics are coupled due to the spectral orthogonalization, satisfying $(\sqrt S\dot\beta_S)^2+(\sqrt R\dot\beta_R)^2=1$.
We further observe that if $q_S\le q_R$, then $\dot\beta_S\ge1/\sqrt{2S}$, and if $q_S>q_R$, then $\dot\beta_R\ge1/\sqrt{2R}$, i.e., the $\beta$ parameter corresponding to the smaller margin increases almost linearly. Therefore, the dependence of the learning times on the target error rate is much milder in the spectral GF dynamics.
\end{itemize}


Based on the ODE above, the following theorem characterizes the learning times $T_S^\delta$ and $T_R^\delta$:
\begin{theorem}\label{theorem:SpectralGF}
Suppose $S\ge R\ge 4$, and Assumption~\ref{assumption:init} holds with
$\alpha_0\le(SR)^{1/4}$ and $\beta_0\le(R/S)^{1/4}$.
Then for any target error rate $\delta\le0.5$, the learning times of spectral GF satisfy
\begin{align*}
T_S^\delta&\simeq(SR)^{1/4}\sqrt{\log[(\delta^{-1}-1)(S-1)]}.\\
T_R^\delta&\simeq\begin{cases}
(SR)^{1/4}\frac{\log[(\delta^{-1}-1)(R-1)]}{\sqrt{\log(R-1)}} & \text{if}\quad\log(\delta^{-1}-1)\le c\cdot\sqrt{\log(S-1)\cdot\log(R-1)};\\
(SR)^{1/4}\sqrt{\log[(\delta^{-1}-1)(S-1)]} & \text{otherwise}.
\end{cases}
\end{align*}
\end{theorem}
\begin{remark}
Compared with the GF dynamics, the learning times under the Spectral GF dynamics has no polynomial dependence on $\delta^{-1}$, which is attributed to the spectral orthogonalization weakening the effect of softmax saturation.
Unlike GF, the phase transition behavior of the learning times $T_S^\delta$ and $T_R^\delta$ under Spectral GF is asymmetric, due to the coupled dynamics of $\beta_S$ and $\beta_R$. We show that $\beta_S$ increases almost linearly in $t$ (See Lemma~\ref{lemma:betaS_lower_betaR_upper}), so $T_S^\delta$ has no phase transition behavior.
The two phases of \(T_R^\delta\) are determined by whether the desired relation accuracy is reached before or after the subject-dependent component becomes well learned (See proof for details).
\end{remark}
\begin{remark}
We now consider the learning-time ratio under the Spectral GF dynamics.
If the target error rate $\delta$ is viewed as a constant, then
\begin{align*}
T_S^\delta\simeq(SR)^{1/4}\sqrt{\log(S-1)},\quad T_R^\delta\simeq(SR)^{1/4}\sqrt{\log(R-1)},\quad\frac{T_S^\delta}{T_R^\delta}\simeq\sqrt{\frac{\log(S-1)}{\log(R-1)}}.
\end{align*}
We also consider the asymptotics of $\delta\to0$. In this case, the learning-time ratio is $T_S^\delta/T_R^\delta\simeq1$.
In both cases, we have $T_S^\delta/T_R^\delta=\tilde\Theta(1)$. Therefore, compared with the GF dynamics, the learning-time ratio under the spectral GF dynamics has no polynomial dependence on $S/R$.
This effect is mainly attributed to the spectral orthogonalization reweighing the subject- and relation-dependent components in the gradient of $\Wb_{OV}$.
\end{remark}

\section{Sign GF and Embedding Dependence}\label{section:SignGF}

\subsection{Orthonormal Embedding Constructions}
\label{subsection:embedding}

As is shown in Remark~\ref{remark:signGF_embedding_dependent}, the Sign GF dynamics rely on the choice of the embedding.
In this section, we construct different embeddings $\bphi(\cdot)$ where the Sign GF dynamics have different behavior.

To construct the full embedding mapping $\bphi(\cdot)$, we first construct embeddings for subject tokens $\bphi_S:\cS\to\RR^S$, for relation tokens $\bphi_R:\cR\to\RR^R$, and for answer tokens $\bphi_A:\cA\to\RR^{SR}$, separately. We then select orthonormal embeddings $\bphi_N$ for the noise tokens (they are irrelevant to the training dynamics, so the choice is arbitrary). We then lift $\bphi_S$, $\bphi_R$, $\bphi_A$, and $\bphi_N$ to a higher-dimensional space (with $\bphi(\EOS)$ as a one-hot vector) by padding the vectors with zeros, so that the subspaces are orthogonal to each other.

\begin{table}[ht]
\caption{Summary of dynamics and learning times for the eight possible combinations of embeddings. ``O'' and ``H'' stand for the one-hot and the Hadamard embeddings, respectively. The $\tilde\Theta(\cdot)$ in the learning times hides logarithmic factors of $\delta^{-1}$ as well as $S$ and $R$.}
\label{table:signGF}
\centering\small
\begin{tabular}{ccc|cccc|ccc}
\toprule
$\bphi_S$ & $\bphi_R$ & $\bphi_A$ & $\dot\alpha_S$ & $\dot\alpha_R$ & $\dot\beta_S$ & $\dot\beta_R$ & $T_S^\delta$ & $T_R^\delta$ & $T_S^\delta/T_R^\delta$\\
\midrule
O & O & O & $2\sqrt R$ & $2\sqrt S$ & $1$ & $1$ & $\tilde\Theta(1)$ & $\tilde\Theta(1)$ & $\tilde\Theta(1)$\\
O & O & H & $\sqrt S$ & $\sqrt R$ & $1$ & $1$ & $\tilde\Theta((R/S)^{1/4})$ & $\tilde\Theta((S/R)^{1/4})$ & \textcolor{red}{$\tilde\Theta(\sqrt{R/S})$}\\
O & H & O & $2\sqrt R$ & $\sqrt{SR}$ & $1$ & $1/\sqrt R$ & $\tilde\Theta(1)$ & $\tilde\Theta(1)$ & $\tilde\Theta(1)$\\
O & H & H & $\sqrt S$ & $1$ & $1$ & $1/\sqrt R$ & $\tilde\Theta((R/S)^{1/4})$ & $\tilde\Theta((SR)^{1/4})$ & \textcolor{red}{$\tilde\Theta(1/\sqrt S)$}\\
H & O & O & $\sqrt{SR}$ & $2\sqrt S$ & $1/\sqrt S$ & $1$ & $\tilde\Theta(1)$ & $\tilde\Theta(1)$ & $\tilde\Theta(1)$\\
H & O & H & $1$ & $\sqrt R$ & $1/\sqrt S$ & $1$ & $\tilde\Theta((SR)^{1/4})$ & $\tilde\Theta((S/R)^{1/4})$ & \textcolor{red}{$\tilde\Theta(\sqrt R)$}\\
H & H & O & $\sqrt{SR}$ & $\sqrt{SR}$ & $1/\sqrt S$ & $1/\sqrt R$ & $\tilde\Theta(1)$ & $\tilde\Theta(1)$ & $\tilde\Theta(1)$\\
H & H & H & $1$ & $1$ & $1/\sqrt S$ & $1/\sqrt R$ & $\tilde\Theta((SR)^{1/4})$ & $\tilde\Theta((SR)^{1/4})$ & $\tilde\Theta(1)$\\
\bottomrule
\end{tabular}
\end{table}

We construct one-hot embeddings and Hadamard embeddings for $\bphi_S$, $\bphi_R$, and $\bphi_A$. We assume that $S=2^p$ and $R=2^q$ for convenience of the Hadamard embedding construction.

\textbf{One-hot embeddings.} All embeddings are distinct one-hot vectors. Specifically, for any $(s, r)$,
\begin{align*}
\bphi_S^{\mathrm{O}}(s)=\eb_s,\quad \bphi_R^{\mathrm{O}}(r)=\eb_r,\quad\bphi_A^{\mathrm{O}}(a_*(s, r))=\eb_s\otimes\eb_r.
\end{align*}

\textbf{Hadamard embeddings.}
Hadamard embeddings are defined by columns of Hadamard matrices:
\begin{gather*}
\Hb_2=\frac1{\sqrt2}\begin{bmatrix}
1 & 1\\
1 & -1
\end{bmatrix},\quad\Hb_S=\Hb_2^{\otimes p},\quad\Hb_R=\Hb_2^{\otimes q};\\
\bphi_S^{\mathrm{H}}(s)=\Hb_S\eb_s,\quad\bphi_R^{\mathrm{H}}(r)=\Hb_R\eb_r,\quad\bphi_A^{\mathrm{H}}(a_*(s, r))=(\Hb_S\eb_s)\otimes(\Hb_R\eb_r).
\end{gather*}

\subsection{Sign GF Dynamics}

The embedding for each family of tokens $\bphi_S$, $\bphi_R$, and $\bphi_A$ can independently be one-hot or Hadamard, yielding $2^3=8$ possible combinations. The following proposition justifies that $\cM$ is an invariant manifold under the Sign GF dynamics with all combinations of embeddings, and gives a general form of the learning times:
\begin{proposition}\label{prop:signGF}
In all possible combinations of the embeddings defined above, if $\Theta\in\cM$, then the element-wise sign of the gradient satisfies $\sign(-\nabla L(\bTheta))\in\cM$. Furthermore, if Assumptions~\ref{assumption:init} holds, then the Sign GF dynamics satisfy $\dot\alpha_S=C_{\alpha_S}$, $\dot\beta_S=C_{\beta_S}$, $\dot\alpha_R=C_{\alpha_R}$, and $\dot\beta_R=C_{\beta_R}$ (details given in Table~\ref{table:signGF}). If $\alpha_0$ and $\beta_0$ are sufficiently small, then the learning times satisfy
\begin{align*}
T_S^\delta=\tilde\Theta(R^{1/4}(C_{\alpha_S}C_{\beta_S})^{-1/2}),\quad T_R^\delta=\tilde\Theta(S^{1/4}(C_{\alpha_R}C_{\beta_R})^{-1/2}).
\end{align*}
\end{proposition}

Assumption~\ref{assumption:init} is not essential and serves to avoid the sign change of the manifold parameters.
Table~\ref{table:signGF} gives the learning times and the learning-time ratios for all combinations of embeddings derived from Proposition~\ref{prop:signGF}. With certain embeddings, the learning-time ratio is $T_S^\delta/T_R^\delta=\tilde\Theta(1)$, matching the result of Spectral GF. However, the ratio can be as large as $\tilde\Theta(\sqrt{R})$ when the embeddings are $(\bphi_S, \bphi_R, \bphi_A)=(\bphi_S^{\mathrm{H}}, \bphi_R^{\mathrm{O}}, \bphi_A^{\mathrm{H}})$. Additionally, different from the GF and Spectral GF dynamics, $T_S^\delta/T_R^\delta$ can be smaller than $1$, indicating that the subject component can achieve the target accuracy earlier than the relation component. The embedding-dependency of Sign GD is also discussed in~\citet{wang2025muon}.

We notice that with the embeddings of interest, the learning times has only logarithmic dependence on $\delta^{-1}$. We conjecture that the element-wise sign can weaken the effect of softmax saturation similar to the spectral orthogonalization even with general embeddings, which is left as future work.

\section{Experiments}

In this section, we provide experiments to justify the theoretical findings. We first simulate the GF and spectral GF dynamics on the invariant manifold $\cM$ and verify the scaling of learning times. We then perform the experiments on the single-layer attention module learning the factual recall task studied in the theoretical part, but study more realistic optimizers with stochastic gradients, momentum, and trajectories out of the invariant manifold. Finally, we perform experiments on a realistic task with synthetic biographical data learned by a small language model.

\subsection{(Normalized) Gradient Flow Dynamics}

The experiment setting is exactly that considered in theoretical studies, so it suffices to simulate the ODEs. We focus on the GF and Spectral GF dynamics.

\subsubsection{Learning-Time Ratio Scaling with $S$ and $R$}

\begin{figure}[ht]
\centering
\subfigure[GF dynamics]{\includegraphics[width=0.48\linewidth]{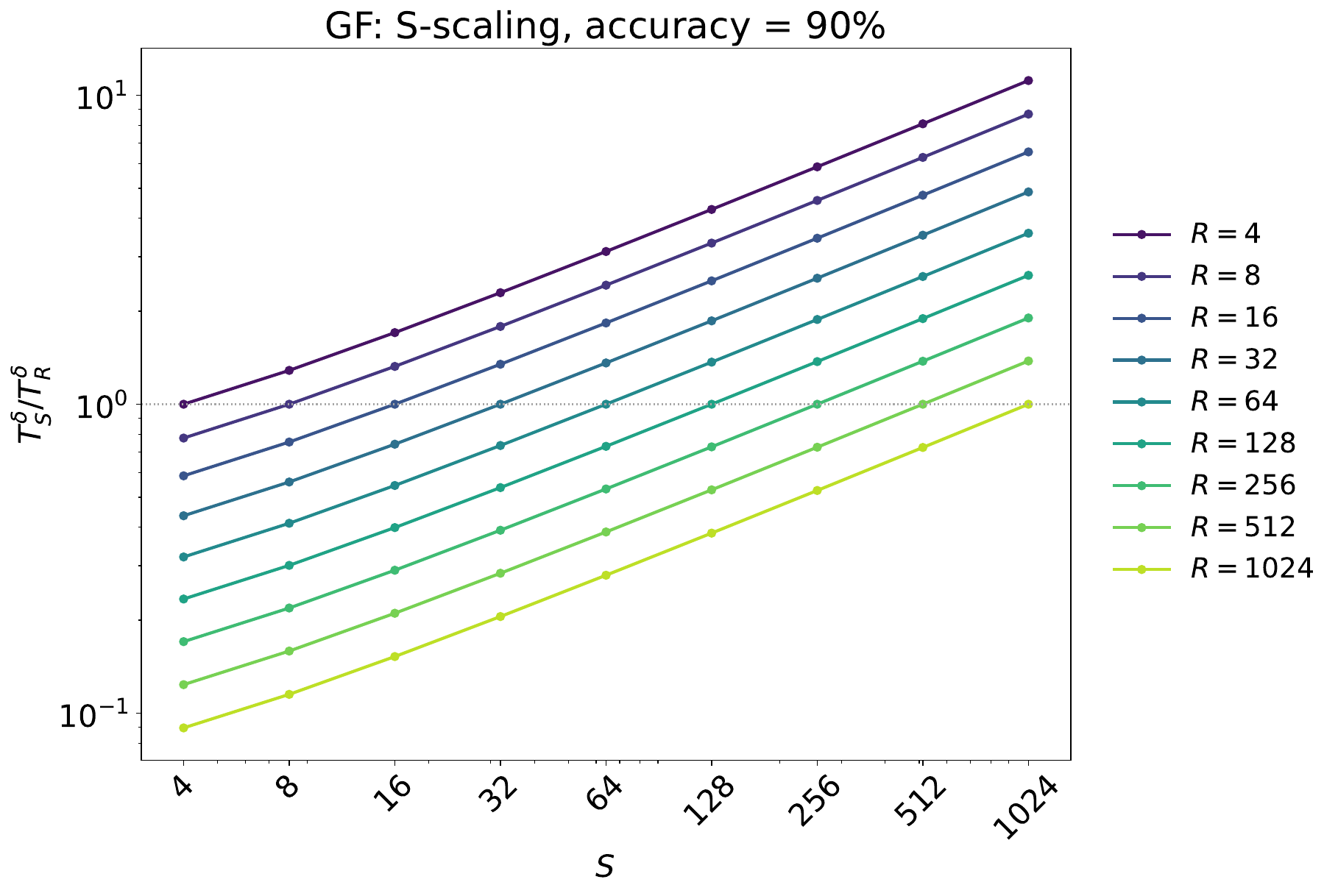}}
\subfigure[Spectral GF dynamics]{\includegraphics[width=0.48\linewidth]{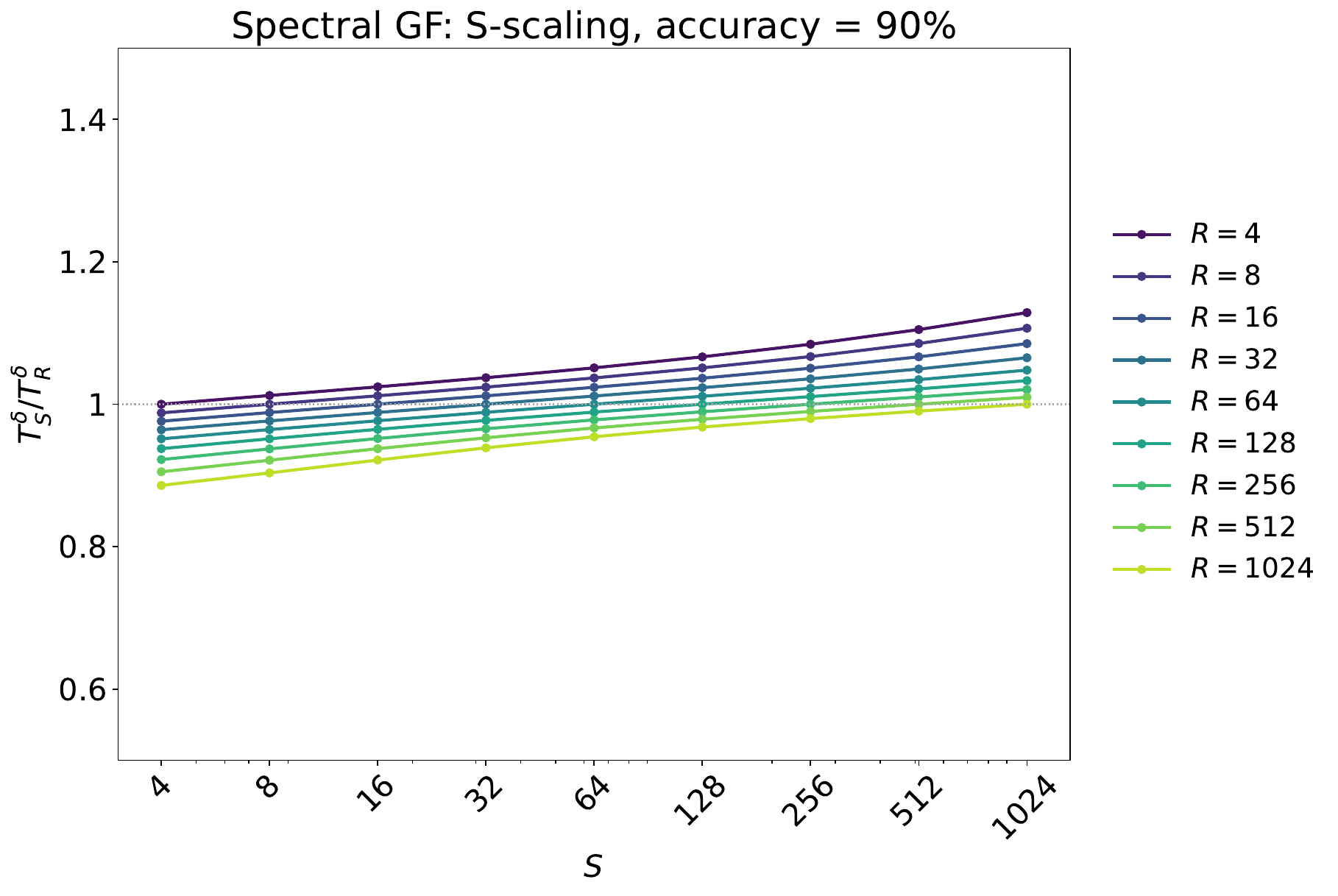}}
\caption{Scaling of learning-time ratio with $S$.}\label{fig:learning_time_ratio_S}
\end{figure}
\begin{figure}[ht]
\centering
\subfigure[GF dynamics]{\includegraphics[width=0.48\linewidth]{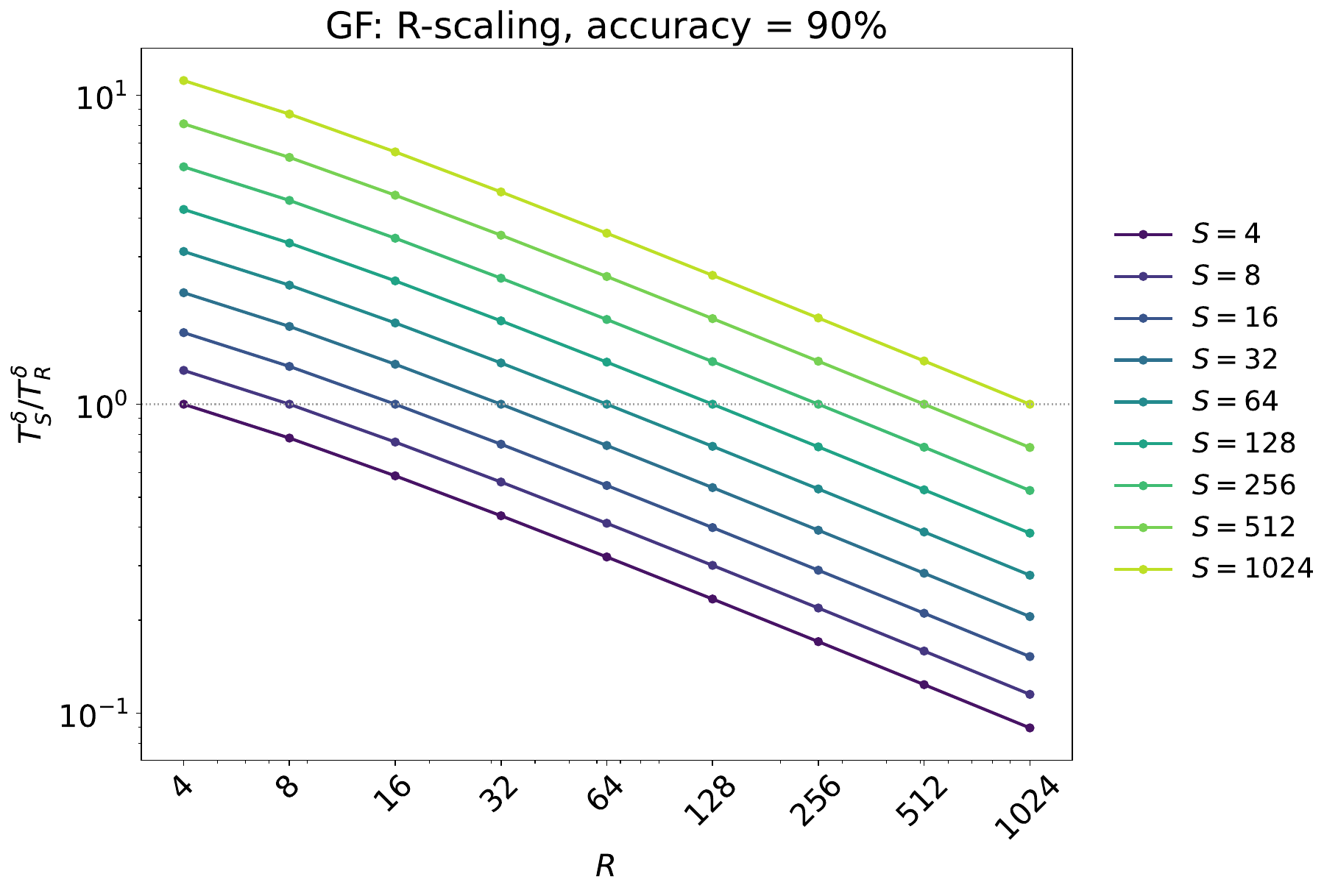}}
\subfigure[Spectral GF dynamics]{\includegraphics[width=0.48\linewidth]{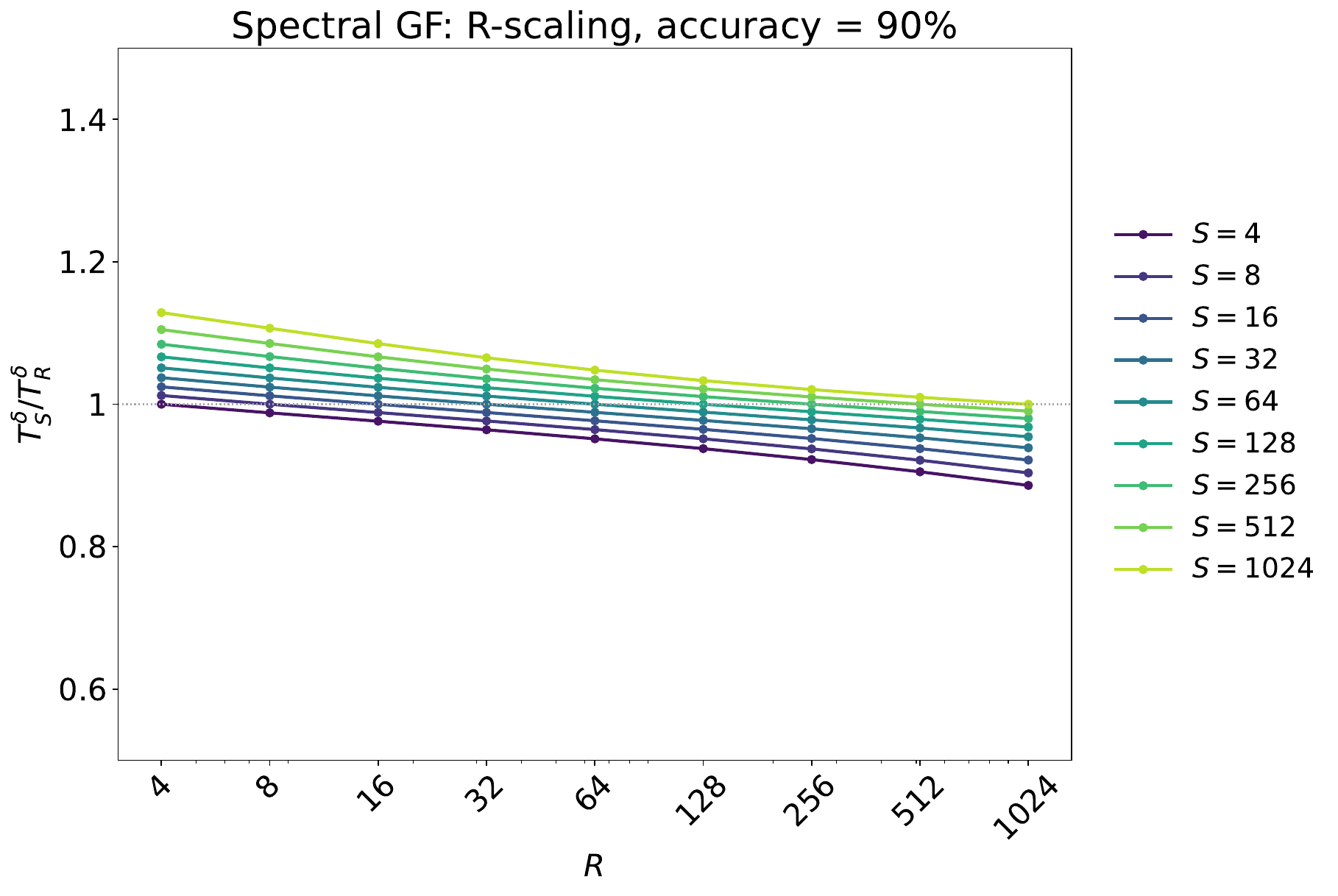}}
\caption{Scaling of learning-time ratio with $R$.}\label{fig:learning_time_ratio_R}
\end{figure}

In this section, we examine the scaling of learning-time ratio with $S$ and $R$. We vary both $S$ and $R$ between $4$ and $1024$ (including the case where $S<R$), and fix $\delta=0.1$, i.e., the target accuracy is 90\%. The results are shown in the log-log plots in Figures~\ref{fig:learning_time_ratio_S} and \ref{fig:learning_time_ratio_R}. In the log-log plot of $T_S^\delta/T_R^\delta$ against $S$ for the GF dynamics, the curves are are almost straight lines with slope close to $0.5$, indicating the ratio's $\sqrt{S}$-dependence on $S$. Similarly, the log-log plot of $T_S^\delta/T_R^\delta$ against $R$ indicates the ratio's $1/\sqrt R$-dependence on $R$. The learning-time ratio of the Spectral GF dynamics remain close to $1$ for all $S$ and $R$.

\subsubsection{Learning-Time Scaling with $\delta$}
\begin{figure}[ht]
\centering
\subfigure[GF dynamics]{\includegraphics[width=0.48\linewidth]{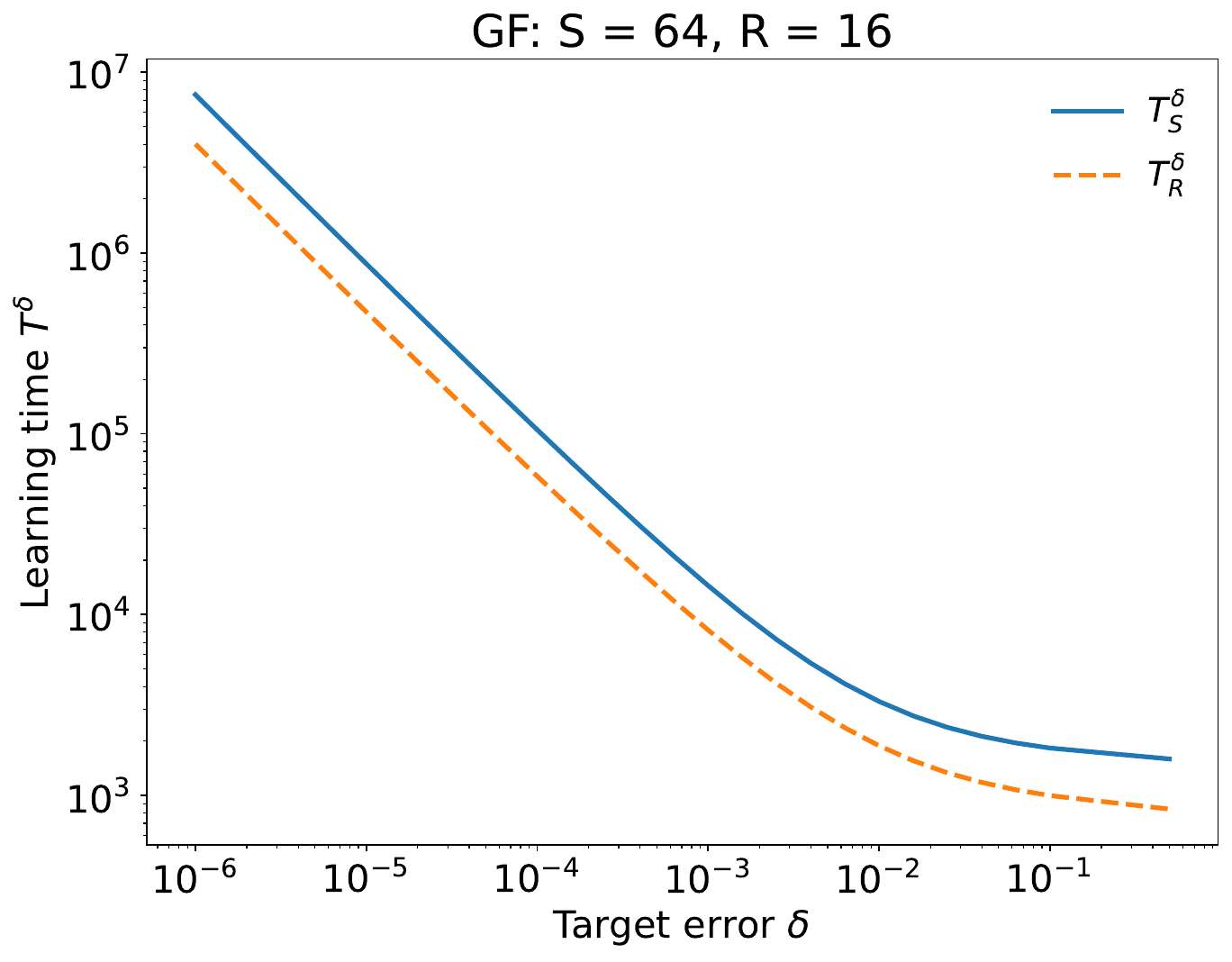}}
\subfigure[Spectral GF dynamics]{\includegraphics[width=0.48\linewidth]{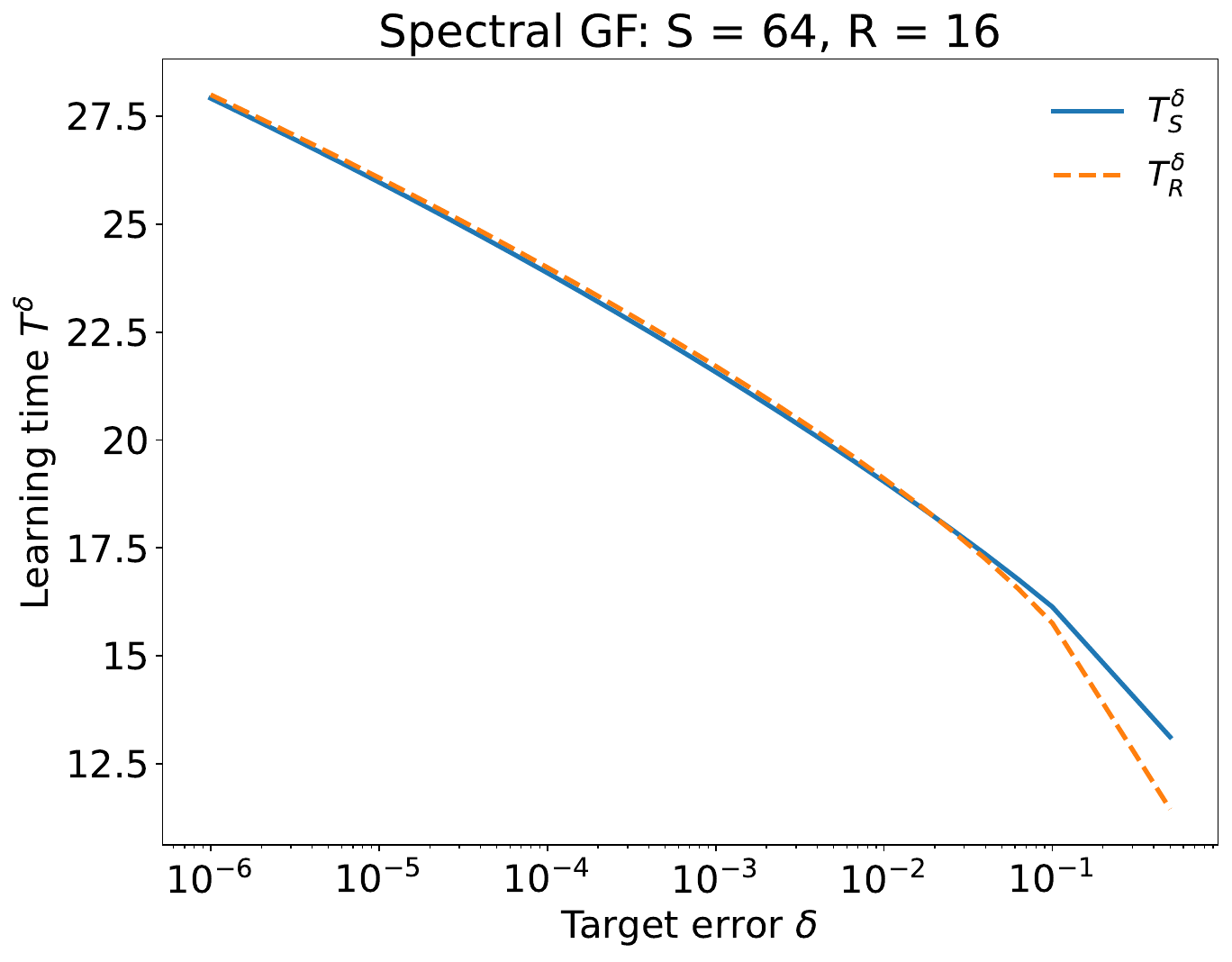}}
\caption{Scaling of learning times with $\delta$.}
\label{fig:scaling_delta}
\end{figure}

In this section, we examine the scaling of learning time with the target error rate $\delta$. We fix $(S, R)=(64, 16)$. Results are shown in Figure~\ref{fig:scaling_delta}.

For the GF dynamics, we observe from the log-log plot that for small $\delta$, both $T_S^\delta$ and $T_R^\delta$ scales with $\delta^{-1}$, while there is a phase transition for large $\delta$, which is consistent with the phase transition given in Theorem~\ref{theorem:GF}. For Spectral GF, Figure~\ref{fig:scaling_delta}(b) shows a logarithmic dependence of the learning times on $\delta^{-1}$. Additionally, $T_R^\delta$ is close to $T_S^\delta$ for small $\delta$, but for larger $\delta$, we have $T_R^\delta<T_S^\delta$. This phenomenon is consistent with the phase transition of $T_R^\delta$ in Theorem~\ref{theorem:SpectralGF}.

\subsection{Single-Layer Attention Training}

In this section, our goal is to show that the theoretical findings hold for more realistic training scenarios with stochastic gradient, momentum updates, and iterates out of the invariant manifold.
We still focus on the original recall data model learned by the single-layer linear attention module. We use a cosine learning schedule for all optimizers, with the peak learning rate being $0.01$ for SGD, $0.02$ for Muon, and $0.0003$ for SGD. Following~\citet{team2025kimi}, we apply a $0.2$ multiplier to the updates of attention weights optimized with Muon, which balances the updates of weights optimized with Muon and Adam. We use the standard 5-step Newton-Schulz iteration to approximate $\msign(\cdot)$, following the standard implementation~\citep{jordan2024muon}. We train the model for 50000 steps with batch size 128. Experiment results are averaged over 3 random seeds of initializations.

\begin{figure}[H]
\centering
\subfigure[SGD]{\includegraphics[width=0.32\linewidth]{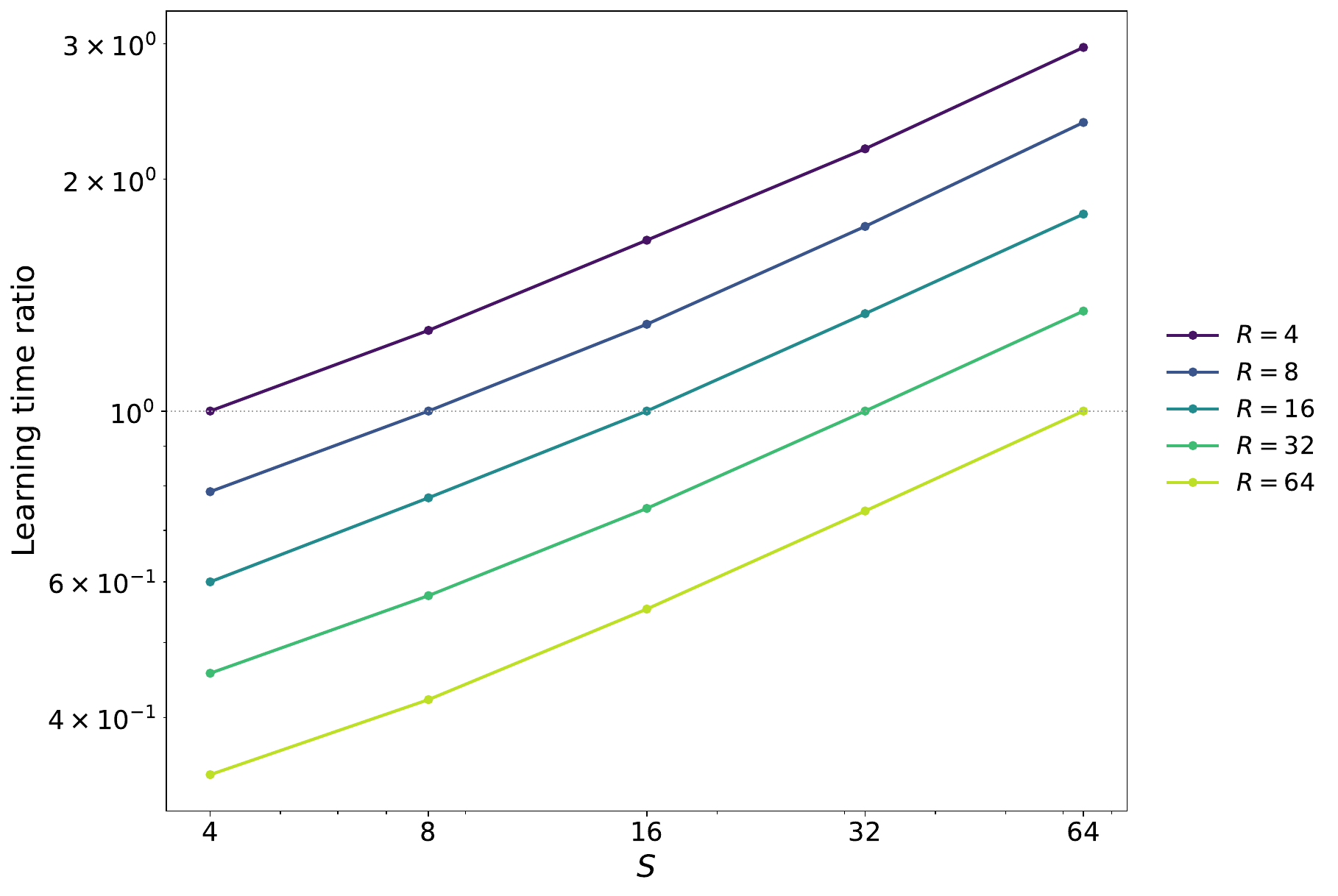}}
\subfigure[Muon]{\includegraphics[width=0.32\linewidth]{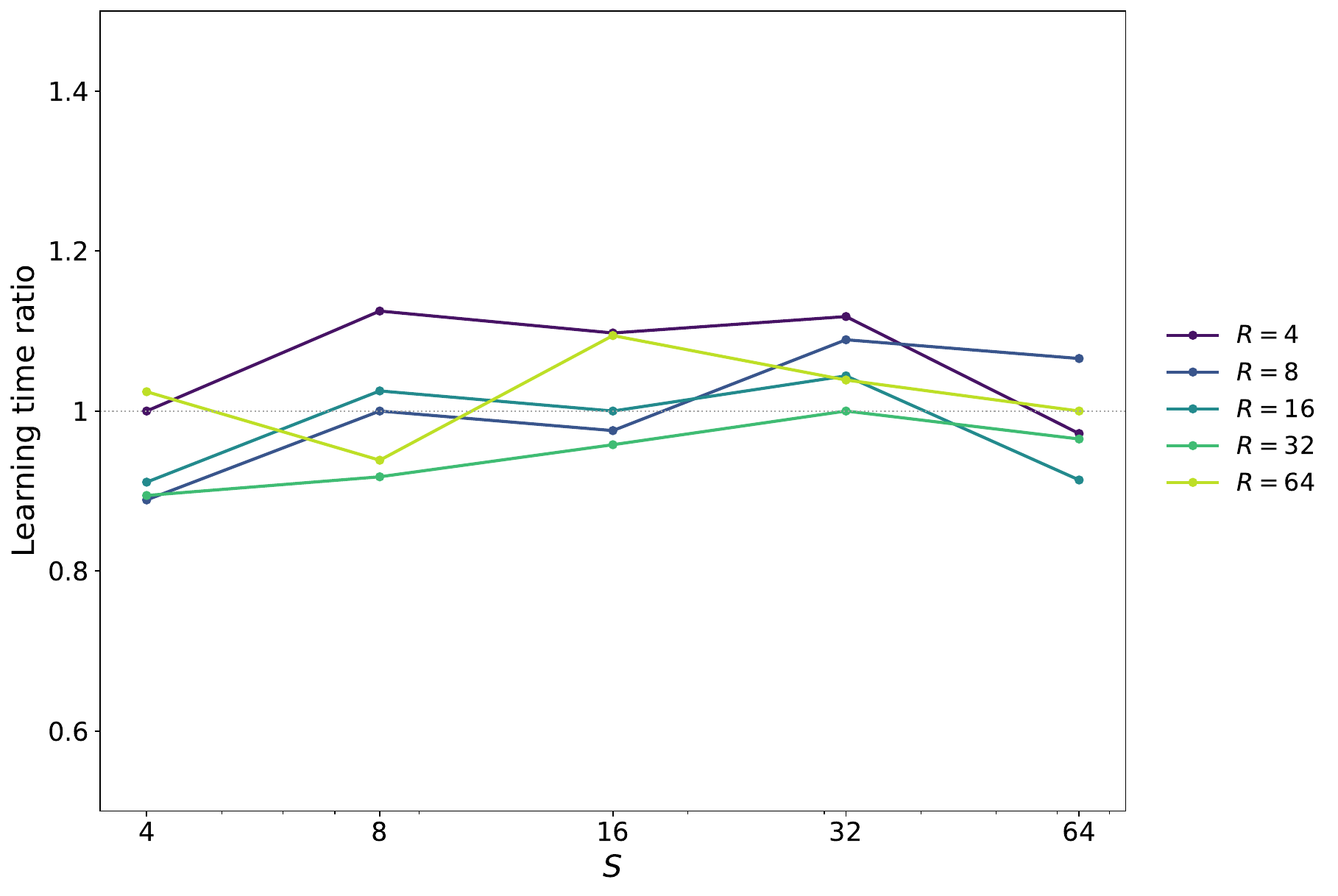}}
\subfigure[AdamW (O, O, O)]{\includegraphics[width=0.32\linewidth]{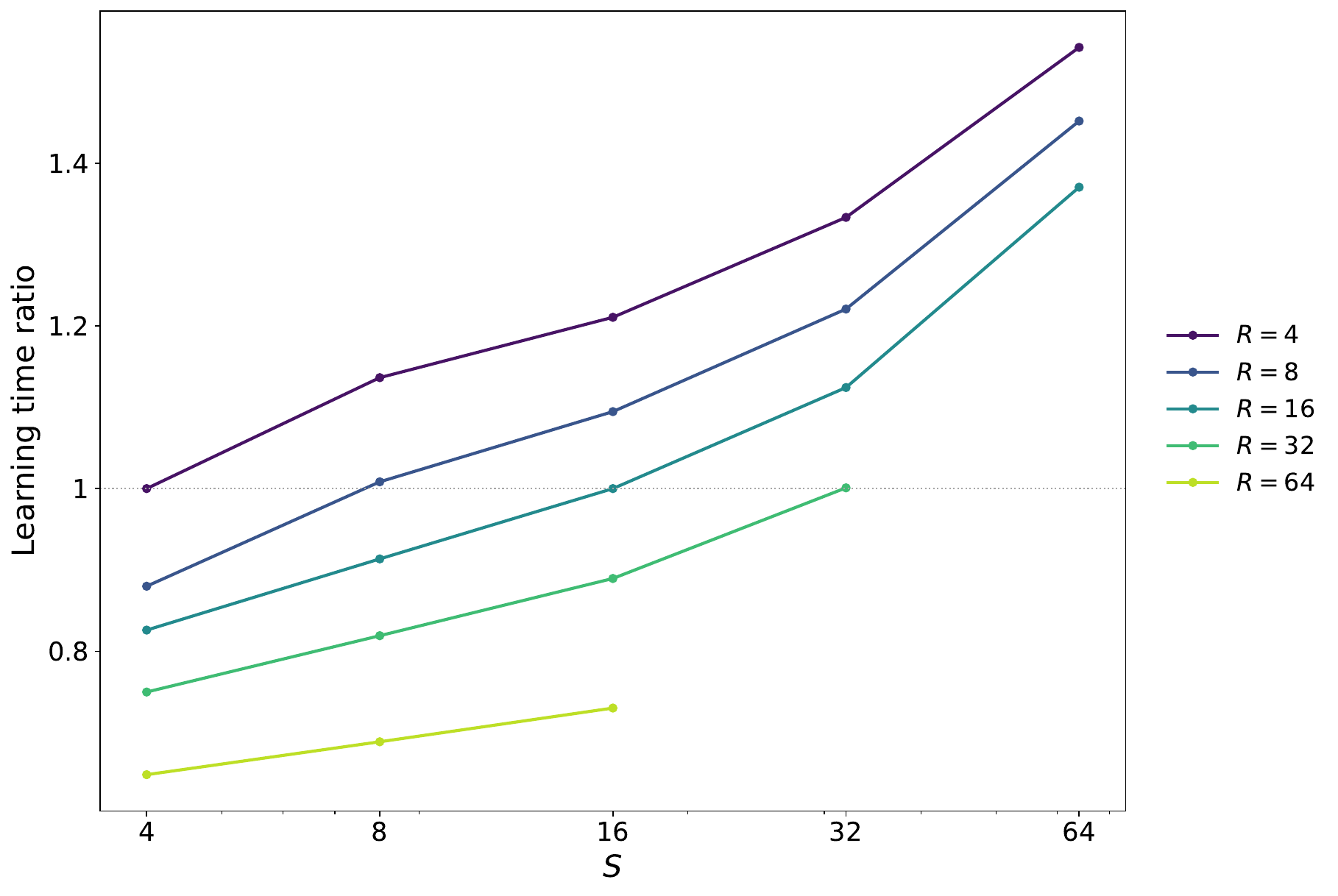}}
\subfigure[AdamW (O, H, H)]{\includegraphics[width=0.32\linewidth]{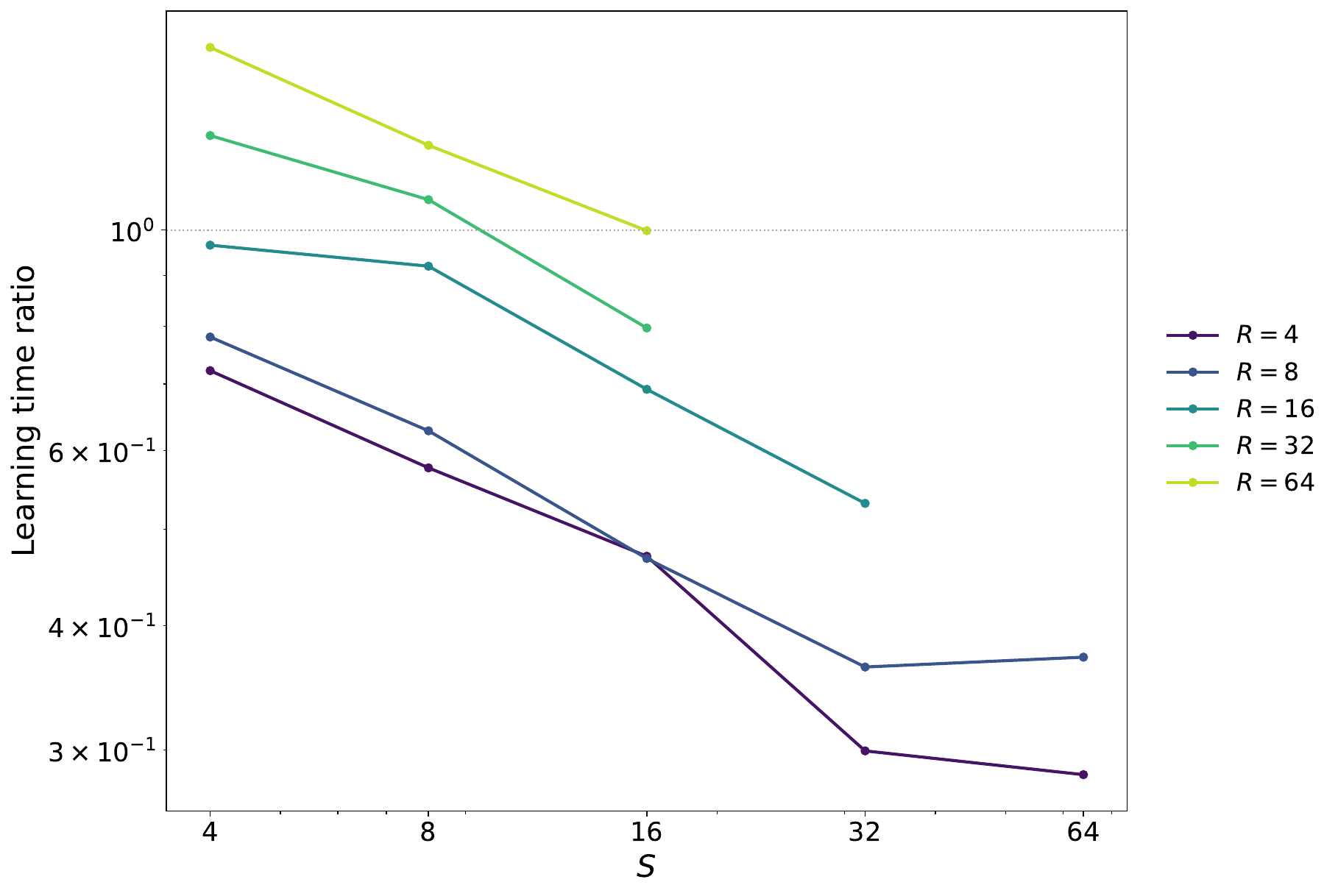}}
\subfigure[AdamW (H, O, H)]{\includegraphics[width=0.32\linewidth]{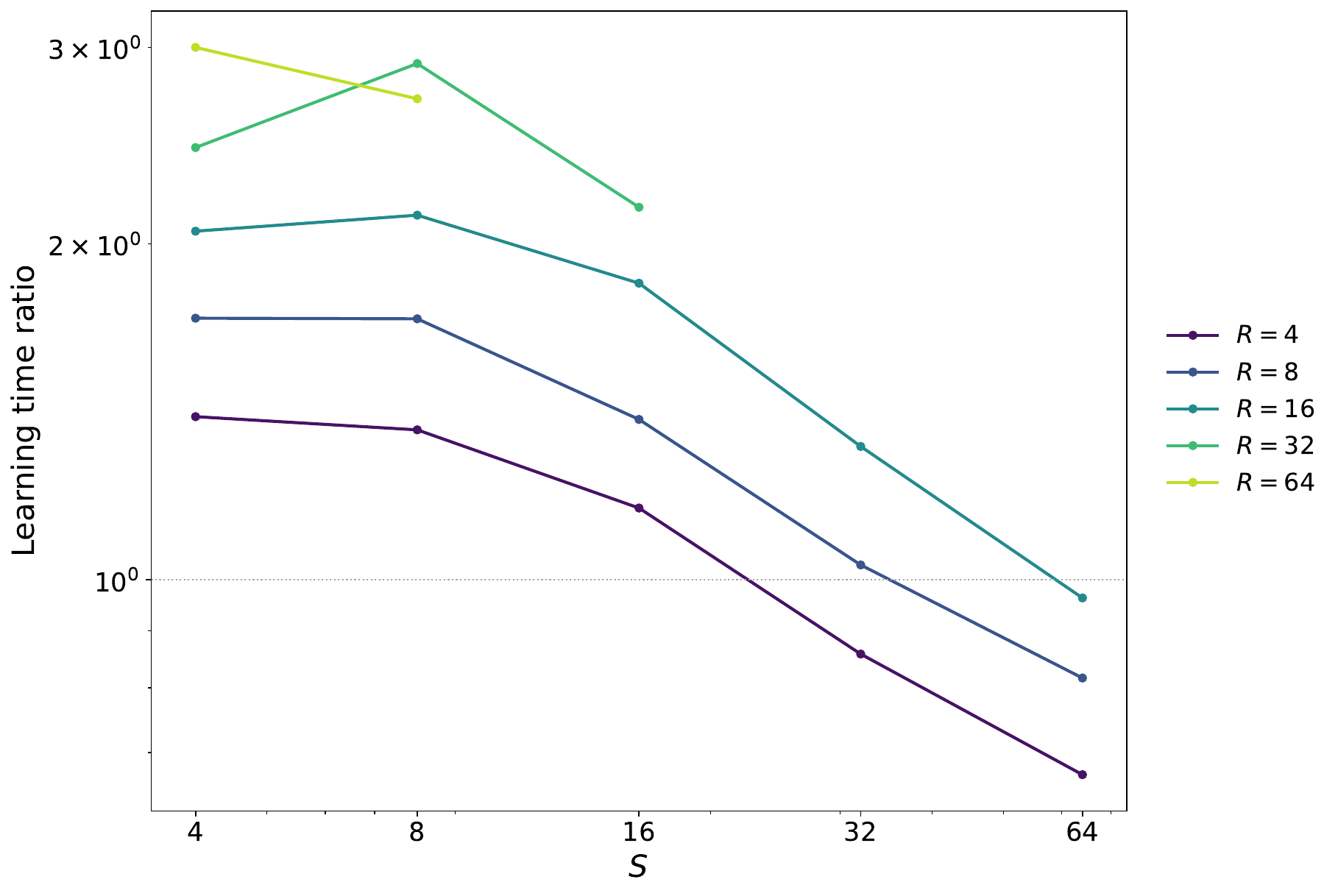}}
\caption{Scaling of learning-time ratio with $S$.}
\label{fig:exp2_S}
\end{figure}

\begin{figure}[H]
\centering
\subfigure[SGD]{\includegraphics[width=0.32\linewidth]{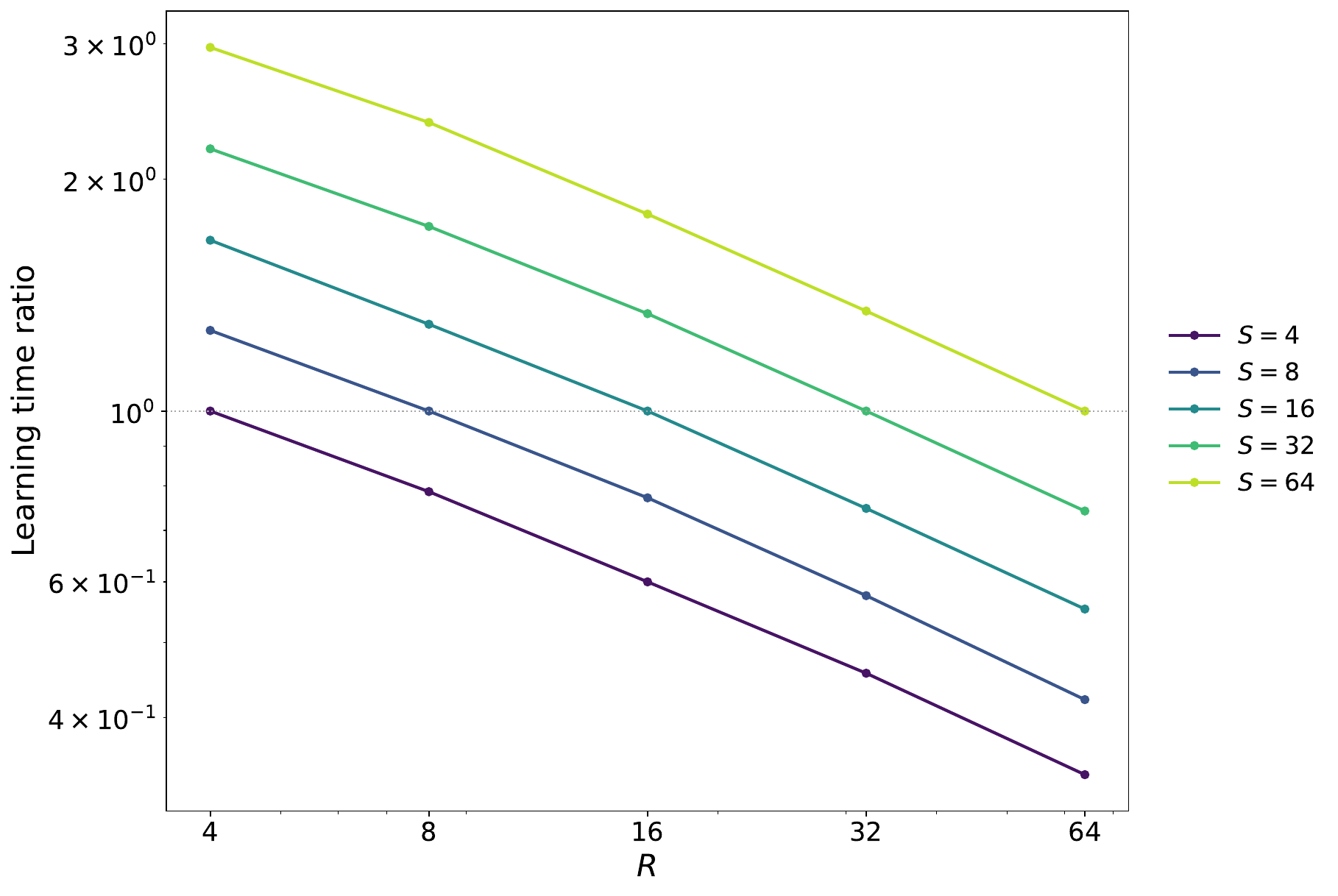}}
\subfigure[Muon]{\includegraphics[width=0.32\linewidth]{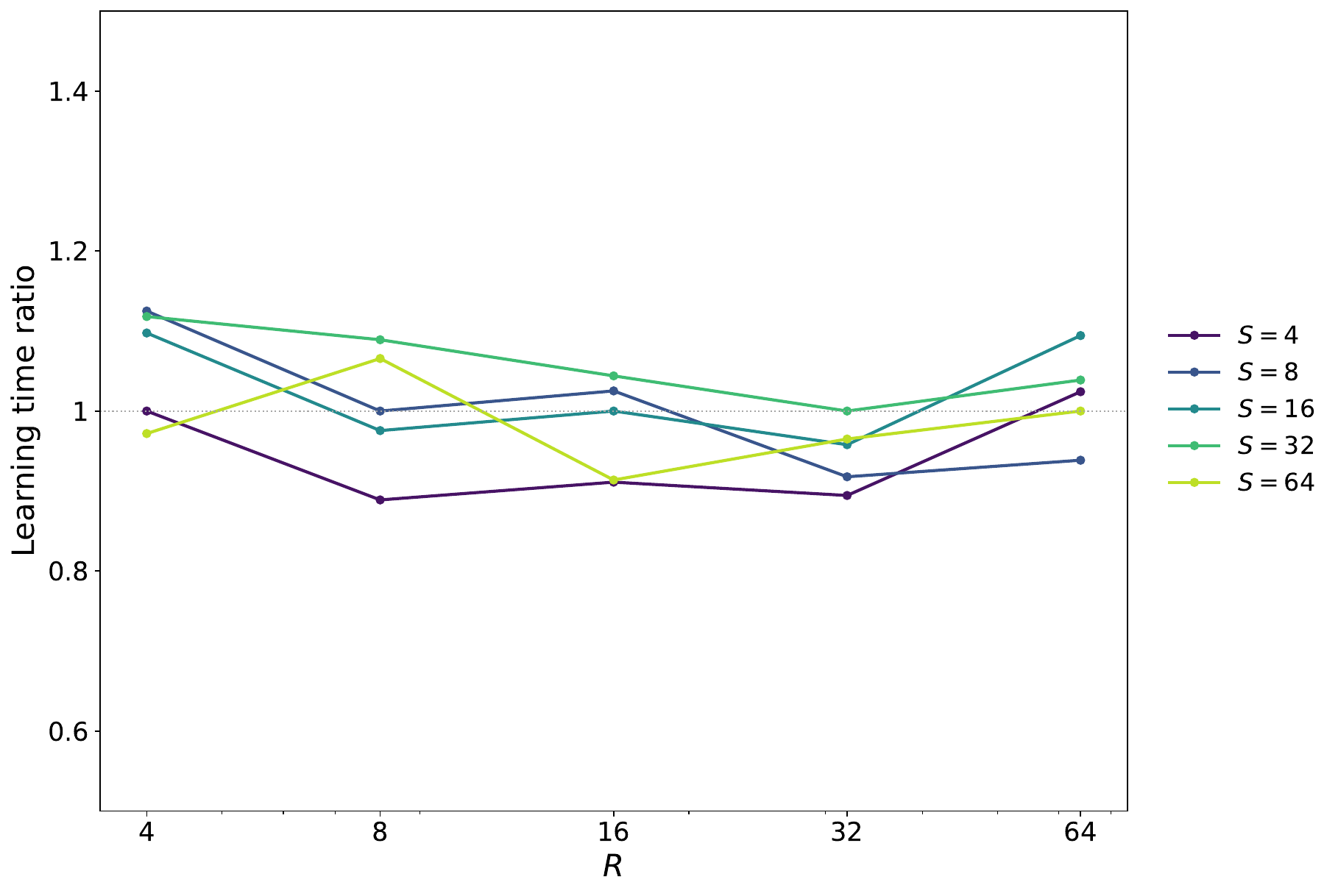}}
\subfigure[AdamW (O, O, O)]{\includegraphics[width=0.32\linewidth]{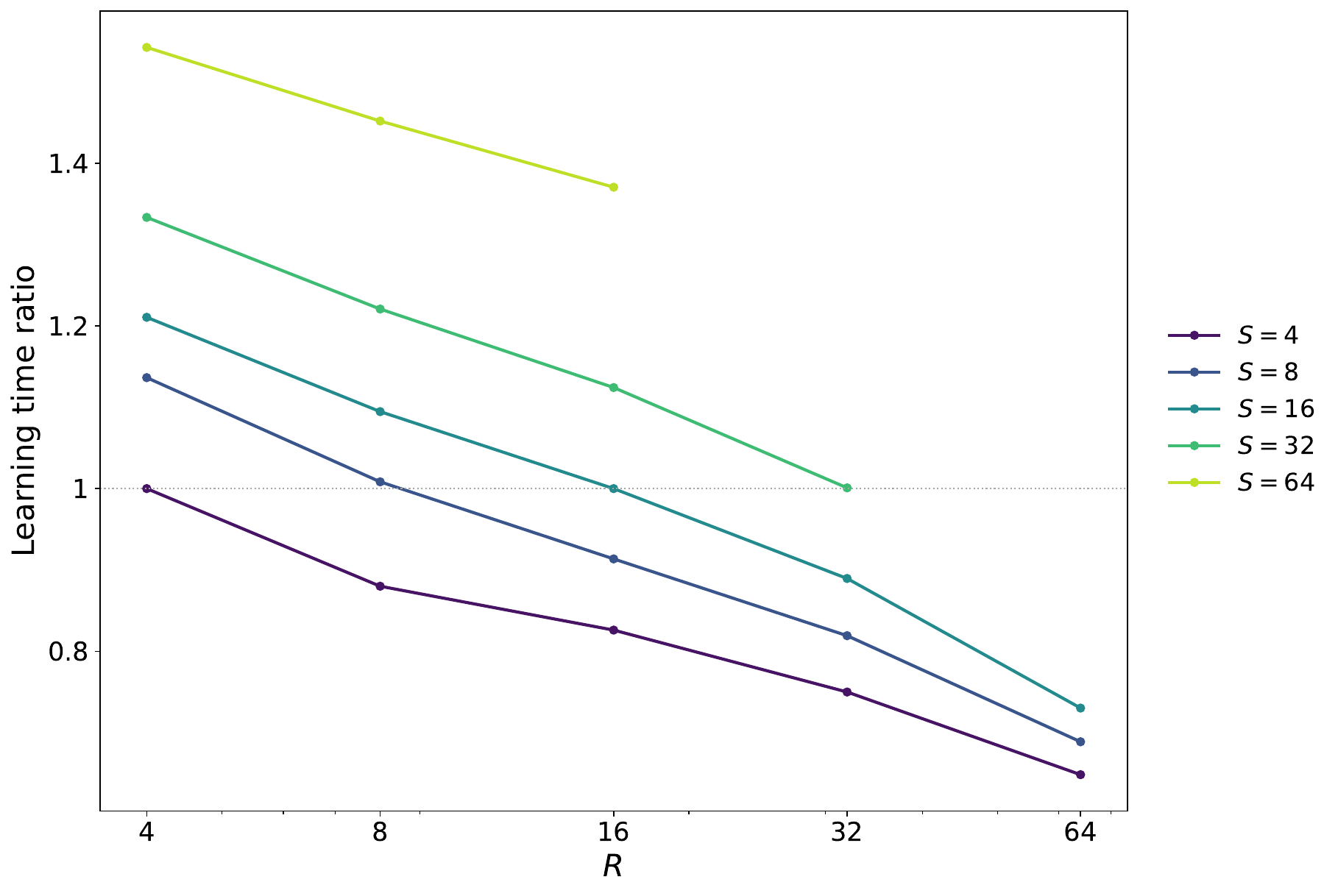}}
\subfigure[AdamW (O, H, H)]{\includegraphics[width=0.32\linewidth]{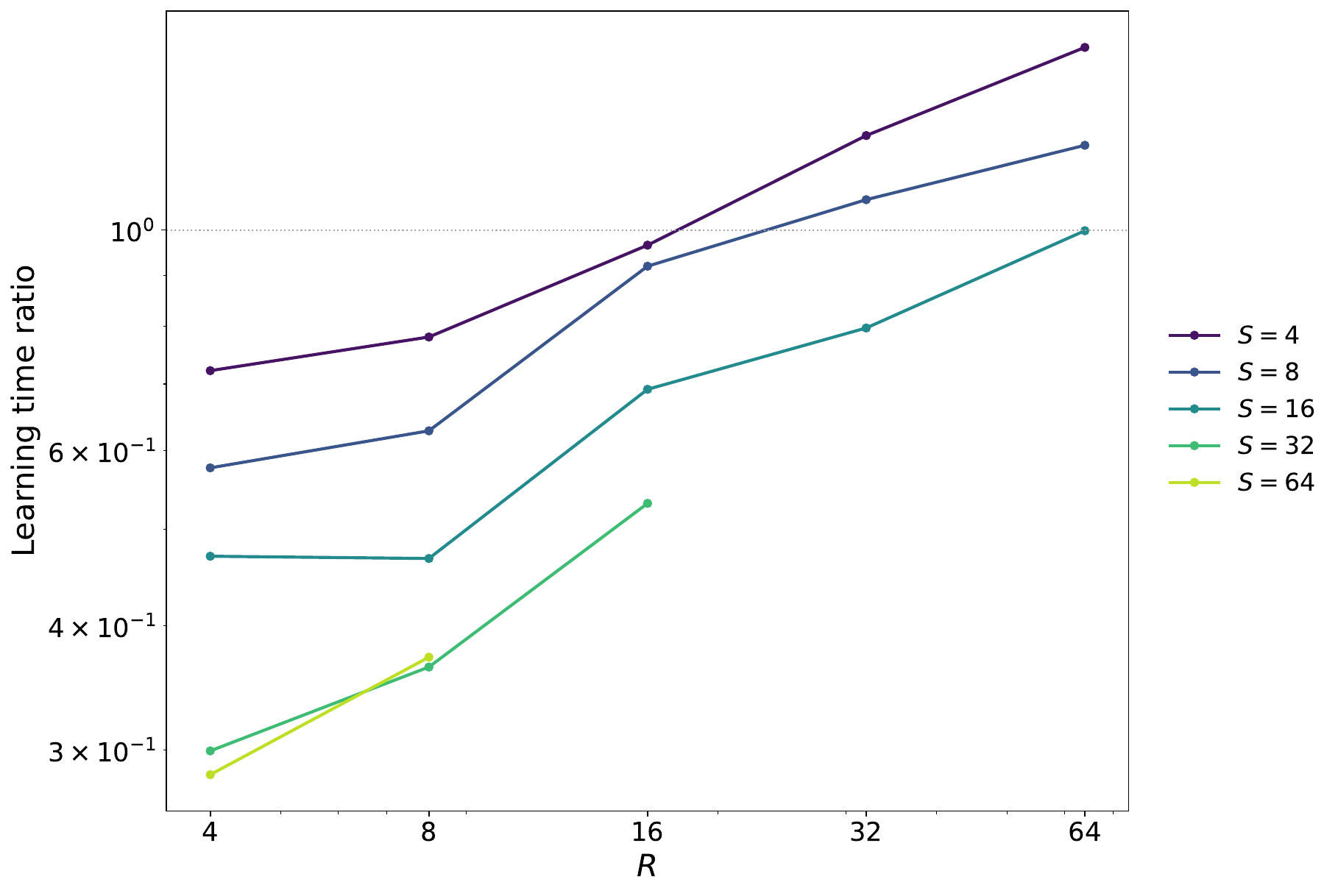}}
\subfigure[AdamW (H, O, H)]{\includegraphics[width=0.32\linewidth]{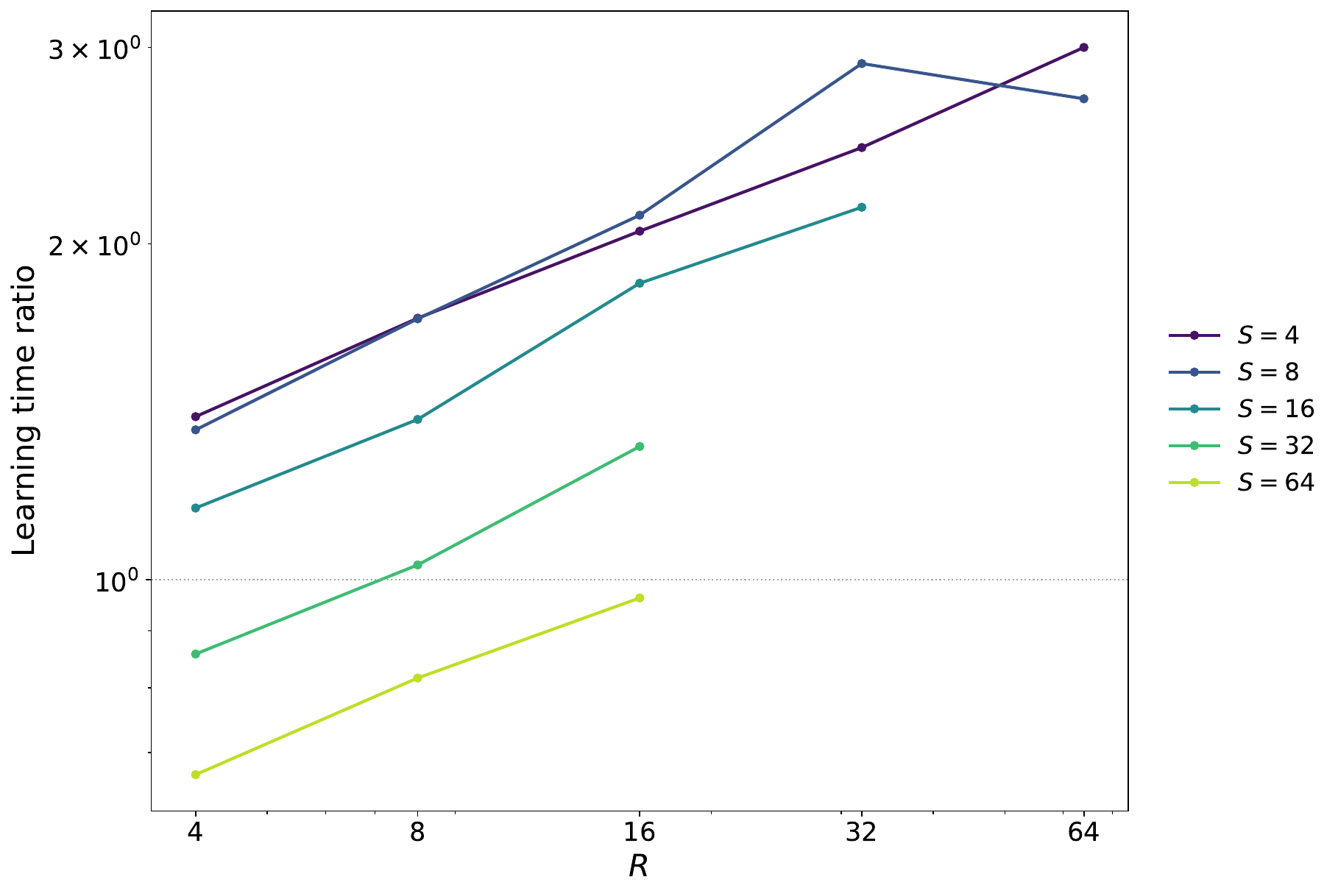}}
\caption{Scaling of learning-time ratio with $R$.}
\label{fig:exp2_R}
\end{figure}

\begin{figure}[H]
\centering
\subfigure[SGD]{\includegraphics[width=0.32\linewidth]{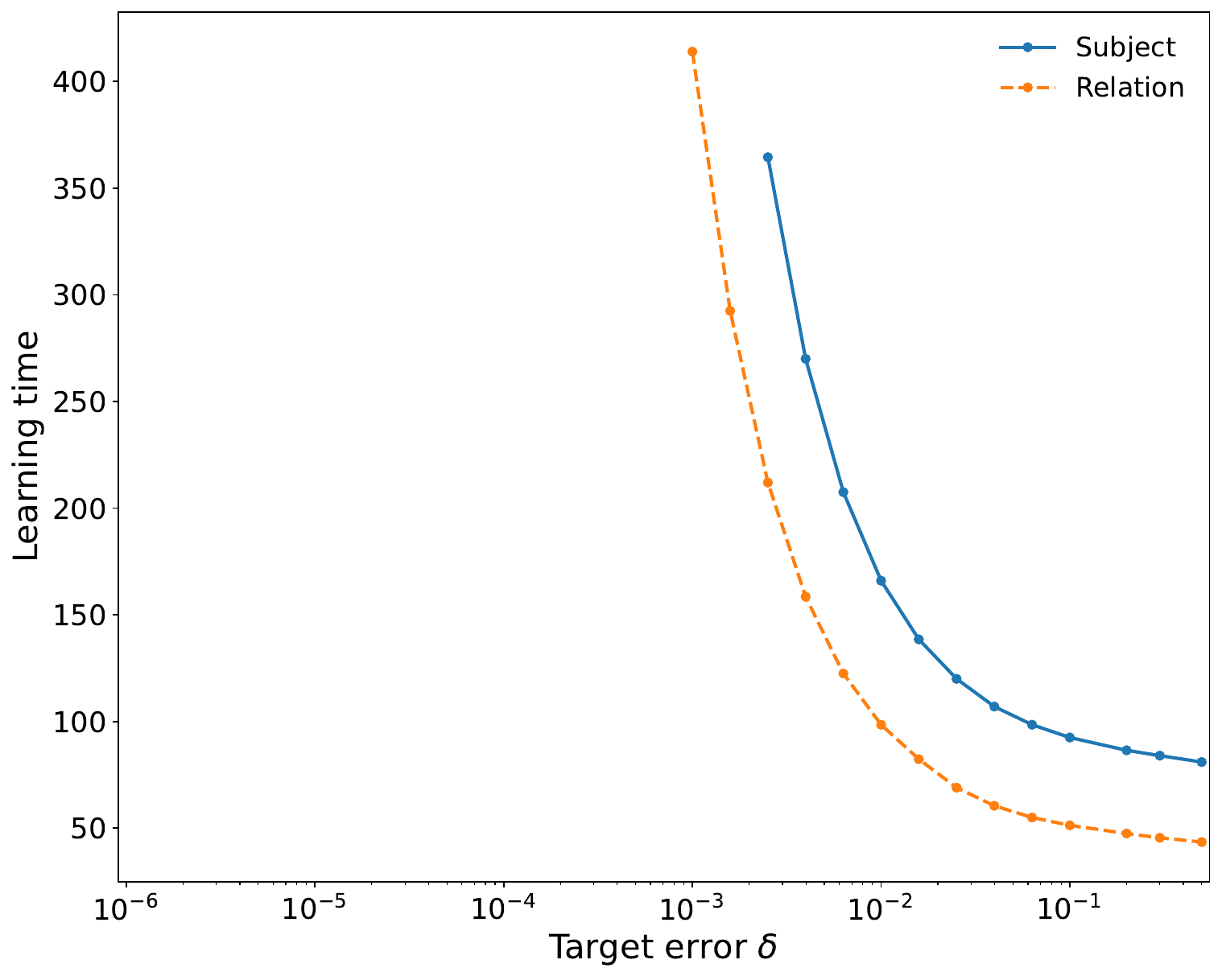}}
\subfigure[Muon]{\includegraphics[width=0.32\linewidth]{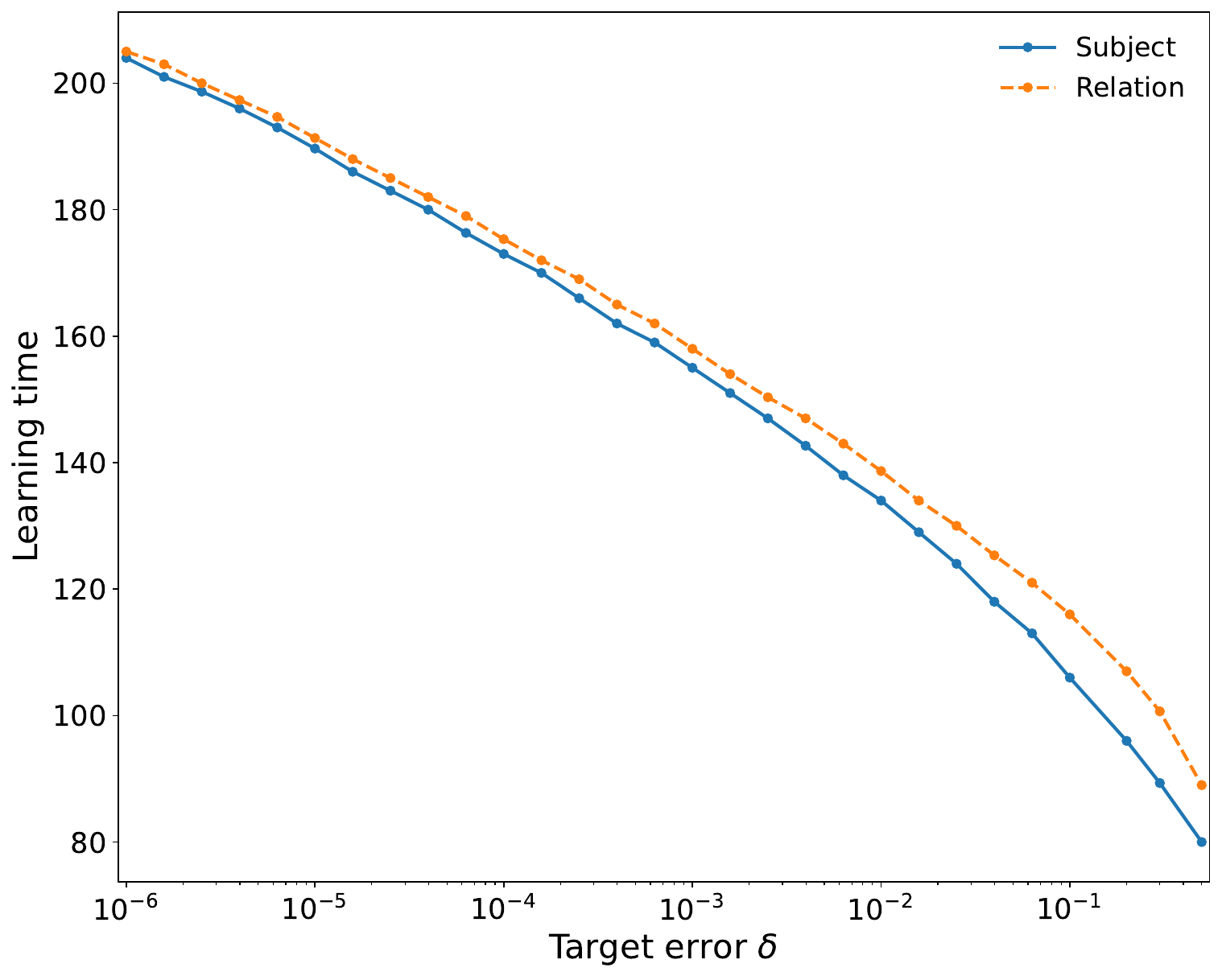}}
\subfigure[AdamW (O, O, O)]{\includegraphics[width=0.32\linewidth]{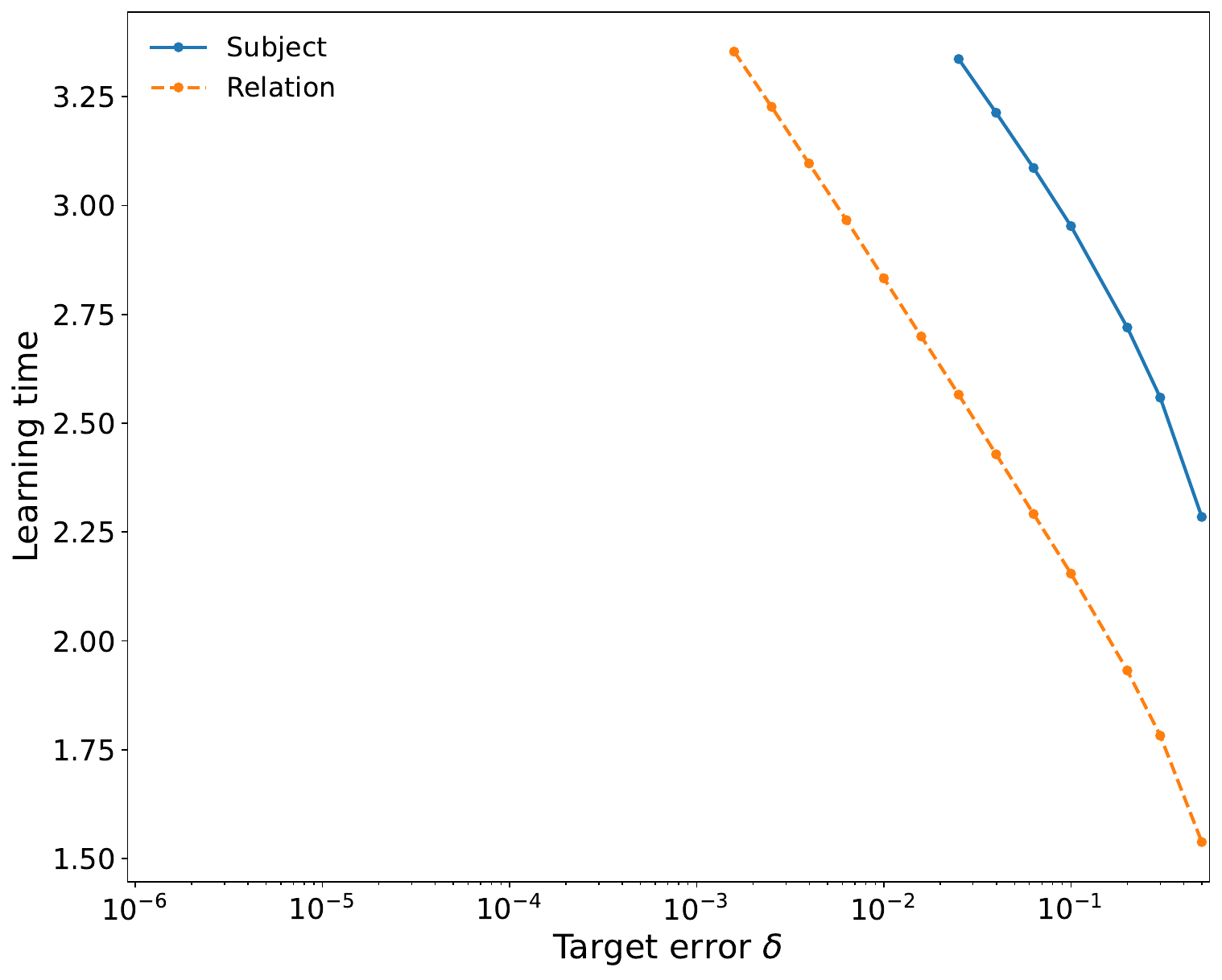}}
\subfigure[AdamW (O, H, H)]{\includegraphics[width=0.32\linewidth]{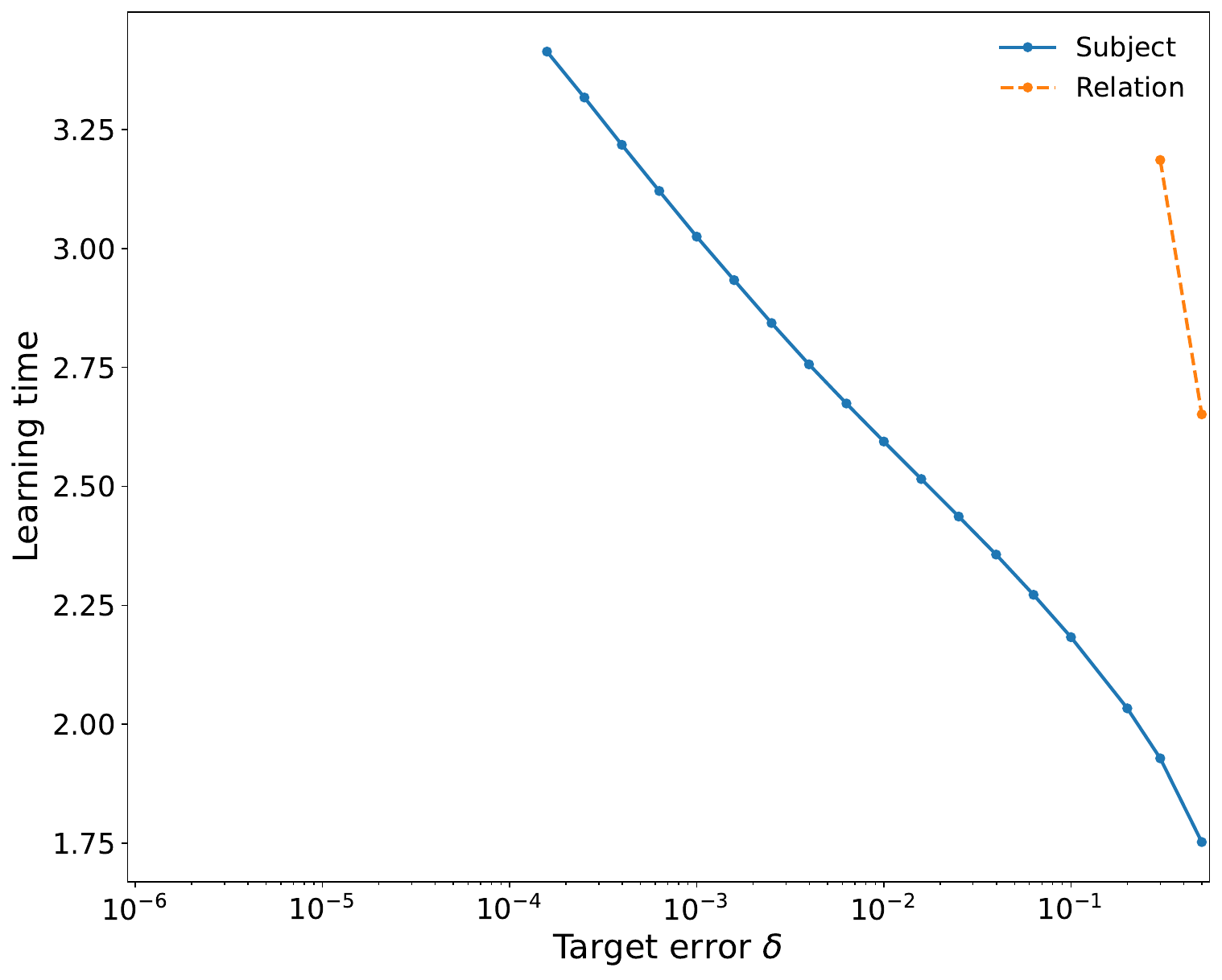}}
\subfigure[AdamW (H, O, H)]{\includegraphics[width=0.32\linewidth]{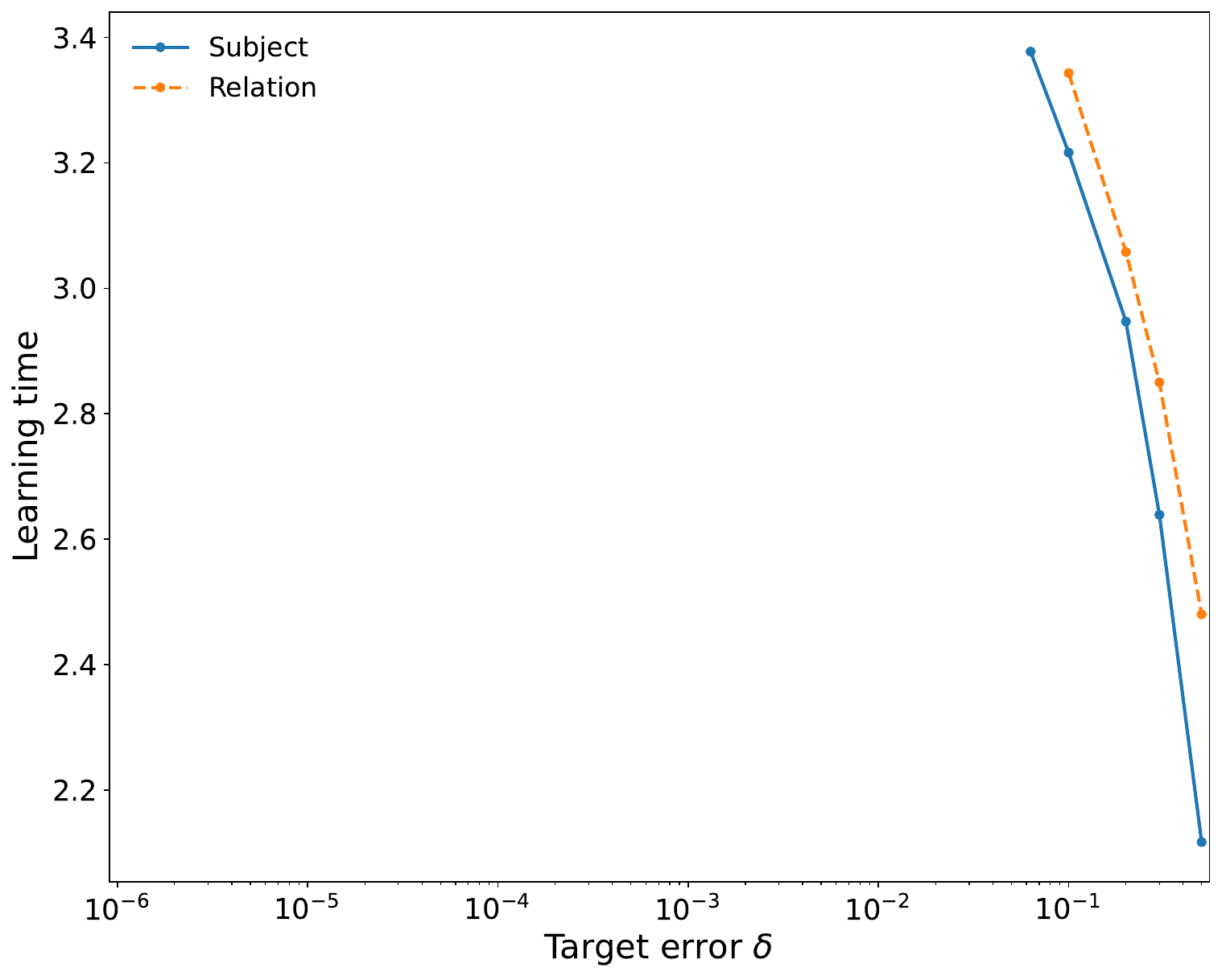}}
\caption{Scaling of learning times with $\delta$.}
\label{fig:exp2_delta}
\end{figure}

We calculate the learning times $T_S^\delta$ and $T_R^\delta$ as the cumulative learning rate. We show the results in Figures~\ref{fig:exp2_S}, \ref{fig:exp2_R}, and \ref{fig:exp2_delta}. We observe that the scaling of the learning-time ratio with $S$ and $R$ is consistent with the ODE simulations and the theoretical predictions.
For scaling of learning times with $\delta$, we can observe the phase transition of the learning times in the SGD dynamics, while the all other optimizers scale logarithmically with $\delta^{-1}$.


\subsection{Realistic Factual Recall in Transformers}

\begin{figure}[ht]
\centering
\includegraphics[width=0.8\linewidth]{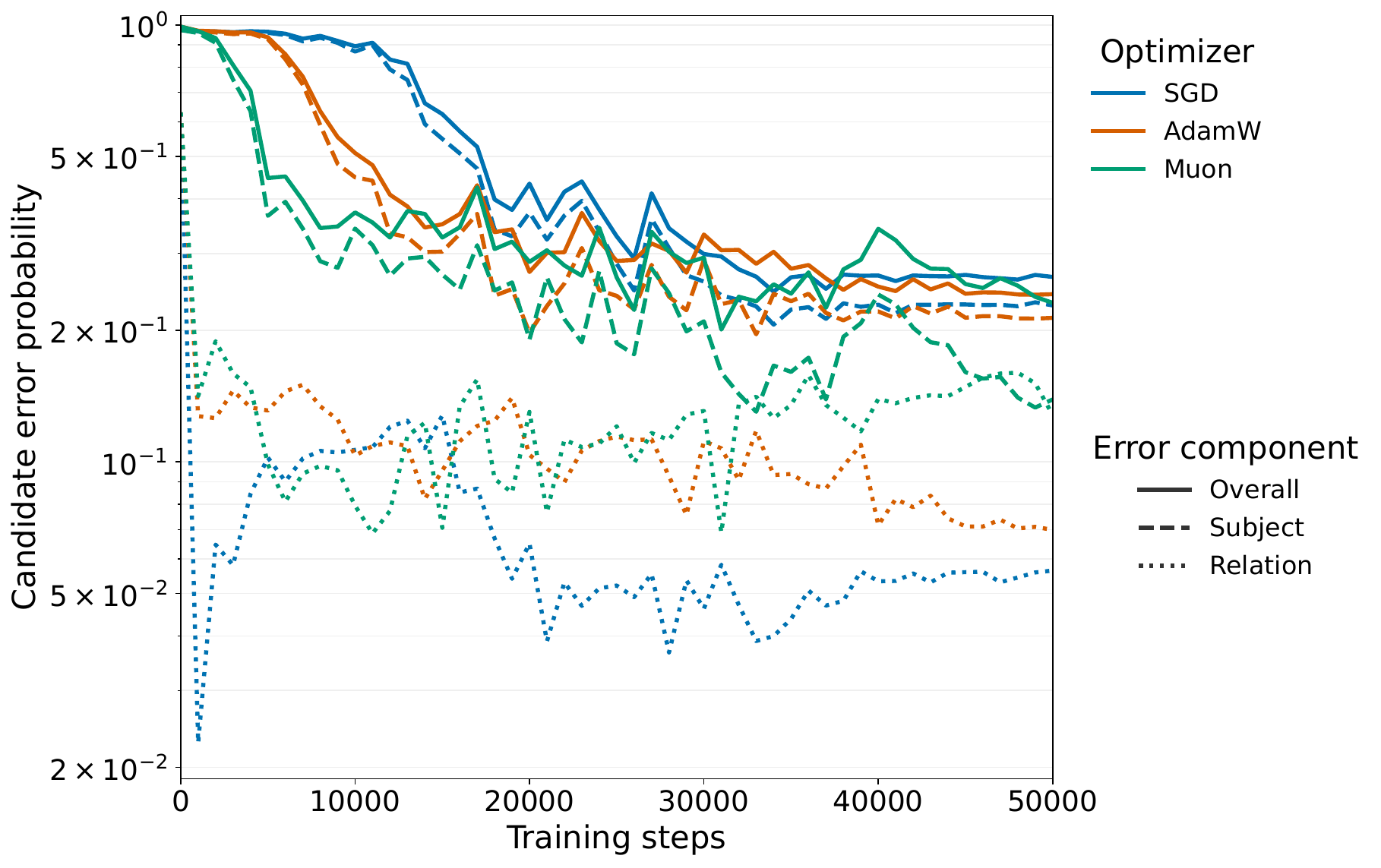}
\caption{SGD, AdamW, and Muon in realistic factual recall task learned by Pythia 35M. The solid line stands for the overall error rate, the dashed line stands for the error probability of the prediction having the wrong subject entity, and the dotted line stands for the error probability of the prediction having the wrong relation entity.}
\label{fig:exp3}
\end{figure}

In this section, we verify the theoretical findings on natural language data learned with a small language model Pythia~\citep{biderman2023pythia} with 35M trainable parameters.

\noindent\textbf{Data model.} Inspired by~\citet{wang2025muon}, we consider a biographical dataset with 8192 fictional people, each with 6 attributes including data of birth, birthplace, university, major, employer, and workplace, so there are 49152 facts in total. The dataset is made up from question-answer pairs, where each attribute corresponds 4 question templates for training and 2 question templates for evaluation. The training and test data are sampled uniformly from the dataset.

\noindent\textbf{Optimizer details.} The optimizer setup is identical to the previous experiment. We apply Muon to each attention head rather than the fused attention weights, which is reported in~\citet{jordan2024muon} to achieve better performance.

The results are shown in Figure~\ref{fig:exp3}. In such a task with much larger number of subjects (8192 people) than relations (6 attributes), the errors of the relation component are much smaller than those of the subject component for all three optimizer. The overall error is thus close to the error of the subject component. We observe that the subject error of SGD does not significantly decay until step 10000, while the subject error decays much earlier in the Muon training dynamics. The performance of AdamW is between SGD and Adam. Although all three optimizers reach similar accuracies after step 30000, Muon has the advantage of learning the subject-dependent component faster than SGD and Adam, indicating that Muon can save the training budget spent on training the subject component.

\section{Conclusion}

In this work, we study the GF, Spectral GF, and Sign GF dynamics in the factual recall task reduced to the invariant manifold, where the model prediction factors into subject and relation components.
We show that Spectral GF, the continuous-time limit of the Muon optimizer, learns the two components at a more balanced rate, thus weakening phase separation of GF.
Additionally, the spectral normalization of Muon and Spectral GF avoids the $\delta^{-1}$ dependence in the learning times caused by softmax saturation.
The Sign GF dynamics, a continuous-time model of Adam, can have diverse behavior due to its dependence on the token embedding.
Our analysis is currently restricted to an ideal setting with a simple data model, the single-head linear attention with merged K-Q and O-V weights in an invariant manifold, access to full gradient, and continuous-time dynamics.
We leave as future work to generalize these assumptions.

\appendix

\section{Discussion about Learning Rate Schedule}\label{section:LR}

The learning rate schedule can be incorporated into the training dynamics through time reparameterization. For a general learning dynamics with learning rate schedule $\gamma(t)$ given by
\begin{align*}
\dot\bTheta(t)=-\gamma(t)\cdot F(\bTheta(t)),
\end{align*}
where $F(\cdot)$ is a mapping from the parameter space to the (normalized) gradient, consider the time reparameterization
\begin{align*}
\tau=g(t):=\int_0^t\gamma(s)\ud s,\quad\tilde\bTheta=\bTheta\circ g^{-1}.
\end{align*}
Then
\begin{align*}
\ud\tilde\bTheta(\tau)/\ud\tau=-F(\tilde\bTheta(\tau)).
\end{align*}
Therefore, under the time reparameterization, the dynamics with an arbitrary learning rate schedule can be transformed into the dynamics with unit learning rate.

Therefore, the relative separation $T_S^\delta/T_R^\delta$ can be viewed as the ratio of the cumulative learning rates for $q_S$ and $q_R$ to achieve the threshold $\log(\delta^{-1}-1)$, and is more intrinsic to the optimization dynamics than either learning time alone.

\section{Proof of Structural Results}\label{section:proof_dynamics}

\subsection{Proof of Proposition~\ref{prop:rotation_invariance}}

\begin{proof}[Proof of Proposition~\ref{prop:rotation_invariance}]
Let $\tilde\Xb$ denote the embedding matrix under the embedding $\tilde\bphi(\cdot)$, then
\begin{align*}
\tilde\Xb=[\Qb\bphi(z_1), \dots, \Qb\bphi(T)]^\top=\Xb\Qb^\top.
\end{align*}
We first notice that for any $a$, the logit satisfies
\begin{align*}
\langle\tilde\bphi(a), \fb(\tilde\Xb; \tilde\bTheta)\rangle&=(\Qb\bphi(a))^\top(\Qb\Wb_{OV}\Qb^\top)\cdot(\Xb\Qb^\top)^\top(\Xb\Qb^\top)\cdot(\Qb\Wb_{KQ}\Qb^\top)\cdot(\Qb\bphi(\EOS))\\
&=\bphi(a)^\top\Wb_{OV}\Xb^\top\Xb\Wb_{KQ}\bphi(\EOS)\\
&=\langle\bphi(a), \fb(\Xb; \bTheta)\rangle.
\end{align*}
where the first equality holds because $\Ub^\top\Ub=\Ib$. Therefore, the logits are invariant under the transformation, so
\begin{align*}
\hat p_{\bTheta}(a|z_{1:T+1})=\hat p_{\tilde\bTheta}(a|z_{1:T+1}).
\end{align*}

We then calculate the gradient of $\ell(z_{1:T+1}; \bTheta)$. The gradient w.r.t. $\Wb_{OV}$ satisfies
\begin{align}\label{eq:grad_ell_WOV}
-\nabla_{\Wb_{OV}}\ell(z_{1:T+1}; \bTheta)=\bigg\{\sum_{a\in\cA}\hat p_{\bTheta}(a|z_{1:T})[\bphi(z_{T+1})-\bphi(a)]\bigg\}\bphi(\EOS)^\top\Wb_{KQ}^\top\Xb^\top\Xb.
\end{align}
The gradient w.r.t. $\Wb_{KQ}$ satisfies
\begin{align}\label{eq:grad_ell_WKQ}
-\nabla_{\Wb_{KQ}}\ell(z_{1:T+1}; \bTheta)=\Xb^\top\Xb\Wb_{OV}^\top\bigg\{\sum_{a\in\cA}\hat p_{\bTheta}(a|z_{1:T})[\bphi(z_{T+1})-\bphi(a)]\bigg\}\bphi(\EOS)^\top.
\end{align}
Under the parameter transform, we notice that
\begin{align*}
&\tilde\bphi(a)\tilde\bphi(\EOS)^\top\tilde\Wb_{KQ}^\top\tilde\Xb^\top\tilde\Xb\\
&=(\Qb\bphi(a))\cdot(\Qb\bphi(\EOS))^\top\cdot(\Qb\Wb_{KQ}\Qb^\top)^\top\cdot(\Xb\Qb^\top)^\top\cdot(\Xb\Qb^\top)\\
&=\Qb\cdot(\bphi(a)\bphi(\EOS)\Wb_{KQ}^\top\Xb^\top\Xb)\cdot\Qb^\top,
\end{align*}
and
\begin{align*}
&\tilde\Xb^\top\tilde\Xb\tilde\Wb_{OV}^\top\tilde\bphi(a)\tilde\bphi(\EOS)^\top\\
&=(\Xb\Qb^\top)^\top\cdot(\Xb\Qb^\top)\cdot(\Qb\Wb_{OV}\Qb^\top)^\top\cdot(\Qb\bphi(a))\cdot(\Qb\bphi(\EOS))^\top\\
&=\Qb\cdot(\Xb^\top\Xb\Wb_{OV}^\top\bphi(a)\bphi(\EOS)^\top)\cdot\Qb^\top.
\end{align*}
Combining the properties above with the fact that $\hat p_{\bTheta}=\hat p_{\tilde\bTheta}$, we have
\begin{align*}
-\nabla_{\tilde\Wb_{OV}}\ell(z_{1:T+1}; \tilde\bTheta)&=\Ub\cdot[-\nabla_{\Wb_{OV}}\ell(z_{1:T+1}; \bTheta)]\cdot\Ub^\top,\\
-\nabla_{\tilde\Wb_{KQ}}\ell(z_{1:T+1}; \tilde\bTheta)&=\Ub\cdot[-\nabla_{\Wb_{KQ}}\ell(z_{1:T+1}; \bTheta)]\cdot\Ub^\top.
\end{align*}
Taking expectations of the sequence $\EE_{z_{1:T+1}}$, we have
\begin{align*}
-\nabla_{\tilde\Wb_{OV}}L(\tilde\bTheta)=\Qb\cdot[-\nabla_{\Wb_{OV}}L(\bTheta)]\cdot\Qb^\top,\quad-\nabla_{\tilde\Wb_{KQ}}L(\tilde\bTheta)=\Qb\cdot[-\nabla_{\Wb_{KQ}}L(\bTheta)]\cdot\Qb^\top.
\end{align*}
Therefore, under the GF dynamics, we have
\begin{align*}
\frac{\ud\tilde\Wb_{OV}(t)}{\ud t}&=\Qb\cdot\frac{\ud\Wb_{OV}(t)}{\ud t}\cdot\Qb^\top\\
&=\Qb\cdot[-\gammaGF\cdot\nabla_{\Wb_{OV}}L(\bTheta)]\cdot\Qb^\top\\
&=-\gammaGF\cdot\nabla_{\tilde\Wb_{OV}}L(\tilde\bTheta).
\end{align*}
Similarly, we have
\begin{align*}
\frac{\ud\tilde\Wb_{KQ}(t)}{\ud t}=-\gammaGF\cdot\nabla_{\tilde\Wb_{KQ}}L(\tilde\bTheta).
\end{align*}
Therefore, $\tilde\bTheta$ follows the GF dynamics. Under the Spectral GF dynamics, we have
\begin{align*}
\frac{\ud\tilde\Wb_{OV}(t)}{\ud t}&=\Qb\cdot\frac{\ud\Wb_{OV}(t)}{\ud t}\cdot\Qb^\top\\
&=\Qb\cdot[-\gammaSpecGF\cdot\msign(\nabla_{\Wb_{OV}}(L(\bTheta)))]\cdot\Qb^\top\\
&=-\gammaSpecGF\cdot\msign(\Qb\cdot\nabla_{\Wb_{OV}}(L(\bTheta))\cdot\Qb^\top)\\
&=-\gammaSpecGF\cdot\msign(\nabla_{\Wb_{OV}}L(\tilde\bTheta)),
\end{align*}
where the third equality holds by Lemma \ref{lemma:msign_rotation}. Similarly,
\begin{align*}
\frac{\ud\tilde\Wb_{KQ}(t)}{\ud t}=-\gammaSpecGF\cdot\msign(\nabla_{\Wb_{kQ}}L(\tilde\bTheta)).
\end{align*}
Therefore, $\tilde\bTheta$ follows the Spectral GF dynamics.
\end{proof}

\subsection{Proof of Proposition~\ref{prop:prediction} and Proposition~\ref{prop:GF}}

\begin{proof}[Proof of Proposition~\ref{prop:prediction} and Proposition~\ref{prop:GF}]
We use $s(z_{1:T})$ and $r(z_{1:T})$ to denote the subject and relation tokens of the sequence $z_{1:T}$, respectively.
Let $C(z, z_{1:T})$ be the number of occurrences of the token $z$ in the sequence $z_{1:T}$. Then
\begin{align}\label{eq:XTX}
\Xb^\top\Xb=\bphi(s(z_{1:T}))\bphi(s(z_{1:T}))^\top+\bphi(r(z_{1:T}))\bphi(r(z_{1:T}))^\top+\sum_{n\in\cN}C(n, z_{1:T})\bphi(n)\bphi(n)^\top.
\end{align}
Recall the definition of the manifold:
\begin{align*}
\Wb_{OV}&=\sum_{s\in\cS}(\alpha_S\bar\vb_s+\gamma_S\bar\vb)\bphi(s)^\top+\sum_{r\in\cR}(\alpha_R\bar\vb_r+\gamma_R\bar\vb)\bphi(r)^\top+\sum_{n\in\cN}\gamma_N\bar\vb\bphi(n)^\top,\\
\Wb_{KQ}&=\bigg(\beta_S\sum_{s\in\cS}\bphi(s)+\beta_R\sum_{r\in\cR}\bphi(r)+\beta_N\sum_{n\in\cN}\bphi(n)\bigg)\bphi(\EOS)^\top.
\end{align*}
We then calculate several quantities important for subsequent derivations. We have
\begin{align*}
\Wb_{OV}\Xb^\top\Xb&=(\alpha_S\bar\vb_{s(z_{1:T})}+\gamma_S\bar\vb)\bphi(s(z_{1:T}))^\top+(\alpha_R\bar\vb_{r(z_{1:T})}+\gamma_R\bar\vb)\bphi(r(z_{1:T}))^\top\\
&\quad+\sum_{n\in\cN}\gamma_N\bar\vb\cdot C(n, z_{1:T})\bphi(n)^\top,
\end{align*}
and
\begin{align*}
\Xb\Xb^\top\Wb_{KQ}\bphi(\EOS)&=\beta_S\bphi(s(z_{1:T}))+\beta_R\bphi(r(z_{1:T}))+\beta_N\sum_{n\in\cN}C(n, z_{1:T})\bphi(n).
\end{align*}
Finally,
\begin{align*}
\fb(\Xb; \bTheta)&=\Wb_{OV}\Xb^\top\Xb\Wb_{KQ}\bphi(\EOS)\\
&=\beta_S(\alpha_S\bar\vb_{s(z_{1:T})}+\gamma_S\bar\vb)+\beta_R(\alpha_R\bar\vb_{r(z_{1:T})}+\gamma_R\bar\vb)+\beta_N\sum_{n\in\cN}C(n, z_{1:T})\cdot\gamma_N\bar\vb\\
&=\beta_S(\alpha_S\bar\vb_{s(z_{1:T})}+\gamma_S\bar\vb)+\beta_R(\alpha_R\bar\vb_{r(z_{1:T})}+\gamma_R\bar\vb)+\beta_N\gamma_N(T-3)\bar\vb,
\end{align*}
where the last equality holds because there are exactly $T-3$ noise tokens in any sequence $z_{1:T}$.

\noindent\textbf{Calculation of $\hat p_{\bTheta}(a|z_{1:T})$.}
For any $(s', r')$, the logits corresponding to $a_*(s', r')$ is
\begin{align*}
\bphi(a_*(s', r'))^\top\fb(\Xb; \bTheta)&=\alpha_S\beta_S\cdot\bar\vb_{s(z_{1:T})}^\top\bphi(a_*(s', r'))+\alpha_R\beta_R\cdot\bar\vb_{r(z_{1:T})}^\top\bphi(a_*(s', r'))\\
&\quad+\underbrace{(\beta_S\gamma_S+\beta_R\gamma_R+(T-3)\beta_N\gamma_N)\bar\vb^\top\bphi(a_*(s', r'))}_{:=C}\\
&=\frac{\alpha_S\beta_S}{\sqrt R}\cdot\delta_{s', s(z_{1:T})}+\frac{\alpha_R\beta_R}{\sqrt S}\cdot\delta_{r', r(z_{1:T})}+C,
\end{align*}
where the last equality holds due to Lemma~\ref{lemma:barv_properties}(a). Since $C$ is a constant for all $(s', r')$, we omit $C$ in subsequent calculations. The normalizing constant is
\begin{align*}
Z&=\sum_{s'\in\cS}\sum_{r'\in\cR}\exp(\alpha_S\beta_S/\sqrt R\cdot\delta_{s', s(z_{1:T})}+\alpha_R\beta_R/\sqrt S\cdot\delta_{r', r(z_{1:T})})\\
&=\bigg[\sum_{s'\in\cS}\exp(\alpha_S\beta_S/\sqrt R\cdot\delta_{s', s(z_{1:T})})\bigg]\cdot\bigg[\sum_{r\in\cR}\exp(\alpha_R\beta_R/\sqrt S\cdot\delta_{r', r(z_{1:T})})\bigg]\\
&=(S-1+e^{\alpha_S\beta_S/\sqrt R})(R-1+e^{\alpha_R\beta_R/\sqrt S}).
\end{align*}
Therefore, the model output satisfies
\begin{align*}
p(a_*(s', r')|z_{1:T})=\frac{\exp(\bphi(a_*(s', r'))^\top\fb(\Xb; \bTheta))}{Z}=\frac{e^{\alpha_S\beta_S/\sqrt R\cdot\delta_{s', s(z_{1:T})}}}{S-1+e^{\alpha_S\beta_S/\sqrt R}}\cdot\frac{e^{\alpha_R\beta_R/\sqrt S\cdot\delta_{r', r(z_{1:T})}}}{R-1+e^{\alpha_R\beta_R/\sqrt S}}.
\end{align*}
\noindent\textbf{Calculation of $-\nabla_{\Wb_{OV}}L(\bTheta)$.} We first calculate the stochastic gradients $-\nabla_{\Wb_{OV}}\ell(z_{1:T+1}; \bTheta)$ as given in \eqref{eq:grad_ell_WOV}. We firstly have
\begin{align*}
&\sum_{a\in\cA}\hat p_{\bTheta}(a|z_{1:T})[\bphi(z_{T+1})-\bphi(a)]=\sum_{s'\in\cS}\sum_{r'\in\cR}\hat p_{\bTheta}(a_*(s', r')|z_{1:T})[\bphi(z_{T+1})-\bphi(a_*(s', r'))]\\
&=\frac{1}{S-1+e^{\alpha_S\beta_S/\sqrt R}}\cdot\frac{1}{R-1+e^{\alpha_R\beta_R/\sqrt S}}\cdot\sum_{s'\in\cS}\sum_{r'\in\cR}[\bphi(z_{T+1})-\bphi(a_*(s', r'))]\\
&\quad+\sum_{s'\in\cS}\sum_{r'\in\cR}\bigg(\hat p_{\bTheta}(a_*(s', r')|z_{1:T})-\frac{1}{S-1+e^{\alpha_S\beta_S/\sqrt R}}\cdot\frac{1}{R-1+e^{\alpha_R\beta_R/\sqrt S}}\bigg)\\
&\quad\cdot[\bphi(z_{T+1})-\bphi(a_*(s', r'))]\\
&=\frac{1}{S-1+e^{\alpha_S\beta_S/\sqrt R}}\cdot\frac{1}{R-1+e^{\alpha_R\beta_R/\sqrt S}}\cdot\Big\{(SR\cdot\bphi(z_{T+1})-\sqrt{SR}\cdot\bar\vb)\\
&\quad+(e^{\alpha_R\beta_R/\sqrt S}-1)\sum_{s'}[\bphi(z_{T+1})-\bphi(a_*(s', r(z_{1:T})))]\\
&\quad+(e^{\alpha_S\beta_S/\sqrt R}-1)\sum_{r'}[\bphi(z_{T+1})-\bphi(a_*(s(z_{1:T}), r'))]\Big\}\\
&=\frac{1}{S-1+e^{\alpha_S\beta_S/\sqrt R}}\cdot\frac{1}{R-1+e^{\alpha_R\beta_R/\sqrt S}}\cdot[(SR\cdot\bphi(z_{T+1})-\sqrt{SR}\cdot\bar\vb)\\
&\quad+(e^{\alpha_R\beta_R/\sqrt S}-1)(S\cdot\bphi(z_{T+1})-\sqrt S\cdot\bar\vb_{r(z_{1:T})})\\
&\quad+(e^{\alpha_S\beta_S/\sqrt R}-1)(R\cdot\bphi(z_{T+1})-\sqrt R\cdot\bar\vb_{s(z_{1:T})})]
\end{align*}
Therefore, the stochastic gradient w.r.t. $\Wb_{OV}$ satisfies
\begin{align}
&-\nabla  _{\Wb_{OV}}\ell(z_{1:T+1}; \bTheta)=\bigg\{\sum_{a\in\cA}\hat p_{\bTheta}(a|z_{1:T})[\bphi(z_{T+1})-\bphi(a)]\bigg\}\bphi(\EOS)^\top\Wb_{KQ}^\top\Xb^\top\Xb\nonumber\\
&=\frac{1}{S-1+e^{\alpha_S\beta_S/\sqrt R}}\cdot\frac{1}{R-1+e^{\alpha_R\beta_R/\sqrt S}}\cdot[(SR\cdot\bphi(z_{T+1})-\sqrt{SR}\cdot\bar\vb)\nonumber\\
&\quad+(e^{\alpha_R\beta_R/\sqrt S}-1)(S\cdot\bphi(z_{T+1})-\sqrt S\cdot\bar\vb_{r(z_{1:T})})\nonumber\\
&\quad+(e^{\alpha_S\beta_S/\sqrt R}-1)(R\cdot\bphi(z_{T+1})-\sqrt R\cdot\bar\vb_{s(z_{1:T})})]\nonumber\\
&\quad\cdot\bigg[\beta_S\bphi(s(z_{1:T}))+\beta_R\bphi(r(z_{1:T}))+\beta_N\sum_{n\in\cN}C(n, z_{1:T})\bphi(n)\bigg]^\top.\label{eq:sgrad_WOV_manifold}
\end{align}
We then take the expectations over $z_{1:T+1}$ to obtain the negative gradient $-\nabla_{\Wb_{OV}}L(\bTheta)$. We consider the three terms in the last bracket separately, and we first calculate the expectation of the noise term: for each $n\in\cN$, the term $C(n, z_{1:T})\cdot\bphi(n)$ is independent with the tokens $s(z_{1:T})$, $r(z_{1:T})$ and $z_{T+1}$. Therefore,
\begin{align*}
&\mathbb{E}_{z_{1:T+1}}\big[(SR\cdot\bphi(z_{T+1})-\sqrt{SR}\cdot\bar\vb)(C(n, z_{1:T})\cdot\bphi(n))^\top\big]\\
&=\EE_{z_{1:T+1}}[SR\cdot\bphi(z_{T+1})-\sqrt{SR}\cdot\bar\vb]\cdot\EE[C(n, z_{1:T})\cdot\bphi(n)^\top]=\zero;\\
&\mathbb{E}_{z_{1:T+1}}\big[(S\cdot\bphi(z_{T+1})-\sqrt S\cdot\bar\vb_{r(z_{1:T})})(C(n, z_{1:T})\cdot\bphi(n))^\top\big]\\
&=\EE_{z_{1:T+1}}[S\cdot\bphi(z_{T+1})-\sqrt S\cdot\bar\vb_{r(z_{1:T})}]\cdot\EE[C(n, z_{1:T})\cdot\bphi(n)^\top]=\zero;\\
&\mathbb{E}_{z_{1:T+1}}\big[(R\cdot\bphi(z_{T+1})-\sqrt R\cdot\bar\vb_{s(z_{1:T})})(C(n, z_{1:T})\cdot\bphi(n))^\top\big]\\
&=\EE_{z_{1:T+1}}[R\cdot\bphi(z_{T+1})-\sqrt R\cdot\bar\vb_{s(z_{1:T})}]\cdot\EE[C(n, z_{1:T})\cdot\bphi(n)^\top]=\zero.
\end{align*}
Therefore,
\begin{align}
&\EE_{z_{1:T}}\bigg[\big((SR\cdot\bphi(z_{T+1})-\sqrt{SR}\cdot\bar\vb)+(e^{\alpha_R\beta_R/\sqrt S}-1)(S\cdot\bphi(z_{T+1})-\sqrt S\cdot\bar\vb_{r(z_{1:T})})\nonumber\\
&\quad+(e^{\alpha_S\beta_S/\sqrt R}-1)(R\cdot\bphi(z_{T+1})-\sqrt R\cdot\bar\vb_{s(z_{1:T})})\big)\bigg(\sum_{n\in\cN}C(n, z_{1:T})\bphi(n)\bigg)^\top\bigg]=\zero.\label{eq:ee_noise_ov}
\end{align}
We then calculate the term corresponding to $\bphi(s(z_{1:T}))$. Note that
\begin{align*}
&\EE_{z_{1:T+1}}[(SR\cdot\bphi(z_{T+1})-\sqrt{SR}\cdot\bar\vb)\bphi(s(z_{1:T}))^\top]\\
&=\frac1{SR}\sum_{s\in\cS}\bigg\{\sum_{r\in\cR}[SR\cdot\bphi(a_*(s, r))-\sqrt{SR}\cdot\bar\vb]\bigg\}\bphi(s)^\top\\
&=\frac1{SR}\sum_{s\in\cS}(SR\cdot\sqrt R\cdot\bar\vb_s-R\cdot\sqrt{SR}\cdot\bar\vb)\bphi(s)^\top=\sqrt R\sum_{s\in\cS}\tilde\vb_s\bphi(s)^\top;\\
&\EE_{z_{1:T+1}}[(S\cdot\bphi(z_{T+1})-\sqrt S\cdot\bar\vb_{r(z_{1:T})})\bphi(s)^\top]\\
&=\frac{1}{SR}\sum_{s\in\cS}\bigg\{\sum_{r\in\cR}[S\cdot\bphi(a_*(s, r))-\sqrt S\cdot\bar\vb_r]\bigg\}\bphi(s)^\top\\
&=\frac1{SR}\sum_{s\in\cS}(S\cdot\sqrt R\cdot\bar\vb_s-\sqrt R\cdot\sqrt S\cdot\bar\vb)\bphi(s)^\top=\frac1{\sqrt R}\sum_{s\in\cS}\tilde\vb_s\bphi(s)^\top;\\
&\EE_{z_{1:T+1}}[(R\cdot\bphi(z_{T+1})-\sqrt R\cdot\bar\vb_{s(z_{1:T})})\bphi(s)^\top]\\
&=\frac1{SR}\sum_{s\in\cS}\bigg\{\sum_{r\in\cR}[R\cdot\bphi(a_*(s, r))-\sqrt R\cdot\bar\vb_{s}]\bigg\}\bphi(s)^\top\\
&=\frac1{SR}\sum_{s\in\cS}(R\cdot\sqrt R\cdot\bar\vb_s-R\cdot\sqrt R\cdot\bar\vb_s)\bphi(s)^\top=\zero.
\end{align*}
Combining all the above, we have
\begin{align}
&\EE_{z_{1:T+1}}\big\{\big[(SR\cdot\bphi(z_{T+1})-\sqrt{SR}\cdot\bar\vb)+(e^{\alpha_R\beta_R/\sqrt S}-1)(S\cdot\bphi(z_{T+1})-\sqrt S\cdot\bar\vb_{r(z_{1:T})})\nonumber\\
&\quad+(e^{\alpha_S\beta_S/\sqrt R}-1)(R\cdot\bphi(z_{T+1})-\sqrt R\cdot\bar\vb_{s(z_{1:T})})\big]\bphi(s(z_{1:T}))^\top\big\}\nonumber\\
&=\Big(\sqrt{R}+\frac{e^{\alpha_R\beta_R/\sqrt S}-1}{\sqrt R}\Big)\sum_{s\in\cS}\tilde\vb_s\bphi(s)^\top=\frac{R-1+e^{\alpha_R\beta_R/\sqrt S}}{\sqrt R}\sum_{s\in\cS}\tilde\vb_s\bphi(s)^\top.\label{eq:ee_s_ov}
\end{align}
Similarly, for the term corresponding to $\bphi(r(z_{1:T}))$, we have
\begin{align}
&\EE_{z_{1:T}}\big\{\big[(SR\cdot\bphi(z_{T+1})-\sqrt{SR}\cdot\bar\vb)+(e^{\alpha_R\beta_R/\sqrt S}-1)(S\cdot\bphi(z_{T+1})-\sqrt S\cdot\bar\vb_{r(z_{1:T})})\nonumber\\
&\quad+(e^{\alpha_S\beta_S/\sqrt R}-1)(R\cdot\bphi(z_{T+1})-\sqrt R\cdot\bar\vb_{s(z_{1:T})})\big]\bphi(r(z_{1:T}))^\top\big\}\nonumber\\
&=\frac{S-1+e^{\alpha_S\beta_S/\sqrt R}}{\sqrt{S}}\sum_{r\in\cR}\tilde\vb_r\bphi(r)^\top.\label{eq:ee_r_ov}
\end{align}
Plugging \eqref{eq:ee_noise_ov}, \eqref{eq:ee_s_ov}, and \eqref{eq:ee_r_ov} into the expectation of \eqref{eq:sgrad_WOV_manifold}, we have
\begin{align*}
-\nabla_{\Wb_{OV}}L(\bTheta)=\frac{\beta_S/\sqrt R}{S-1+e^{\alpha_S\beta_S/\sqrt R}}\sum_{s\in\cS}\tilde\vb_s\bphi(s)^\top+\frac{\beta_R/\sqrt S}{R-1+e^{\alpha_R\beta_R/\sqrt S}}\sum_{r\in\cR}\tilde\vb_r\bphi(r)^\top.
\end{align*}

\noindent\textbf{Calculation of $-\nabla_{\Wb_{KQ}}L(\bTheta)$.}
We also start from the stochastic gradient. Similar to the analysis of the noise-related terms in $-\nabla_{\Wb_{OV}}\ell(z_{1:T+1}; \bTheta)$, the noise-related term in $\Xb^\top\Xb\Wb_{OV}^\top$, i.e.,
\begin{align*}
\sum_{n\in\cN}C(n, z_{1:T})\cdot\bphi(n)\bar\vb^\top
\end{align*}
is independent from $s(z_{1:T})$, $r(z_{1:T})$, and $z_{T+1}$, and thus has zero contribution in the expectation. We also notice that due to Lemma~\ref{lemma:barv_properties}(a, b),
\begin{align*}
&\bar\vb^\top(SR\cdot\bphi(z_{T+1})-\sqrt{SR}\cdot\bar\vb)=SR\cdot\frac1{\sqrt{SR}}-\sqrt{SR}\cdot1=0\\
&\bar\vb^\top(S\cdot\bphi(z_{T+1})-\sqrt S\cdot\bar\vb_{r(z_{1:T})})=S\cdot\frac1{\sqrt{SR}}-\sqrt S\cdot\frac1{\sqrt R}=0\\
&\bar\vb^\top(R\cdot\bphi(z_{T+1})-\sqrt R\cdot\bar\vb_{s(z_{1:T})})=R\cdot\frac1{\sqrt{SR}}-\sqrt R\cdot\frac1{\sqrt{S}}=0.
\end{align*}
so we also drop the terms related to $\bar\vb$ in $\Xb^\top\Xb\Wb_{OV}^\top$. Therefore,
\begin{align*}
&-\nabla_{\Wb_{KQ}}L(\bTheta)=\EE_{z_{1:T+1}}\Big\{(\alpha_S\bphi(s(z_{1:T}))\bar\vb_{s(z_{1:T})}^\top+\alpha_R\bphi(r(z_{1:T}))\bar\vb_{r(z_{1:T})}^\top)\\
&\quad\cdot\frac{1}{S-1+e^{\alpha_S\beta_S/\sqrt R}}\cdot\frac{1}{R-1+e^{\alpha_R\beta_R/\sqrt S}}\cdot[(SR\cdot\bphi(z_{T+1})-\sqrt{SR}\cdot\bar\vb)\\
&\quad+(e^{\alpha_R\beta_R/\sqrt S}-1)(S\cdot\bphi(z_{T+1})-\sqrt S\cdot\bar\vb_{r(z_{1:T})})\\
&\quad+(e^{\alpha_S\beta_S/\sqrt R}-1)(R\cdot\bphi(z_{T+1})-\sqrt R\cdot\bar\vb_{s(z_{1:T})})]\cdot\bphi(\EOS)^\top\Big\}.
\end{align*}
By Lemma~\ref{lemma:barv_properties}(a, b), we have
\begin{align*}
&\bar\vb_{s(z_{1:T})}^\top(SR\cdot\bphi(z_{T+1})-\sqrt{SR}\cdot\bar\vb)=SR\cdot\frac1{\sqrt R}-\sqrt{SR}\cdot\frac1{\sqrt S}=(S-1)\sqrt R;\\
&\bar\vb_{s(z_{1:T})}^\top(S\cdot\bphi(z_{T+1})-\sqrt S\cdot\bar\vb_{r(z_{1:T})})=S\cdot\frac1{\sqrt R}-\sqrt{S}\cdot\frac1{\sqrt{SR}}=\frac{S-1}{\sqrt R};\\
&\bar\vb_{s(z_{1:T})}^\top(R\cdot\bphi(z_{T+1})-\sqrt R\cdot\bar\vb_{s(z_{1:T})})=R\cdot\frac1{\sqrt R}-\sqrt R\cdot1=0;\\
&\bar\vb_{r(z_{1:T})}^\top(SR\cdot\bphi(z_{T+1})-\sqrt{SR}\cdot\bar\vb)=SR\cdot\frac1{\sqrt S}-\sqrt{SR}\cdot\frac1{\sqrt R}=(R-1)\sqrt S;\\
&\bar\vb_{r(z_{1:T})}^\top(S\cdot\bphi(z_{T+1})-\sqrt S\cdot\bar\vb_{r(z_{1:T})})=S\cdot\frac1{\sqrt S}-\sqrt{S}\cdot1=0;\\
&\bar\vb_{r(z_{1:T})}^\top(R\cdot\bphi(z_{T+1})-\sqrt R\cdot\bar\vb_{s(z_{1:T})})=R\cdot\frac1{\sqrt S}-\sqrt R\cdot\frac{1}{\sqrt{SR}}=\frac{R-1}{\sqrt{S}}.
\end{align*}
Plugging the above into the expression of $-\nabla_{\Wb_{KQ}}L(\bTheta)$, we have
\begin{align*}
&-\nabla_{\Wb_{KQ}}L(\bTheta)\\
&=\EE_{z_{1:T+1}}\Big[\frac{(S-1)\alpha_S/\sqrt R}{S-1+e^{\alpha_S\beta_S/\sqrt R}}\bphi(s(z_{1:T}))+\frac{(R-1)\alpha_R/\sqrt S}{R-1+e^{\alpha_R\beta_R/\sqrt S}}\bphi(r(z_{1:T}))\Big]\cdot\bphi(\EOS)^\top\\
&=\bigg[\frac{(S-1)/(S\sqrt R)\cdot\alpha_S}{S-1+e^{\alpha_S\beta_S/\sqrt R}}\sum_{s\in\cS}\bphi(s)+\frac{(R-1)/(R\sqrt S)\cdot\alpha_R}{R-1+e^{\alpha_R\beta_R/\sqrt S}}\sum_{r\in\cR}\bphi(r)\bigg]\cdot\bphi(\EOS)^\top.
\end{align*}

\end{proof}

\subsection{Proof of Proposition~\ref{prop:spectralGF}}
We need the following lemma to calculate the spectral orthogonalization on $\nabla_{\Wb_{OV}}(\bTheta)$:
\begin{lemma}\label{lemma:msign_grad}
Let $\tilde\vb_s$ and $\tilde\vb_r$ be defined in \eqref{eq:def_tildev}. Let $C_S$ and $C_R$ be arbitrary constants. Then
\begin{align*}
&\msign\bigg(C_S\sum_{s\in\cS}\tilde\vb_s\bphi(s)^\top+C_R\sum_{r\in\cR}\tilde\vb_r\bphi(r)^\top\bigg)\\
&=\sign(C_S)\sum_{s\in\cS}\tilde\vb_s\bphi(s)^\top+\sign(C_R)\sum_{r\in\cR}\tilde\vb_r\bphi(r)^\top.
\end{align*}
\end{lemma}
\begin{proof}[Proof of Proposition~\ref{prop:spectralGF}]
By Lemma~\ref{lemma:msign_grad}, we have
\begin{align*}
\msign(-\nabla_{\Wb_{OV}}L(\bTheta))=\sign(\beta_S)\cdot\sum_{s\in\cS}\tilde\vb_s\bphi(r)^\top+\sign(\beta_R)\cdot\sum_{r\in\cR}\tilde\vb_r\bphi(r)^\top.
\end{align*}
We notice that $-\nabla_{\Wb_{KQ}}L(\bTheta)$ is a rank-1 vector
\begin{align*}
&-\nabla_{\Wb_{KQ}}L(\bTheta)\\
&=\underbrace{\bigg[\frac{(S-1)/(S\sqrt R)\cdot\alpha_S}{S-1+e^{\alpha_S\beta_S/\sqrt R}}\sum_{s\in\cS}\bphi(s)+\frac{(R-1)/(R\sqrt S)\cdot\alpha_R}{R-1+e^{\alpha_R\beta_R/\sqrt S}}\sum_{r\in\cR}\bphi(r)\bigg]}_{\ub}\cdot\bphi(\EOS)^\top.
\end{align*}
Therefore, it suffices to normalize the matrix by $\|\ub\|$, which satisfies
\begin{align*}
\|\ub\|_2^2&=\frac{(S-1)^2/(S^2R)\cdot\alpha_S^2}{(S-1+e^{\alpha_S\beta_S/\sqrt R})^2}\cdot S+\frac{(R-1)^2/(R^2S)\cdot\alpha_R^2}{(R-1+e^{\alpha_R\beta_R/\sqrt S})^2}\cdot R\\
&=\frac1{SR}\bigg[\frac{(S-1)^2\alpha_S^2}{(S-1+e^{\alpha_S\beta_S/\sqrt R})^2}+\frac{(R-1)^2\alpha_R^2}{(R-1+e^{\alpha_R\beta_R/\sqrt S})^2}\bigg].
\end{align*}
Therefore,
\begin{align*}
&\msign(-\nabla_{\Wb_{KQ}}L(\bTheta))=\bigg[\frac{(S-1)^2\alpha_S^2}{(S-1+e^{\alpha_S\beta_S/\sqrt R})^2}+\frac{(R-1)^2\alpha_R^2}{(R-1+e^{\alpha_R\beta_R/\sqrt S})^2}\bigg]^{-1/2}\\
&\quad\cdot\bigg[\frac{(S-1)/\sqrt S\cdot\alpha_S}{S-1+e^{\alpha_S\beta_S/\sqrt R}}\sum_{s\in\cS}\bphi(s)+\frac{(R-1)/\sqrt R\cdot\alpha_R}{R-1+e^{\alpha_R\beta_R/\sqrt S}}\sum_{r\in\cR}\bphi(r)\bigg]\cdot\bphi(\EOS)^\top
\end{align*}
\end{proof}

\subsection{Proof of Proposition~\ref{prop:signGF}}

In this section, we provide the proof of Proposition~\ref{prop:signGF}, which is essentially the calculation of $\sign(-\nabla L(\bTheta))$. Due to the expression of the gradient in Proposition~\ref{prop:GF} and the construction of the embeddings where the subject and relation tokens have disjoint supports, it suffices to calculate the element-wise signs of the following matrices:
\begin{align*}
\sum_{s\in\cS}\tilde\vb_s\bphi(s)^\top,\quad\sum_{r\in\cR}\tilde\vb_r\bphi(r)^\top,\quad\bigg(\sum_{s\in\cS}\bphi(s)\bigg)\bphi(\EOS)^\top,\quad\bigg(\sum_{r\in\cR}\bphi(r)\bigg)\bphi(\EOS)^\top.
\end{align*}
Without loss of generality, we can define $\tilde\vb_s$ and $\tilde\vb_r$ on $\bphi_A$, and replace $\bphi(s), \bphi(r)$ with $\bphi_S(s), \bphi_R(r)$, instead of on the full embedding. The following lemmas based on these notations give the exact expressions of $\sum_{s\in\cS}\tilde\vb_s\bphi(s)^\top$ and $\sum_{r\in\cR}\tilde\vb_r\bphi(r)^\top$.

\begin{lemma}\label{lemma:subject-object}
With different combinations of the embeddings, the element-wise sign of the matrix $\sum_{s\in\cS}\tilde\vb_s\bphi(s)^\top$ satisfy
\begin{itemize}[leftmargin=*]
\item Subject one-hot, answer one-hot:
\begin{align*}
\sign\bigg(\sum_{s\in\cS}\tilde\vb_s\bphi(s)^\top\bigg)=\sum_{s\in\cS}(2\sqrt R\cdot\bar\vb_s-\sqrt{SR}\cdot\bar\vb)\bphi(s)^\top.
\end{align*}
\item Subject one-hot, answer Hadamard:
\begin{align*}
\sign\bigg(\sum_{s\in\cS}\tilde\vb_s\bphi(s)^\top\bigg)=\sqrt S\sum_{s\in\cS}\tilde\vb_s\bphi(s)^\top.
\end{align*}
\item Subject Hadamard, answer one-hot:
\begin{align*}
\sign\bigg(\sum_{s\in\cS}\tilde\vb_s\bphi(s)^\top\bigg)=\sqrt{SR}\sum_{s\in\cS}\tilde\vb_s\bphi(s)^\top.
\end{align*}
\item Subject Hadamard, answer Hadamard:
\begin{align*}
\sign\bigg(\sum_{s\in\cS}\tilde\vb_s\bphi(s)^\top\bigg)=\sum_{s\in\cS}\tilde\vb_s\bphi(s)^\top.
\end{align*}
\end{itemize}
\end{lemma}
\begin{lemma}\label{lemma:relation-object}
With different combinations of the embeddings, the element-wise sign of the matrix $\sum_{r\in\cR}\tilde\vb_r\bphi(r)^\top$ satisfy
\begin{itemize}[leftmargin=*]
\item Subject one-hot, answer one-hot:
\begin{align*}
\sign\bigg(\sum_{r\in\cR}\tilde\vb_r\bphi(r)^\top\bigg)=\sum_{r\in\cR}(2\sqrt S\cdot\bar\vb_r-\sqrt{SR}\cdot\bar\vb)\bphi(r)^\top.
\end{align*}
\item Relation one-hot, answer Hadamard:
\begin{align*}
\sign\bigg(\sum_{r\in\cR}\tilde\vb_r\bphi(r)^\top\bigg)=\sqrt S\sum_{r\in\cR}\tilde\vb_r\bphi(r)^\top.
\end{align*}
\item Relation Hadamard, answer one-hot:
\begin{align*}
\sign\bigg(\sum_{r\in\cR}\tilde\vb_r\bphi(r)^\top\bigg)=\sqrt{SR}\sum_{r\in\cR}\tilde\vb_r\bphi(r)^\top.
\end{align*}
\item Relation Hadamard, answer Hadamard:
\begin{align*}
\sign\bigg(\sum_{r\in\cR}\tilde\vb_r\bphi(r)^\top\bigg)=\sum_{r\in\cR}\tilde\vb_r\bphi(r)^\top.
\end{align*}
\end{itemize}
\end{lemma}

\begin{lemma}\label{lemma:SR_mean}
With different embeddings, the element-wise sign of $(\sum_{s\in\cS}\bphi(s))\bphi(\EOS)^\top$ satisfies
\begin{itemize}[leftmargin=*]
\item When the subject embeddings are one-hot, then
\begin{align*}
\sign\bigg(\bigg[\sum_{s\in\cS}\bphi(s)\bigg]\bphi(\EOS)^\top\bigg)=\bigg[\sum_{s\in\cS}\bphi(s)\bigg]\bphi(\EOS)^\top.
\end{align*}
\item When the subject embeddings are Hadamard, then
\begin{align*}
\sign\bigg(\bigg[\sum_{s\in\cS}\bphi(s)\bigg]\bphi(\EOS)^\top\bigg)=\frac1{\sqrt S}\bigg[\sum_{s\in\cS}\bphi(s)\bigg]\bphi(\EOS)^\top.
\end{align*}
\end{itemize}
The relation embeddings satisfy similar properties.
\end{lemma}
Lemma~\ref{lemma:SR_mean} holds by noticing that
\begin{align*}
\sum_{s\in\cS}\eb_s=\one_S,\quad\sum_{s\in\cS}\Hb_S\eb_s=\Hb_S\one_S=\sqrt S\eb_1.
\end{align*}

\begin{proof}[Proof of Proposition~\ref{prop:signGF}]

Proposition~\ref{prop:signGF} holds by combining the three lemmas above.
\end{proof}

\section{Proof of Learning Time Results}\label{section:proof_learning_time}

\subsection{Proof of Theorem~\ref{theorem:GF}}

\begin{proof}[Proof of Theorem~\ref{theorem:GF}]
We first consider the ODE of $\alpha_S$ and $\beta_S$, and the ODE of $\alpha_R$ and $\beta_R$ can be analyzed similarly.
The derivative of the margin $q_S$ satisfies
\begin{align*}
\frac{\ud q_S}{\ud t}=\frac{\alpha_S}{\sqrt R}\cdot\frac{\ud\beta_S}{\ud t}+\frac{\ud\alpha_S}{\ud t}\cdot\frac{\beta_S}{\sqrt R}=\frac{(S-1)\alpha_S^2/S+\beta_S^2}{R(S-1+e^{\alpha_S\beta_S/\sqrt R})}=\frac{\alpha_S^2/S+\beta_S^2/(S-1)}{R(1+e^{q_S})}.
\end{align*}
We aim to express the right-hand side as a function of $q_S$. Notice that $\alpha_S^2/S-\beta_S^2/(S-1):=C_S$ is a constant because
\begin{align*}
\frac{\ud[\alpha_S^2/S-\beta_S^2/(S-1)]}{\ud t}=2\Big[\frac{\ud\alpha_S}{\ud t}\cdot\frac{\alpha_S}S-\frac{\ud\beta_S}{\ud t}\cdot\frac{\beta_S}{S-1}\Big]=0.
\end{align*}
We also notice that
\begin{align*}
\alpha_S^2/S+\beta_S^2/(S-1)&=\sqrt{[\alpha_S^2/S-\beta_S^2/(S-1)]^2+4\alpha_S^2\beta_S^2/(S(S-1))}\\
&=\sqrt{C_S^2+4R[q_S+\log(S-1)]^2/(S(S-1))}.
\end{align*}
Therefore, the derivative of $q_S$ can be written as the following expression with only $q_S$ and the constant $C_S$:
\begin{align*}
\frac{\ud q_S}{\ud t}=\frac{\sqrt{C_S^2/R^2+4[q_S+\log(S-1)]^2/(SR(S-1))}}{1+e^{q_S}}.
\end{align*}
Separating variables and taking the integral, we have
\begin{align*}
T_S^\delta&=\int_{\alpha_0\beta_0/\sqrt R-\log(S-1)}^{\log(\delta^{-1}-1)}\frac{1+e^{q_S}}{\sqrt{C_S^2/R^2+4[q_S+\log(S-1)]^2/(SR(S-1))}}\ud q_S\\
&=\underbrace{\int_{\alpha_0\beta_0/\sqrt R-\log(S-1)}^{\log(\delta^{-1}-1)}\frac{\ud q_S}{\sqrt{C_S^2/R^2+4[q_S+\log(S-1)]^2/(SR(S-1))}}}_{I_{S, 1}}\\
&\quad+\underbrace{\int_{\alpha_0\beta_0/\sqrt R-\log(S-1)}^{1-\log(S-1)}\frac{e^{q_S}\ud q_S}{\sqrt{C_S^2/R^2+4[q_S+\log(S-1)]^2/(SR(S-1))}}}_{I_{S, 2}}\\
&\quad+\underbrace{\int_{1-\log(S-1)}^{\log(\delta^{-1}-1)}\frac{e^{q_S}\ud q_S}{\sqrt{C_S^2/R^2+4[q_S+\log(S-1)]^2/(SR(S-1))}}}_{I_{S, 3}}
\end{align*}
For the term $I_{S, 1}$, we directly calculate the integral:
\begin{align*}
I_{S, 1}&=\frac{\sqrt{SR(S-1)}}{2}\bigg[\log\bigg(\log(\delta^{-1}-1)(S-1)+\sqrt{[\log(\delta^{-1}-1)(S-1)]^2+\frac{S(S-1)C_S^2}{4R}}\bigg)\\
&\quad-\log\bigg(\frac{\alpha_0\beta_0}{\sqrt R}+\sqrt{\frac{\alpha_0^2\beta_0^2}{R}+\frac{S(S-1)C_S^2}{4R}}\bigg)\bigg].
\end{align*}
Note that $S(S-1)C_S^2/(4R)$ is a lower-order term because
\begin{align}
\frac{S(S-1)C_S^2}{4R}&=\frac{S(S-1)}{4R}[\alpha_0^2/S-\beta_0^2/(S-1)]^2\nonumber\\
&\le\frac{S(S-1)}{4R}\max\Big\{\frac{\alpha_0^4}{S^2}, \frac{\beta_0^4}{(S-1)^2}\Big\}\nonumber\\
&\le\frac{S}{4R(S-1)}\le1,\label{eq:CS_bound}
\end{align}
where the first inequality holds because $|A-B|\le\max\{A, B\}$, the second inequality holds due to the assumption $\alpha_0, \beta_0\le1$, and the last inequality holds because $S, R\ge2$. We also have
\begin{align*}
\frac{\alpha_0\beta_0}{\sqrt R}+\sqrt{\frac{\alpha_0^2\beta_0^2}{R}+\frac{S(S-1)C_S^2}{4R}}=\frac1{2\sqrt R}\Big[\Big(\frac{S-1}{S}\Big)^{1/4}\alpha_0+\Big(\frac{S}{S-1}\Big)^{1/4}\beta_0\Big]^2\simeq\frac{(\alpha_0+\beta_0)^2}{2\sqrt R}
\end{align*}
Therefore, $I_{S, 1}$ is approximated as
\begin{align}\label{eq:IS1_final}
I_{S, 1}\simeq\sqrt{SR(S-1)}\log\frac{\sqrt R\log(\delta^{-1}-1)(S-1)}{\alpha_0+\beta_0}.
\end{align}

We then show that $I_{S, 2}$ is controlled by $I_{S, 1}$, i.e.,
\begin{align}\label{eq:IS2<IS1}
I_{S, 2}\lesssim I_{S, 1}.
\end{align}
It suffices to show that $e^{q_S}\lesssim1$, which holds because
\begin{align*}
e^{q_S}\le e/(S-1)\le e,
\end{align*}
where the first inequality holds because $q_S\le1-\log(S-1)$, and the second inequality holds because $S\ge2$.

For the term $I_{S, 3}$, we use the bound for $C_S$ in \eqref{eq:CS_bound} and have
\begin{align}\label{eq:IS3}
I_{S, 3}&=\frac{\sqrt{SR(S-1)}}2\int_{1-\log(S-1)}^{\log(\delta^{-1}-1)}\frac{e^{q_S}\ud q_S}{\sqrt{\frac{S(S-1)C_S^2}{4R}+[q_S+\log(S-1)]^2}}\nonumber\\
&\simeq\sqrt{SR(S-1)}\int_{1-\log(S-1)}^{\log(\delta^{-1}-1)}\frac{e^{q_S}\ud q_S}{q_S+\log(S-1)}\nonumber\\
&=\sqrt{\frac{SR}{S-1}}\int_1^{\log(\delta^{-1}-1)(S-1)}\frac{e^u\ud u}{u}\nonumber\\
&\simeq\sqrt{SR(S-1)}\cdot\frac{\delta^{-1}-1}{\log(\delta^{-1}-1)(S-1)},
\end{align}
where the first equality holds because $\frac{S(S-1)C_S^2}{4R}\lesssim1$ but $q_S+\log(S-1)\ge1$, and the last equality holds due to Lemma~\ref{lemma:Ei_estimate}. Combining \eqref{eq:IS1_final}, \eqref{eq:IS2<IS1}, and \eqref{eq:IS3}, we have
\begin{align*}
T_S^\delta\simeq\sqrt{SR(S-1)}\bigg[\log\frac{\sqrt R\log(S-1)}{\alpha_0+\beta_0}+\frac{\delta^{-1}-1}{\log(\delta^{-1}-1)(S-1)}\bigg].
\end{align*}
We similarly have
\begin{align*}
T_R^\delta\simeq\sqrt{SR(R-1)}\bigg[\log\frac{\sqrt S\log(R-1)}{\alpha_0+\beta_0}+\frac{\delta^{-1}-1}{\log(\delta^{-1}-1)(R-1)}\bigg].
\end{align*}
\end{proof}

\subsection{Proof of Theorem~\ref{theorem:SpectralGF}}

The following lemma gives a lower bound for $\beta_S$ and an upper bound for $\beta_R$:
\begin{lemma}\label{lemma:betaS_lower_betaR_upper}
Suppose that Assumption~\ref{assumption:init} holds. Then the Spectral GF dynamics satisfy
\begin{align*}
\beta_S(t)&\ge\frac{t}{2\sqrt S}+\frac{1+\sqrt{R/S}}{2}\beta_0,\\
\beta_R(t)&\le\frac{t}{\sqrt{2R}}+\frac{1+\sqrt{S/R}}{2}\beta_0.
\end{align*}
\end{lemma}
The following lemma characterizes the learning times $T_S^{0.5}$ and $T_R^{0.5}$, as well as the corresponding manifold parameters:
\begin{lemma}\label{lemma:T0.5}
Under the same assumptions as Theorem~\ref{theorem:SpectralGF}, the learning times of the spectral GF for the subject and relation component to reach $0.5$ accuracy satisfy
\begin{align*}
T_S^{0.5}\simeq(SR)^{1/4}\sqrt{\log(S-1)},\quad T_R^{0.5}\simeq(SR)^{1/4}\sqrt{\log(R-1)},
\end{align*}
and the manifold parameters satisfy
\begin{align*}
&\alpha_S(T_S^{0.5})\simeq(SR)^{1/4}\sqrt{\log(S-1)},\quad\beta_S(T_S^{0.5})\simeq(R/S)^{1/4}\sqrt{\log(S-1)};\\
&\alpha_S(T_R^{0.5})\simeq(SR)^{1/4}\sqrt{\log(R-1)},\quad\beta_R(T_R^{0.5})\simeq(S/R)^{1/4}\sqrt{\log(R-1)}.
\end{align*}
\end{lemma}
We now present the proof of Theorem~\ref{theorem:SpectralGF}.
\begin{proof}[Proof of Theorem~\ref{theorem:SpectralGF}]
\noindent\textbf{Characterization of $T_S^\delta$.}
We notice that the derivative of $\beta_S$ satisfies
\begin{align*}
\frac{\ud\beta_S}{\ud t}=\frac{(1+e^{q_R})/\sqrt S}{\sqrt{(1+e^{q_S})^2+(1+e^{q_R})^2}}\le1/\sqrt S,
\end{align*}
so combining the above with the lower bound of $\beta_S$ in Lemma~\ref{lemma:betaS_lower_betaR_upper}, we have
\begin{align*}
\frac{t}{2\sqrt S}+\frac{1+\sqrt{R/S}}{2}\beta_0\le\beta_S(t)\le\frac{t}{\sqrt S}+\beta_0.
\end{align*}
Since $0<\sqrt{R/S}\le1$, we have $\beta_S(t)\simeq t/\sqrt S+\beta_0$.
According to the definition of $T_S^\delta$, we have
\begin{align*}
\sqrt R\log(\delta^{-1}-1)(S-1)=\alpha_S(T_S^\delta)\cdot\beta_S(T_S^\delta)\simeq(T_S^\delta+\alpha_0)(T_S^\delta/\sqrt S+\beta_0).
\end{align*}
Due to the assumptions $\alpha_0\le(SR)^{1/4}$ and $\beta_0\le(R/S)^{1/4}$, we have
\begin{align*}
T_S^\delta\simeq(SR)^{1/4}\sqrt{\log(\delta^{-1}-1)(S-1)}.
\end{align*}

\noindent\textbf{Characterization of $T_R^\delta$.} We develop the characterization of $T_R^\delta$ in 5 steps.

\textit{Step 1: Universal upper bound of $T_R^\delta$.}
Since $\beta_R$ is increasing, we have for any time $t\ge T_R^{0.5}$,
\begin{align*}
\beta_R(t)\ge\beta_R(T_R^{0.5})=\frac{\sqrt S\log(R-1)}{\alpha_R(T_R^{0.5})}.
\end{align*}
Combining the above with the definition of $T_R^\delta$, we have
\begin{align*}
\log(\delta^{-1}-1)(R-1)&=\frac{\alpha_R(T_R^\delta)\cdot\beta_R(T_R^\delta)}{\sqrt S}\\
&\ge\frac{\alpha_R(T_R^\delta)}{\alpha_R(T_R^{0.5})}\cdot\log(R-1)\\
&\simeq\frac{T_R^\delta}{T_R^{0.5}}\cdot\log(R-1)\\
&\simeq(SR)^{-1/4}\sqrt{\log(R-1)}\cdot T_R^\delta,
\end{align*}
where the third line holds because $\alpha_0=(SR)^{1/4}\lesssim T_R^{0.5}\le T_R^\delta$, and the last line holds due to Lemma~\ref{lemma:T0.5}. We thus have
\begin{align}\label{eq:TS_universal_upper_bound}
T_R^\delta\lesssim(SR)^{1/4}\frac{\log(\delta^{-1}-1)(R-1)}{\sqrt{\log(R-1)}}.
\end{align}

\textit{Step 2: The above upper bound is tight when $T_R^\delta<T_S^{0.5}$.}
The above upper bound relies on using $\beta_R(T_R^{0.5})$ to lower bound $\beta_R(T_R^\delta)$. We now show that when $T_R^\delta<T_S^{0.5}$, we have $\beta_R(T_R^\delta)\simeq\beta_R(T_R^{0.5})$, so the above upper bound is tight. Note that when $T_R^{0.5}\le t\le T_S^{0.5}$, we have $q_S\le0\le q_R$, and
\begin{align}\label{eq:dbetaR_dt_upper}
\frac{\ud\beta_R}{\ud t}=\frac{1/\sqrt R}{\sqrt{1+(\frac{1+e^{q_R}}{1+e^{q_S}})^2}}\le\frac{1/\sqrt R}{\sqrt{1+(\frac{1+e^{q_R}}{2})^2}}\le\frac{2}{\sqrt R e^{q_R}},
\end{align}
where the first inequality holds because $q_S\ge1$, and the second inequality holds because $1+(\frac{1+e^{q_R}}{2})^2\ge\frac{e^{2q_R}}{4}$.
We then derive a lower bound for $q_R$ when $T_R^{0.5}\le t\le T_S^{0.5}$:
\begin{align}
q_R(t)&=\frac{\alpha_R(t)\beta_R(t)}{\sqrt S}-\log(R-1)\nonumber\\
&=\frac{\alpha_R(t)\beta_R(t)-\alpha_R(T_R^{0.5})\beta_R(T_R^{0.5})}{\sqrt S}\nonumber\\
&\ge\frac{[\alpha_R(t)-\alpha_R(T_R^{0.5})]\beta_R(T_R^{0.5})}{\sqrt S}\nonumber\\
&=(t-T_R^{0.5})\cdot\frac{\log(R-1)}{\alpha_R(T_R^{0.5})},\label{eq:qR_lower}
\end{align}
where the second equality holds due to the definition of $T_R^{0.5}$, and inequality holds because $\beta_R(t)\ge\beta_R(T_R^{0.5})$ for $t\ge T_R^{0.5}$, and the last equality holds due to the definition of $T_R^{0.5}$ and $\alpha_R(t)=t+\alpha_0$. Plugging \eqref{eq:qR_lower} into \eqref{eq:dbetaR_dt_upper}, we have
\begin{align*}
\beta_R(T_R^\delta)-\beta_R(T_R^{0.5})&\le\int_{T_R^{0.5}}^{T_R^\delta}\frac{2\ud t}{\sqrt R e^{\frac{\log(R-1)}{\alpha_R(T_R^{0.5})}\cdot(t-T_R^{0.5})}}\\
&\le\int_{T_R^{0.5}}^{\infty}\frac{2\ud t}{\sqrt R e^{\frac{\log(R-1)}{\alpha_R(T_R^{0.5})}\cdot(t-T_R^{0.5})}}\\
&=\frac{2\alpha_R(T_R^{0.5})}{\sqrt R\log(R-1)}\\
&\frac{\simeq(S/R)^{1/4}}{\sqrt{\log(R-1)}}
\end{align*}
where the last equality holds due to Lemma~\ref{lemma:T0.5}. Since $\beta_R(T_R^{0.5})\simeq(S/R)^{1/4}\sqrt{\log(R-1)}$ according to Lemma~\ref{lemma:T0.5}, we have $\beta(T_R^\delta)\simeq\beta_R(T_R^{0.5})$, and the derivation of the upper bound is tight, i.e.,
\begin{align*}
T_R^\delta\simeq(SR)^{1/4}\frac{\log(\delta^{-1}-1)(R-1)}{\sqrt{\log R-1}}.
\end{align*}

\textit{Step 3: Lower bound of $T_R^\delta$ when $T_R^\delta\ge T_S^{0.5}$.}
Due to the definition of $T_R^\delta$, we have
\begin{align*}
\log(\delta^{-1}-1)(R-1)=\frac{\alpha_R(T_R^\delta)\beta_R(T_R^\delta)}{\sqrt S}\le\frac{T_R^\delta+\alpha_0}{\sqrt S}\cdot\Big(\frac{T_R^\delta}{\sqrt{2R}}+\frac{1+\sqrt{S/R}}{2}\beta_0\Big),
\end{align*}
where the last equality holds due to Lemma~\ref{lemma:betaS_lower_betaR_upper}. Since $\alpha_0\le(SR)^{1/4}$ and $\beta_0\le(R/S)^{1/4}$, we have
\begin{align*}
T_R^\delta&\gtrsim\max\{T_S^{0.5}, (SR)^{1/4}\sqrt{\log(\delta^{-1}-1)(R-1)}\}\\
&\simeq(SR)^{1/4}\sqrt{\max\{\log(S-1), \log(\delta^{-1}-1)+\log(R-1)\}}\\
&\simeq(SR)^{1/4}\sqrt{\log(\delta^{-1}-1)(S-1)}.
\end{align*}
where the first equality holds due to Lemma~\ref{lemma:T0.5}, and the last equality holds because $\max\{A, B\}\simeq A+B$ and $S\ge R$.

\textit{Step 4: Upper bound of $T_R^\delta$ when $T_R^\delta\ge T_S^{0.5}$.} In Step 3, we derived the lower bound of $T_R^\delta$ which has the same order as $T_S^\delta$. If $T_S^\delta\ge T_R^\delta$, then the bound is tight, so we focus on the case where $T_S^\delta<T_R^\delta$. We study the dynamics of $m(t):=\min\{q_S(t), q_R(t)\}$ for $t\ge\tau:=\max\{T_S^{0.5}, T_R^{0.5}\}$. Due to the definition and the assumption $T_S^\delta<T_R^\delta$, we immediately have
\begin{align}\label{eq:m(tau)=0}
m(\tau)=0,\quad m(T_R^\delta)=q_R(T_R^\delta)=\log(\delta^{-1}-1).
\end{align}
We then calculate the derivative of $m(t)$, and it suffices to consider the derivatives of $q_S$ and $q_R$. Suppose that $q_S(t)\le q_R(t)$ for a given $t$, then we have
\begin{align*}
\frac{\ud\beta_S(t)}{\ud t}=\frac{(1+e^{q_R})/\sqrt S}{\sqrt{(1+e^{q_S})^2+(1+e^{q_R})^2}}\ge\frac1{\sqrt{2S}}.
\end{align*}
The derivative of $m(t)$ thus satisfies
\begin{align*}
\frac{\ud m(t)}{\ud t}&=\frac{\ud q_S(t)}{\ud t}=\frac{\ud\alpha_S(t)}{\ud t}\cdot\frac{\beta_S(t)}{\sqrt R}+\frac{\alpha_S(t)}{\sqrt R}\cdot\frac{\ud\beta_R(t)}{\ud t}\\
&\ge\frac{\beta_S(t)}{\sqrt R}+\frac{t+t_0}{\sqrt{2RS}}\\
&=\frac{q_S(t)+\log(S-1)}{t+t_0}+\frac{t+t_0}{\sqrt{2RS}},
\end{align*}
where the last equality holds due to the definition of $q_S$. Similarly, when $q_S(t)\ge q_R(t)$, we have
\begin{align*}
\frac{\ud m(t)}{\ud t}\ge\frac{q_R(t)+\log(R-1)}{t+t_0}+\frac{t+t_0}{\sqrt{2RS}}.
\end{align*}
Combining the two cases, noting that $S\ge R$, we have
\begin{align*}
\frac{\ud m(t)}{\ud t}\ge\frac{m(t)+\log(R-1)}{t+t_0}+\frac{t+t_0}{\sqrt{2RS}}.
\end{align*}
We thus have
\begin{align*}
\frac{\ud}{\ud t}\bigg(\frac{m(t)+\log(R-1)}{t+\alpha_0}\bigg)&=\frac{1}{t+\alpha_0}\cdot\frac{\ud m(t)}{\ud t}-\frac{m(t)+\log(R-1)}{(t+\alpha_0)^2}\\
&\ge\frac{m(t)+\log(R-1)}{(t+\alpha_0)^2}+\frac1{\sqrt{2SR}}-\frac{m(t)+\log(R-1)}{(t+\alpha_0)^2}\\
&=\frac1{\sqrt{2SR}}.
\end{align*}
Taking the integral from $t=\tau$ to $t=T_R^\delta$, we have
\begin{align*}
\frac{T_R^\delta-\tau}{\sqrt{2SR}}&\le\frac{m(T_R^\delta)+\log(R-1)}{T_R^\delta+\alpha_0}-\frac{m(\tau)+\log(R-1)}{\tau+\alpha_0}\\
&=\frac{\log(\delta^{-1}-1)(R-1)}{T_R^\delta+\alpha_0}-\frac{\log(R-1)}{\tau+\alpha_0},
\end{align*}
where the equality holds due to \eqref{eq:m(tau)=0}. Solving the inequality, we have
\begin{align*}
T_R^\delta&\le\frac12\Big(\tau-\alpha_0-\frac{\sqrt{2SR}\log(R-1)}{\tau+\alpha_0}\Big)\\
&\quad+\sqrt{\frac14\Big(\tau+\alpha_0+\frac{\sqrt{2SR}\log(R-1)}{\tau+\alpha_0}\Big)^2+\sqrt{2SR}\log(\delta^{-1}-1)}\\
&\le\frac12\Big(\tau-\alpha_0-\frac{\sqrt{2SR}\log(R-1)}{\tau+\alpha_0}\Big)\\
&\quad+\frac12\Big(\tau+\alpha_0+\frac{\sqrt{2SR}\log(\delta^{-1}-1)}{\tau+\alpha_0}\Big)+\sqrt{\sqrt{2SR}\log(\delta^{-1}-1)}\\
&=\tau+(2SR)^{1/4}\sqrt{\log(\delta^{-1}-1)}\\
&\simeq(SR)^{1/4}(\sqrt{\log(S-1)}+\sqrt{\log(\delta^{-1}-1)})\\
&\simeq(SR)^{1/4}\sqrt{\log(\delta^{-1}-1)(S-1)},
\end{align*}
where the second inequality holds because $\sqrt{A^2+B^2}\le A+B$, and the second equality holds due to Lemma~\ref{lemma:T0.5}.

\textit{Step 5: Combining all results together.} We have so far derived matching upper and lower bounds of $T_R^\delta$ under two regimes $T_R^\delta<T_S^{0.5}$ and $T_R^\delta\ge T_S^{0.5}$. However, these two regimes are given by conditions that are not directly related to $S$, $R$, and $\delta$. We now discuss when these two approximations apply. We first notice that Steps 1-4 indicate that
\begin{align*}
T_R^\delta\simeq(SR)^{1/4}\cdot\min\bigg\{\frac{\log(\delta^{-1}-1)(R-1)}{\sqrt{\log(R-1)}}, \sqrt{\log(\delta^{-1}-1)(S-1)}\bigg\}.
\end{align*}
When we have
\begin{align*}
\log(\delta^{-1}-1)\le c\cdot\sqrt{\log(S-1)\cdot\log(R-1)},
\end{align*}
then
\begin{align*}
\frac{\log(\delta^{-1}-1)(R-1)}{\sqrt{\log(R-1)}}&\lesssim\frac{\sqrt{\log(S-1)\cdot\log(R-1)}+\log(R-1)}{\sqrt{\log(R-1)}}\\
&=\sqrt{\log(S-1)}+\sqrt{\log(R-1)}\\
&\simeq\sqrt{\log(S-1)}\\
&\le\sqrt{\log(\delta^{-1}-1)(S-1)},
\end{align*}
where the third line holds because $S\ge R$. On the other hand, we aim to prove that 
\begin{align*}
\frac{\log(\delta^{-1}-1)(R-1)}{\sqrt{\log(R-1)}}\gtrsim\sqrt{\log(\delta^{-1}-1)(S-1)}
\end{align*}
under the condition of
\begin{align*}
\log(\delta^{-1}-1)\ge c\cdot\sqrt{\log(S-1)\cdot\log(R-1)}.
\end{align*}
Following the similar logic of the previous deduction, we have
\begin{align*}
\frac{\log(\delta^{-1}-1)(R-1)}{\sqrt{\log(R-1)}}&\gtrsim\sqrt{\log(S-1)}.
\end{align*}
Therefore, if $\log(\delta^{-1}-1)\le\log(S-1)$, then
\begin{align*}
\sqrt{\log(S-1)}\simeq\sqrt{\log(\delta^{-1}-1)(S-1)}.
\end{align*}
Otherwise, when $\log(\delta^{-1}-1)\ge\log(S-1)$, we have
\begin{align*}
\frac{\log(\delta^{-1}-1)(R-1)}{\sqrt{\log(R-1)}}\lesssim\frac{\log(\delta^{-1}-1)}{\sqrt{\log(R-1)}}\lesssim\frac{\log(\delta^{-1}-1)}{\sqrt{\log(\delta^{-1}-1)}}\simeq\sqrt{\log(\delta^{-1}-1)(S-1)}
\end{align*}
where the first two inequalities hold because $\log(R-1)\le\log(S-1)\lesssim\log(\delta^{-1}-1)$, and the last equality holds because $\log(S-1)\le\log(\delta^{-1}-1)$. Therefore, the final result is
\begin{align*}
T_R^\delta\simeq\begin{cases}
(SR)^{1/4}\frac{\log(\delta^{-1}-1)(R-1)}{\sqrt{\log(R-1)}} & \text{if}\quad\log(\delta^{-1}-1)\le c\cdot\sqrt{\log(S-1)\cdot\log(R-1)};\\
(SR)^{1/4}\sqrt{\log(\delta^{-1}-1)(S-1)} & \text{otherwise}.
\end{cases}
\end{align*}
\end{proof}

\subsection{Proof of Learning Times for Sign GF}

The following theorem provides general results about the learning times:
\begin{theorem}\label{theorem:SignGF}
Consider the general ODE
\begin{align*}
\dot\alpha_S=C_{\alpha_S},\quad\dot\beta_S=C_{\beta_S},\quad\dot\alpha_R=C_{\alpha_R},\quad\dot\beta_R=C_{\beta_R}.
\end{align*}
Assume that
\begin{align*}
\alpha_0\le\min\bigg\{R^{1/4}\sqrt{\frac{C_{\alpha_S}}{C_{\beta_S}}}, S^{1/4}\sqrt{\frac{C_{\alpha_R}}{C_{\beta_R}}}\bigg\};\quad\beta_0\le\min\bigg\{R^{1/4}\sqrt{\frac{C_{\beta_S}}{C_{\alpha_S}}}, S^{1/4}\sqrt{\frac{C_{\beta_R}}{C_{\alpha_R}}}\bigg\}.
\end{align*}
Then the learning times satisfy
\begin{align*}
T_S^\delta\simeq R^{1/4}\sqrt{\frac{\log(\delta^{-1}-1)(R-1)}{C_{\alpha_S}\cdot C_{\beta_S}}},\quad T_R^\delta\simeq S^{1/4}\sqrt{\frac{\log(\delta^{-1}-1)(S-1)}{C_{\alpha_R}\cdot C_{\beta_R}}}.
\end{align*}
\end{theorem}

\begin{proof}[Proof of Theorem~\ref{theorem:SignGF}]
We firstly notice that
\begin{align*}
\alpha_S=C_{\alpha_S}\cdot t+\alpha_0,\quad\beta_S=C_{\beta_S}\cdot t+\beta_0.
\end{align*}
By definition of $T_S^\delta$, we have
\begin{align*}
\sqrt R\log(\delta^{-1}-1)(S-1)=(C_{\alpha_S}\cdot T_S^\delta+\alpha_0)\cdot(C_{\beta_S}\cdot T_S^\delta+\beta_0).
\end{align*}
Since
\begin{align*}
\alpha_0\le R^{1/4}\sqrt{C_{\alpha_S}/C_{\beta_S}},\quad\beta_0\le R^{1/4}\sqrt{C_{\beta_S}/C_{\alpha_S}},
\end{align*}
we have $T_S^\delta\gtrsim R^{1/4}/\sqrt{C_{\alpha_S}\cdot C_{\alpha_R}}$. We thus have
\begin{align*}
(C_{\alpha_S}\cdot T_S^\delta+\alpha_0)\cdot(C_{\beta_S}\cdot T_S^\delta+\beta_0)\simeq(C_{\alpha_S}C_{\beta_S})\cdot (T_S^\delta)^2.
\end{align*}
Therefore,
\begin{align*}
T_S^\delta\simeq R^{1/4}\sqrt{\frac{\log(\delta^{-1}-1)(R-1)}{C_{\alpha_S}\cdot C_{\beta_S}}}.
\end{align*}
The chracterization of $T_R^\delta$ is similar.
\end{proof}

\section{Proof of Lemmas in Appendix~\ref{section:proof_dynamics} and Appendix~\ref{section:proof_learning_time}}

\subsection{Proof of Lemma~\ref{lemma:msign_grad}}

\begin{proof}[Proof of Lemma~\ref{lemma:msign_grad}]
We assume without loss of generality that $C_S, C_R\neq0$. The case of $C_S=0$ or $C_R=0$ can be handled in a similar way.
Let $\Vb_S\in\RR^{d\times S}$ and $\Vb_R\in\RR^{d\times R}$ be the matrices formed by horizontally stacking $\{\tilde\vb_s\}_{s\in\cS}$ and $\{\tilde\vb_r\}_{r\in\cR}$, respectively. Let $\bPhi_S\in\RR^{d\times S}$ and $\bPhi_R\in\RR^{d\times R}$ be the matrices formed by horizontally stacking $\{\bphi(s)\}_{s\in\cS}$ and $\{\bphi(r)\}_{r\in\cR}$, respectively. Then
\begin{align*}
C_R\sum_{s\in\cS}\tilde\vb_s\phi(s)+C_R\sum_{r\in\cR}\tilde\vb_r\bphi(r)=C_S\Vb_S\bPhi_S^\top+C_R\Vb_R\bPhi_R^\top.
\end{align*}
Note that $(\bPhi_S, \bPhi_R)$ forms an orthonormal basis:
\begin{align}\label{eq:Phi_properties}
\bPhi_S^\top\bPhi_S=\Ib_S,\quad\bPhi_R^\top\bPhi_R=\Ib_R,\quad\bPhi_S^\top\bPhi_R=\zero.
\end{align}
However, $(\Vb_S, \Vb_R)$ do not have such properties: By Lemma~\ref{lemma:barv_properties}(c), we have
\begin{align*}
\Vb_S^\top\Vb_S=\Ib_S-\one_{S, S}/S,\quad\Vb_R^\top\Vb_R=\Ib_R-\one_{R, R}/R,\quad\Vb_S^\top\Vb_R=\zero.
\end{align*}
To solve this problem, we define the matrix $\Qb_S$:
\begin{align*}
\Qb_S=\begin{bmatrix}
1/\sqrt2 & 1/\sqrt6 & \cdots & 1/\sqrt{S(S-1)}\\
-1/\sqrt 2 & 1/\sqrt6 & \cdots & 1/\sqrt{S(S-1)}\\
& -2/\sqrt6 & \cdots & 1/\sqrt{S(S-1)}\\
& & \ddots & \vdots\\
& & & (1-S)/\sqrt{S(SS1)}
\end{bmatrix}\in\RR^{S\times(S-1)}.
\end{align*}
We note that
\begin{align*}
\Qb_S^\top\Qb_S=\Ib_{S-1},\quad\Qb_S\Qb_S^\top=\Ib_S-\one_{S, S}/S,
\end{align*}
so
\begin{align}\label{eq:VQQ=V}
\Vb_S\Qb_S\Qb_S^\top=\Vb_S-\frac{\Vb\cdot\one_S\cdot\one_S^\top}{S}=\Vb_S-\bigg[\sum_{s\in\cS}(\bar\vb_s-\bar\vb/\sqrt S)\bigg]\cdot\frac{\one_S^\top}{S}=\Vb_S,
\end{align}
where the last inequality holds due to the definition of $\bar\vb_s$ and $\bar\vb_r$.
We also define $\Qb_R$ similarly, and we similarly have $\Vb_R\Qb_R\Qb_R^\top=\Vb_R$. Therefore,
\begin{align}\label{eq:SVD}
&C_S\Vb_S\bPhi_S^\top+C_R\Vb_R\bPhi_R^\top=C_S\Vb_S\Qb_S\Qb_S^\top\bPhi_S^\top+C_R\Vb_R\Qb_R\Qb_R^\top\bPhi_R\\
&=\begin{bmatrix}
\sign(C_S)\cdot\Vb_S\Qb_S & \sign(C_R)\cdot\Vb_R\Qb_R
\end{bmatrix}\cdot\begin{bmatrix}
|C_S|\Ib_{S-1} & \zero \\ \zero & |C_R|\Ib_{R-1}
\end{bmatrix}\cdot\begin{bmatrix}
\bPhi_S\Qb_S & \bPhi_R\Qb_R
\end{bmatrix}^\top.
\end{align}
We will show that \eqref{eq:SVD} provides an SVD of $C_S\Vb_S\bPhi_S^\top+C_R\Vb_R\bPhi_R^\top$, and it suffices to show that the first and last terms are left and right singular vectors:

\noindent\textbf{Left singular vectors.}
We notice that
\begin{align*}
&\begin{bmatrix}
\sign(C_S)\cdot\Vb_S\Qb_S & \sign(C_R)\cdot\Vb_R\Qb_R
\end{bmatrix}^\top\cdot\begin{bmatrix}
\sign(C_S)\cdot\Vb_S\Qb_S & \sign(C_R)\cdot\Vb_R\Qb_R
\end{bmatrix}\\
&=\begin{bmatrix}
\Qb_S^\top\Vb_S^\top\Vb_S\Qb_S & \sign(C_SC_R)\cdot\Qb_S^\top\Vb_S^\top\Vb_R\Qb_R\\
\sign(C_SC_R)\cdot\Qb_R^\top\Vb_R^\top\Vb_S\Qb_S & \Qb_R^\top\Vb_R^\top\Vb_R\Qb_R
\end{bmatrix},
\end{align*}
where $\Vb_S^\top\Vb_R=\Vb_R^\top\Vb_S=\zero$, and
\begin{align*}
&\Qb_S^\top\Vb_S^\top\Vb_S\Qb_S=\Qb_S^\top(\Ib_S-\one_{S, S}/S)\Qb_S=\Qb_S^\top\Qb_S-(\one_S^\top\Qb_S)^\top(\one_S^\top\Qb_S)/S=\Ib_{S-1},\\
&\Qb_R^\top\Vb_R^\top\Vb_R\Qb_R=\Qb_R^\top(\Ib_R-\one_{R, R}/S)\Qb_R=\Qb_R^\top\Qb_R-(\one_R^\top\Qb_R)^\top(\one_R^\top\Qb_R)/R=\Ib_{R-1}.
\end{align*}
Here we used the properties that $\one_S^\top\Qb_S=\zero$ and $\one_R^\top\Qb_R=\zero$ as well as $\Qb_S^\top\Qb_S=\Ib_{S-1}$ and $\Qb_R^\top\Qb_R=\Ib_{R-1}$.

\noindent\textbf{Right singular vectors.}
We notice that
\begin{align*}
&\begin{bmatrix}
\bPhi_S\Qb_S & \bPhi_R\Qb_R
\end{bmatrix}^\top\begin{bmatrix}
\bPhi_S\Qb_S & \bPhi_R\Qb_R
\end{bmatrix}=\begin{bmatrix}
\Qb_S^\top\bPhi_S^\top\bPhi_S\Qb_S & \Qb_S^\top\bPhi_S^\top\bPhi_R\Qb_R\\
\Qb_R^\top\bPhi_R^\top\bPhi_S\Qb_S & \Qb_R^\top\bPhi_R^\top\bPhi_R\Qb_R
\end{bmatrix}\\
&=\begin{bmatrix}
\Qb_S^\top\Qb_S & \Qb_S^\top\Qb_R\\
\Qb_R^\top\Qb_S & \Qb_R^\top\Qb_R
\end{bmatrix}=\begin{bmatrix}
\Ib_{S-1} & \zero\\
\zero & \Ib_{R-1}
\end{bmatrix}.
\end{align*}
where the second equality holds due to \eqref{eq:Phi_properties}.

Therefore, \eqref{eq:SVD} provides an SVD, so
\begin{align*}
&\msign(C_S\Vb_S\bPhi_S^\top+C_R\Vb_R\bPhi_R^\top)=\sign(C_S)\Vb_S\Qb_S\Qb_S^\top\bPhi_S^\top+\sign(C_R)\Vb_R\Qb_R\Qb_R^\top\bPhi_R^\top\\
&=\sign(C_S)\Vb_S\bPhi_S^\top+\sign(C_R)\Vb_R\bPhi_R^\top.
\end{align*}
where the second equality holds due to \eqref{eq:VQQ=V}.
\end{proof}

\subsection{Proof of Lemma~\ref{lemma:subject-object} and Lemma~\ref{lemma:relation-object}}

We only show the proof of Lemma~\ref{lemma:subject-object}, and the proof of Lemma~\ref{lemma:subject-object} is similar.
\begin{proof}[Proof of Lemma~\ref{lemma:subject-object}]
We first calculate $\tilde\vb_s$ under both types of embeddings.

\noindent\textbf{One-hot embeddings.} We firstly have
\begin{align*}
\bar\vb_s&=\frac1{\sqrt R}\sum_{r\in\cR}\bphi_A^{\mathrm{O}}(a_*(s, r))=\frac1{\sqrt R}\sum_{r\in\cR}\eb_s\otimes\eb_r=\frac1{\sqrt R}\eb_s\otimes\one_R;\\
\bar\vb&=\frac1{\sqrt{SR}}\sum_{s\in\cS}\sum_{r\in\cR}\bphi_A^{\mathrm{O}}(a_*(s, r))=\frac1{\sqrt{SR}}\sum_{s\in\cS}\sum_{r\in\cR}\eb_s\otimes\eb_r=\frac1{\sqrt{SR}}\one_S\otimes\one_R.
\end{align*}
We thus have
\begin{align*}
\tilde\vb_s=\bar\vb_s-\bar\vb/\sqrt S=(\eb_s-\one_S/S)\otimes\frac{\one_R}{\sqrt R}.
\end{align*}

\noindent\textbf{Hadamard embeddings.} We firstly have
\begin{align*}
\Hb_S\one_S=\Hb_2^{\otimes p}\one_2^{\otimes p}=(\Hb_2\one_2)^{\otimes p}=(\sqrt2\eb_1)^p=\sqrt S\eb_1.
\end{align*}
Similarly, $\Hb_R\one_R=\sqrt R\eb_1$. We thus have
\begin{align*}
\bar\vb_s&=\frac1{\sqrt R}\sum_{r\in\cR}\bphi_A^{\mathrm{H}}(a_*(s, r))=\frac1{\sqrt R}\sum_{r\in\cR}(\Hb_S\eb_s)\otimes(\Hb_R\eb_r)\\
&=(\Hb_S\eb_s)\otimes\frac{\Hb_R\one_R}{\sqrt R}=(\Hb_S\eb_s)\otimes\eb_1;\\
\bar\vb&=\frac1{\sqrt{SR}}\sum_{s\in\cS}\sum_{r\in\cR}\bphi_A^{\mathrm{H}}(a_*(s, r))=\frac1{\sqrt{SR}}\sum_{s\in\cS}\sum_{r\in\cR}(\Hb_S\eb_s)\otimes(\Hb_R\eb_r)\\
&=\frac{\Hb_S\one_S}{\sqrt S}\otimes\frac{\Hb_R\one_R}{\sqrt R}=\eb_1\otimes\eb_1.
\end{align*}
Therefore,
\begin{align*}
\tilde\vb_s=\bar\vb_s-\bar\vb/\sqrt S=(\Hb_S\eb_s-\eb_1/\sqrt S)\otimes\eb_1.
\end{align*}
We then study the four possible combinations.

\noindent\textbf{Subject one-hot, answer one-hot.} In this case,
\begin{align*}
\sum_{s\in\cS}\tilde\vb_s\bphi_S^{\mathrm{O}}(s)^\top&=\sum_{s\in\cS}\Big[\Big(\eb_s-\frac{\one_S}{S}\Big)\otimes\frac{\one_R}{\sqrt R}\Big]\cdot\eb_s^\top\\
&=\bigg[\sum_{s\in\cS}\Big(\eb_s-\frac{\one_S}{S}\Big)\eb_s^\top\bigg]\otimes\frac{\one_R}{\sqrt R}\\
&=(\Ib_S-\one_{S, S}/S)\otimes\frac{\one_R}{\sqrt R}.
\end{align*}
Thus, the element-wise sign is
\begin{align*}
\sign\bigg(\sum_{s\in\cS}\tilde\vb_s\bphi_S^{\mathrm{O}}(s)^\top\bigg)=(2\Ib_S-\one_{S, S})\otimes\one_R.
\end{align*}
Since
\begin{align*}
\sum_{s\in\cS}\bar\vb_s\bphi_S^{\mathrm{O}}(s)^\top&=\sum_{s\in\cS}\Big(\eb_s\otimes\frac{\one_R}{\sqrt R}\Big)\eb_s^\top=\frac{1}{\sqrt R}\Ib_S\otimes\one_R,\\
\sum_{s\in\cS}\bar\vb\bphi_S^{\mathrm{O}}(s)^\top&=\sum_{s\in\cS}\frac{1}{\sqrt{SR}}(\one_S\otimes\one_R)\eb_s^\top=\frac{1}{\sqrt SR}\one_{S, S}\otimes\one_R,
\end{align*}
we have
\begin{align*}
\sum_{s\in\cS}\tilde\vb_s\bphi_S^{\mathrm{O}}(s)^\top&=\sum_{s\in\cS}(2\sqrt R\cdot\bar\vb_s-\sqrt{SR}\cdot\bar\vb)\bphi_S^{\mathrm{O}}(s)^\top.
\end{align*}

\noindent\textbf{Subject one-hot, answer Hadamard.} In this case,
\begin{align*}
\sum_{s\in\cS}\tilde\vb_s\bphi_S^{\mathrm{O}}(s)^\top&=\sum_{s\in\cS}[(\Hb_S\eb_s-\eb_1/\sqrt S)\otimes\eb_1]\eb_s^\top\\
&=\bigg[\sum_{s\in\cS}(\Hb_S\eb_s-\eb_1/\sqrt S)\eb_s^\top\bigg]\otimes\eb_1\\
&=(\Hb_S-\eb_1\one_S^\top/\sqrt S)\otimes\eb_1.
\end{align*}
Note that the first row of $\Hb_S$ is exactly $\one_S^\top/\sqrt S$, so the first row of $\Hb_S-\eb_1\one_S^\top/\sqrt S$ is zero. All other entries of the matrix are $\pm1/\sqrt S$.
Thus, the element-wise sign is
\begin{align*}
\sign\bigg(\sum_{s\in\cS}\tilde\vb_s\bphi_S^{\mathrm{O}}(s)^\top\bigg)=\sqrt S\sum_{s\in\cS}\tilde\vb_s\bphi_S^{\mathrm{O}}(s)^\top
\end{align*}

\noindent\textbf{Subject Hadamard, answer one-hot.} In this case,
\begin{align*}
\sum_{s\in\cS}\tilde\vb_s\bphi_S^{\mathrm{O}}(s)^\top&=\sum_{s\in\cS}\Big[\Big(\eb_s-\frac{\one_S}{S}\Big)\otimes\frac{\one_R}{\sqrt R}\Big]\cdot(\Hb_S\eb_s)^\top\\
&=\bigg[\sum_{s\in\cS}\Big(\eb_s-\frac{\one_S}{S}\Big)\eb_s^\top\Hb_S\bigg]\otimes\frac{\one_R}{\sqrt R}\\
&=(\Hb_S-\one_{S, S}\Hb_S/S)\otimes\frac{\one_R}{\sqrt R}\\
&=(\Hb_S-\one_S\eb_1^\top/\sqrt S)\otimes\frac{\one_R}{\sqrt R}.
\end{align*}
Similar to the previous case, the first column of $\Hb_S-\one_S\eb_1^\top/\sqrt S$ is zero, and all other entries are $\pm1/\sqrt S$. Considering the additional $1/\sqrt R$ factor on $\one/\sqrt R$, we have
\begin{align*}
\sign\bigg(\sum_{s\in\cS}\tilde\vb_s\bphi_S^{\mathrm{O}}(s)^\top\bigg)=\sqrt{SR}\sum_{s\in\cS}\tilde\vb_s\bphi_S^{\mathrm{O}}(s)^\top.
\end{align*}

\noindent\textbf{Subject Hadamard, answer Hadamard.} In this case,
\begin{align*}
\sum_{s\in\cS}\tilde\vb_s\bphi_S^{\mathrm{O}}(s)^\top&=\sum_{s\in\cS}[(\Hb_S\eb_s-\eb_1/\sqrt S)\otimes\eb_1](\Hb_S\eb_s)^\top\\
&=\bigg[\sum_{s\in\cS}(\Hb_S\eb_s-\eb_1/\sqrt S)\eb_s^\top\Hb_S\bigg]\otimes\eb_1\\
&=(\Hb_S^2-\eb_1\one_S^\top\Hb_S/\sqrt S)\otimes\eb_1.
\end{align*}
Since
\begin{align*}
\Hb_S^2=[(\Hb_2/\sqrt 2)^2]^{\otimes p}=(\Ib_2)^{\otimes p}=\Ib_S,
\end{align*}
we have
\begin{align*}
\sum_{s\in\cS}\tilde\vb_s\bphi_S^{\mathrm{O}}(s)^\top=(\Ib_S-\eb_1\eb_1^\top)\otimes\eb_1.
\end{align*}
Therefore,
\begin{align*}
\sign\bigg(\sum_{s\in\cS}\tilde\vb_s\bphi_S^{\mathrm{O}}(s)^\top\bigg)=\sum_{s\in\cS}\tilde\vb_s\bphi_S^{\mathrm{O}}(s)^\top.
\end{align*}
\end{proof}

\subsection{Proof of Lemma~\ref{lemma:betaS_lower_betaR_upper}}

\begin{proof}[Proof of Lemma~\ref{lemma:betaS_lower_betaR_upper}]
We first show that $\sqrt S\beta_S\ge\sqrt R\beta_R$. If $S=R$, then $\beta_S\equiv\beta_R$. We now assume that $S>R$. At initialization, we have $\sqrt S\beta_0>\sqrt R\beta_0$. Suppose there exists time when $\sqrt S\beta_S\le\sqrt R\beta_R$, then define
\begin{align*}
t_0=\inf\{t: \sqrt S\beta_S(t)\le\sqrt R\beta_R\}.
\end{align*}
Note that the derivative of $\sqrt S\beta_S-\sqrt R\beta_R$ satisfies
\begin{align*}
\frac{\ud(\sqrt S\beta_S-\sqrt R\beta_R)}{\ud t}&=\frac{e^{q_S}}{\sqrt{(1+e^{q_S})^2+(1+e^{q_R})^2}}(e^{q_R-q_R}-1)\\
&=\frac{e^{q_S}}{\sqrt{(1+e^{q_S})^2+(1+e^{q_R})^2}}\Big(\frac{S-1}{R-1}e^{\frac{t+\alpha_0}{\sqrt{SR}}(\sqrt S\beta_S-\sqrt R\beta_R)}-1\Big)
\end{align*}
where the last equality holds due to the definition of $q_S$ and $q_R$ as well as $\alpha_S(t)=\alpha_R(t)=t+\alpha_0$. Since at $t=t_0$, we have $\sqrt S\beta_S(t_0)-\sqrt R\beta_R(t_0)=0$, the derivative at $t=t_0$ satisfies
\begin{align*}
\frac{\ud(\sqrt S\beta_S-\sqrt R\beta_R)}{\ud t}\bigg|_{t=t_0}=\frac{e^{q_S}}{\sqrt{(1+e^{q_S})^2+(1+e^{q_R})^2}}\cdot\frac{S-R}{R-1}>0.
\end{align*}
This means that there exists $\epsilon>0$ such that $\sqrt S\beta_S(t_0-\epsilon)-\sqrt R\beta_R(t_0-\epsilon)<0$, which contradicts against definition of $t_0$. Therefore, $\sqrt S\beta_S\ge\sqrt R\beta_R$ at any time, and we have
\begin{align}\label{eq:sum_bounds_2}
2\sqrt R\beta_R(t)\le\sqrt S\beta_S(t)+\sqrt R\beta_R(t)\le2\sqrt S\beta_S(t).
\end{align}

We now get back to the proof of the upper and lower bounds, and we notice that
\begin{align*}
\Big[\frac{\ud(\sqrt S\beta_S)}{\ud t}\Big]^2+\Big[\frac{\ud(\sqrt R\beta_R)}{\ud t}\Big]^2=1.
\end{align*}
Note that for positive numbers $A, B$, we have $\sqrt{A^2+B^2}\le A+B\le\sqrt{2(A^2+B^2)}$. Therefore,
\begin{align*}
1\le\frac{\ud(\sqrt S\beta_S+\sqrt R\beta_R)}{\ud t}\le\sqrt2.
\end{align*}
Taking the integral, we have
\begin{align}\label{eq:sum_bounds_1}
t+(\sqrt S+\sqrt R)\beta_0\le\sqrt S\beta_S(t)+\sqrt R\beta_R(t)\le\sqrt2t+(\sqrt S+\sqrt R)\beta_0.
\end{align}
Combining \eqref{eq:sum_bounds_1} and \eqref{eq:sum_bounds_2}, we have
\begin{align*}
\beta_S(t)&\ge\frac{t}{2\sqrt S}+\frac{1+\sqrt{R/S}}{2}\beta_0,\\
\beta_R(t)&\le\frac{t}{\sqrt{2R}}+\frac{1+\sqrt{S/R}}{2}\beta_0.
\end{align*}
\end{proof}

\subsection{Proof of Lemma~\ref{lemma:T0.5}}

\begin{proof}[Proof of Lemma~\ref{lemma:T0.5}]
For any $t\le T_S^{0.5}$, we have $q_S(t)\le0$. Thus, for any $t\le T_S^{0.5}$, the time derivative of $\beta_S$ satisfies
\begin{align*}
\frac{\ud\beta_S(t)}{\ud t}&=\frac{(1+e^{q_R})/\sqrt S}{\sqrt{(1+e^{q_S})^2+(1+e^{q_R})^2}}\le\frac{(1+e^{q_R})/\sqrt S}{\sqrt{4+(1+e^{q_R})^2}}\le\frac1{\sqrt{5S}},
\end{align*}
where the first inequality holds because $e^{q_S}\le1$, and the second inequality holds because $e^{q_R}\ge0$. We also trivially have $\ud\beta_S/\ud t\le1/\sqrt S$. Therefore, for any $t\le T_S^{0.5}$, we have
\begin{align*}
\beta_S(t)\simeq t/\sqrt{S}+\beta_0.
\end{align*}
According to the definition of $T_S^{0.5}$, we have
\begin{align*}
\log(S-1)=\frac{\alpha_S(T_S^{0.5})\cdot\beta_S(T_S^{0.5})}{\sqrt R}\simeq\frac{(t+\alpha_0)(t/\sqrt S+\beta_0)}{\sqrt R}
\end{align*}
Due to the assumptions of $\alpha_0\le(SR)^{1/4}$ and $\beta_0\le(R/S)^{1/4}$, we have
\begin{align*}
T_S^{0.5}\simeq(SR)^{1/4}\sqrt{\log(S-1)},\quad\alpha_S(T_S^{0.5})\simeq(R/S)^{1/4}\sqrt{\log(S-1)}.
\end{align*}
We similarly obtain the expressions of the relation-dependent learning time and manifold parameters.
\end{proof}

\section{Auxiliary Lemmas}
\begin{lemma}\label{lemma:msign_rotation}
Let $\Mb\in\RR^{m\times n}$ be any matrix, and $\Qb_1\in\RR^{m\times m}$, $\Qb_2\in\RR^{n\times n}$ be any orthogonal matrices. Then
\begin{align*}
\msign(\Qb_1\Mb\Qb_2^\top)=\Qb_1\cdot\msign(\Mb)\cdot\Qb_2^\top.
\end{align*}
\end{lemma}
\begin{proof}
Let the SVD of $\Mb$ be $\Mb=\Ub\bSigma\Vb^\top$. Then 
\begin{align*}
\Qb_1\Mb\Qb_2^\top=(\Qb_1\Ub)\cdot\bSigma\cdot(\Qb_2\Vb)^\top.
\end{align*}
Note that $\Qb_1\Ub$ and $\Qb_2\Vb$ are left and right singular vectors because
\begin{align*}
(\Qb_1\Ub)^\top(\Qb_1\Ub)&=\Ub^\top(\Qb_1^\top\Qb_1)\Ub=\Ub^\top\Ub=\Ib_r,\\
(\Qb_2\Vb)^\top(\Qb_2\Vb)&=\Vb^\top(\Qb_2^\top\Qb_2)\Vb=\Vb^\top\Vb=\Ib_r.
\end{align*}
Therefore, the spectral orthogonalization of $\Qb_1\Mb\Qb_2^\top$ is
\begin{align*}
\msign(\Qb_1\Mb\Qb_2^\top)=(\Qb_1\Ub)(\Qb_2\Vb^\top)=\Qb_1\cdot\msign(\Mb)\cdot\Qb_2^\top.
\end{align*}
\end{proof}

\begin{lemma}\label{lemma:barv_properties}
Let $\bar\vb_s$, $\bar\vb_r$, and $\bar\vb$ be defined in~\eqref{eq:def_barv}, and $\tilde\vb_s$, $\tilde\vb_r$ be defined in~\eqref{eq:def_tildev}. Then the following properties hold
\begin{itemize}[leftmargin=*]
\item[(a).] For any $s', r'$, we have
\begin{align*}
\bar\vb_s^\top\bphi(a_*(s', r'))=\delta_{s, s'}/\sqrt R,\quad\bar\vb_r^\top\bphi(a_*(s', r'))=\delta_{r, r'}/\sqrt S,\quad\bar\vb^\top\bphi(a_*(s', r'))=1/\sqrt{SR}.
\end{align*}
\item[(b).] For any $s, r$ and $s', r'$, we have
\begin{gather*}
\bar\vb_{s'}^\top\bar\vb_s=\delta_{s, s'},\quad\bar\vb_{r'}^\top\bar\vb_r=\delta_{r, r'},\quad\bar\vb_s^\top\bar\vb_r=\frac1{\sqrt{SR}},\\
\bar\vb_s^\top\bar\vb=1/\sqrt S,\quad\bar\vb_r^\top\bar\vb=1/\sqrt R,\quad\bar\vb^\top\bar\vb=1.
\end{gather*}
\item[(c).] For any $s, r$ and $s', r'$, we have
\begin{align*}
\tilde\vb_{s'}^\top\tilde\vb_s=\delta_{s, s'}-1/S,\quad\tilde\vb_{r'}^\top\tilde\vb_r=\delta_{r, r'}-1/R,\quad\tilde\vb_s^\top\tilde\vb_r=0.
\end{align*}
\end{itemize}
\end{lemma}
\begin{proof}
\noindent\textbf{Part (a).} This part holds due to the definition in \eqref{eq:def_barv} and because embeddings $\bphi(a)$ are orthonormal.

\noindent\textbf{Part (b).}
We firstly have
\begin{align*}
\bar\vb_{s'}\bar\vb_s=\frac1{\sqrt R}\sum_{r'\in\cR}\bphi(a_*(s', r'))^\top\bar\vb_s=\frac1{\sqrt R}\cdot R\cdot\frac{\delta_{s, s'}}{\sqrt R}=\delta_{s, s'},
\end{align*}
where the second equality holds due to Part (a). Similarly, we have $\bar\vb_{r'}^\top\bar\vb_r=\delta_{r, r'}$. For $\bar\vb_s^\top\bar\vb_r$, we have
\begin{align*}
\bar\vb_s^\top\bar\vb_r=\frac1{\sqrt R}\sum_{r'\in\cR}\bphi(a_*(s, r'))^\top\bar\vb_r=\frac1{\sqrt R}\sum_{r'\in\cR}\frac{\delta_{r, r'}}{\sqrt S}=\frac1{\sqrt{SR}},
\end{align*}
where the second equality holds due to Part (a). Furthermore,
\begin{align*}
\bar\vb_s^\top\bar\vb=\frac1{\sqrt R}\sum_{r'\in\cR}\bphi(a_*(s, r'))^\top\bar\vb=\frac1{\sqrt R}\cdot R\cdot\frac1{\sqrt{SR}}=\frac1{\sqrt S}.
\end{align*}
Similarly, we have $\bar\vb_r^\top\bar\vb=1/\sqrt R$. Finally, we have
\begin{align*}
\bar\vb^\top\bar\vb=\frac1{\sqrt{SR}}\sum_{s', r'}\bphi(a_*(s, r))^\top\bar\vb=\frac1{\sqrt{SR}}\cdot SR\cdot\frac1{\sqrt{SR}}=1.
\end{align*}

\noindent\textbf{Part (c).} We firstly have
\begin{align*}
\tilde\vb_{s'}^\top\tilde\vb_s&=(\bar\vb_{s'}-\bar\vb/\sqrt S)^\top(\bar\vb_s-\bar\vb/\sqrt S)\\
&=\bar\vb_{s'}^\top\bar\vb_s-\frac{\bar\vb_{s'}^\top\bar\vb}{\sqrt S}-\frac{\bar\vb_s^\top\bar\vb}{\sqrt S}+\frac{\bar\vb^\top\bar\vb}{S}\\
&=\delta_{s, s'}-\frac1{\sqrt S}\cdot\frac1{\sqrt S}-\frac1{\sqrt S}\cdot\frac1{\sqrt S}+\frac1S\cdot1\\
&=\delta_{s, s'}-\frac1S,
\end{align*}
where the third equality holds due to Part (b).
Similarly, $\tilde\vb_{r'}^\top\tilde\vb_r=\delta_{r, r'}-1/R$. Finally,
\begin{align*}
\tilde\vb_s^\top\tilde\vb_r&=\bar\vb_s^\top\bar\vb_r-\frac{\bar\vb_s^\top\bar\vb}{\sqrt R}-\frac{\bar\vb_r^\top\bar\vb}{\sqrt S}+\frac{\bar\vb^\top\bar\vb}{\sqrt{RS}}\\
&=\frac1{\sqrt{SR}}
\end{align*}
where the second equality holds due to Part (c).
\end{proof}

\begin{lemma}\label{lemma:Ei_estimate}
Assume that $M\ge2$, then
\begin{align*}
\int_1^M\frac{e^u}{u}\ud u\simeq\frac{e^M}{M}.
\end{align*}
\end{lemma}
\begin{proof}
Applying integral by parts, we have
\begin{align*}
\int_1^M\frac{e^u}{u}\ud u=\frac{e^M}{M}-e+\int_1^M\frac{e^u}{u^2}\ud u.
\end{align*}
Since $M\ge2$, we have
\begin{align*}
\int_1^M\frac{e^u}{u}\ud u\ge\frac{e^M}{M}-e\simeq\frac{e^M}{M}.
\end{align*}
It then suffices to prove that
\begin{align*}
\int_1^M\frac{e^u}{u^2}\ud u\lesssim\frac{e^M}{M}.
\end{align*}
To prove this, we notice that
\begin{align*}
\int_1^M\frac{e^u}{u^2}\ud u=\underbrace{\int_1^{M/2}\frac{e^u}{u^2}\ud u}_{J_1}+\underbrace{\int_{M/2}^M\frac{e^u}{u^2}\ud u}_{J_2}.
\end{align*}
For the term $J_1$, we have
\begin{align*}
J_1\le e^{M/2}\int_1^{M/2}\frac{\ud u}{u^2}\le e^{M/2}\int_1^\infty\frac{\ud u}{u^2}=e^{M/2}\lesssim\frac{e^M}{M}.
\end{align*}
where the first inequality holds because $u\le M/2$. For the term $J_2$, we have
\begin{align*}
J_2&\le\int_{M/2}^M\frac{e^u\ud u}{(M/2)^2}\le\frac{4e^M}{M^2}\lesssim\frac{e^M}{M}.
\end{align*}
where the first inequality holds because $u\ge M/2$, and the second inequality holds because $\int_{M/2}^Me^u\ud u=e^M-e^{M/2}\le e^M$.
\end{proof}

\bibliography{arxiv}
\bibliographystyle{ims}

\end{document}